\pdfoutput=1

\documentclass[11pt]{article}
\usepackage{arabtex}
\usepackage[table]{xcolor}

\usepackage[final]{acl}

\usepackage{times}
\usepackage{latexsym}
\usepackage{booktabs}
\usepackage{worldflags}

\usepackage{twemojis}
\usepackage{scalerel}

\usepackage{listings}

\usepackage{booktabs}
\usepackage{array}  % Required for the 'p' column type
\usepackage{arydshln} % Import the arydshln package for dashed lines in tables
\usepackage{xcolor, colortbl}
\usepackage{tcolorbox}
\usepackage{color,soul} 

\usepackage{graphicx}
\usepackage{xspace}

\newcommand{\ourdataset}{\textsc{Palm}\xspace}

\usepackage[T1]{fontenc}
\usepackage[utf8]{inputenc}

\usepackage{microtype}

\usepackage{inconsolata}

\usepackage{graphicx}
\usepackage{soul}
\usepackage{amssymb}
\usepackage{appendix}
\usepackage{amsmath}  % Import the amsmath package for \numberwithin
\usepackage{arabtex}

\usepackage{multirow,array}
\usepackage{subcaption}  % This is the modern package for subfigures
\usepackage{amssymb}% http://ctan.org/pkg/amssymb
\usepackage{pifont}% http://ctan.org/pkg/pifont
\newcommand{\bluecheck}{\color{teal}\ding{51}}%
\newcommand{\redxmark}{\color{red}\ding{55}}%

\usepackage{utf8}
\usepackage{placeins}  % Add this in the preamble
\setcode{utf8}
\usepackage{capt-of} % Allows captions outside floating environments
\usepackage{caption}   % for \captionof
\usepackage{cuted}

\title{
\raisebox{-2.1ex}{\protect\includegraphics[height=3.8\fontcharht\font`\B]{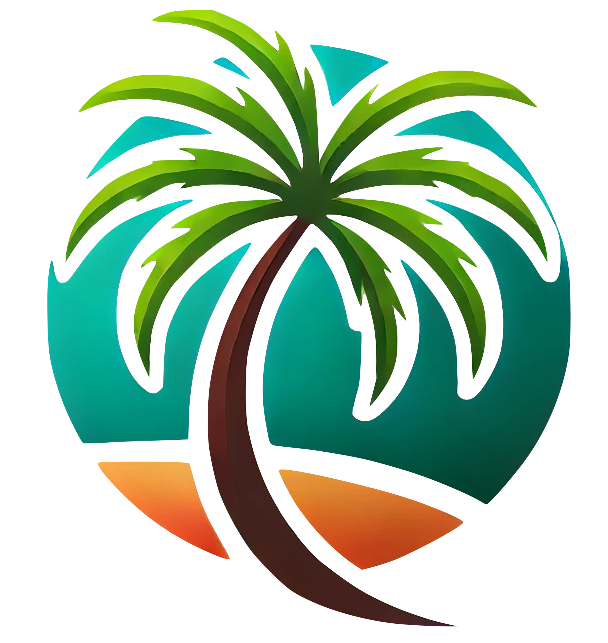}} Palm: A Culturally Inclusive and Linguistically Diverse Dataset
for Arabic LLMs}

\author{
    \begin{minipage}[t]{\textwidth}
        \centering
        \normalfont
        Fakhraddin Alwajih\textsuperscript{1}\thanks{~~Equal contribution}, 
        Abdellah El Mekki\textsuperscript{1,2}\footnotemark[1],  
        Samar Mohamed Magdy\textsuperscript{1,2},
        Abdelrahim A. Elmadany\textsuperscript{1}, 
        Omer Nacar\textsuperscript{5}, 
        ElMoatez Billah Nagoudi\textsuperscript{1,5}, 
        Reem Abdel-Salam\textsuperscript{7}, 
        Hanin Atwany\textsuperscript{2},
        Youssef Nafea\textsuperscript{2},
        Abdulfattah Mohammed Yahya\textsuperscript{7}, 
        Rahaf Alhamouri\textsuperscript{9}, 
        Hamzah A. Alsayadi\textsuperscript{10}, 
        Hiba Zayed\textsuperscript{4}, 
        Sara Shatnawi\textsuperscript{2}, 
        Serry Sibaee\textsuperscript{5}, 
        Yasir Ech-Chammakhy\textsuperscript{6}, 
        Walid Al-Dhabyani\textsuperscript{7}, 
        Marwa Mohamed Ali\textsuperscript{10}, 
        Imen Jarraya\textsuperscript{5}, 
        Ahmed Oumar El-Shangiti\textsuperscript{2}, 
        Aisha Alraeesi\textsuperscript{2}, 
        Mohammed Anwar Al-Ghrawi\textsuperscript{11}, 
        Abdulrahman S. Al-Batati\textsuperscript{5}, 
        Elgizouli Mohamed\textsuperscript{12}, 
        Noha Taha Elgindi\textsuperscript{13}, 
        Muhammed Saeed\textsuperscript{2}, 
        Houdaifa Atou\textsuperscript{6}, 
        Issam Ait Yahia\textsuperscript{6}, 
        Abdelhak Bouayad\textsuperscript{6}, 
        Mohammed Machrouh\textsuperscript{6}, 
        Amal Makouar\textsuperscript{6}, 
        Dania Alkawi\textsuperscript{5}, 
        Mukhtar Mohamed\textsuperscript{2}, 
        Safaa Taher Abdelfadil\textsuperscript{2}, 
        Amine Ziad Ounnoughene\textsuperscript{15}, 
        Rouabhia Anfel\textsuperscript{16}, 
        Rwaa Assi\textsuperscript{4}, 
        Ahmed Sorkatti\textsuperscript{12}, 
        Mohamedou Cheikh Tourad\textsuperscript{14}, 
        Anis Koubaa\textsuperscript{17}, \\
        Ismail Berrada\textsuperscript{6}, 
        Mustafa Jarrar\textsuperscript{4,18}, 
        Shady Shehata\textsuperscript{2,3},\\
        Muhammad Abdul-Mageed\textsuperscript{1,2,3} \\ [1em]
        {
            \textsuperscript{1} The University of British Columbia, 
            \textsuperscript{2} MBZUAI, 
            \textsuperscript{3} Invertible AI, \\
            \textsuperscript{4} Birzeit University, 
            \textsuperscript{5} Prince Sultan University, \\
            \textsuperscript{6} UM6P,
            \textsuperscript{7} Cairo University,
            \textsuperscript{8} Prince Sultan University,\\
            \textsuperscript{9} JUST,
            \textsuperscript{10} Ain Shams University,
            \textsuperscript{11} Damascus University,\\
            \textsuperscript{12} University of Khartoum,
            \textsuperscript{13} Menoufiya University,
            \textsuperscript{14} University of Nouakchott,\\
            \textsuperscript{15} National Polytechnic School of Algiers,
            \textsuperscript{16} Full Sail University,
            \textsuperscript{17} Alfaisal University,\\
            \textsuperscript{18} Hamad Bin Khalifa University 
        } \\ [0.5em]
        \texttt{\normalsize \{fakhr.alwajih, muhammad.mageed\}@ubc.ca}
    \end{minipage}
}

\begin{document}

\onecolumn

\maketitle

\leavevmode
\vspace{350pt}

\begin{strip}
\centering
\includegraphics[width=0.90\textwidth]{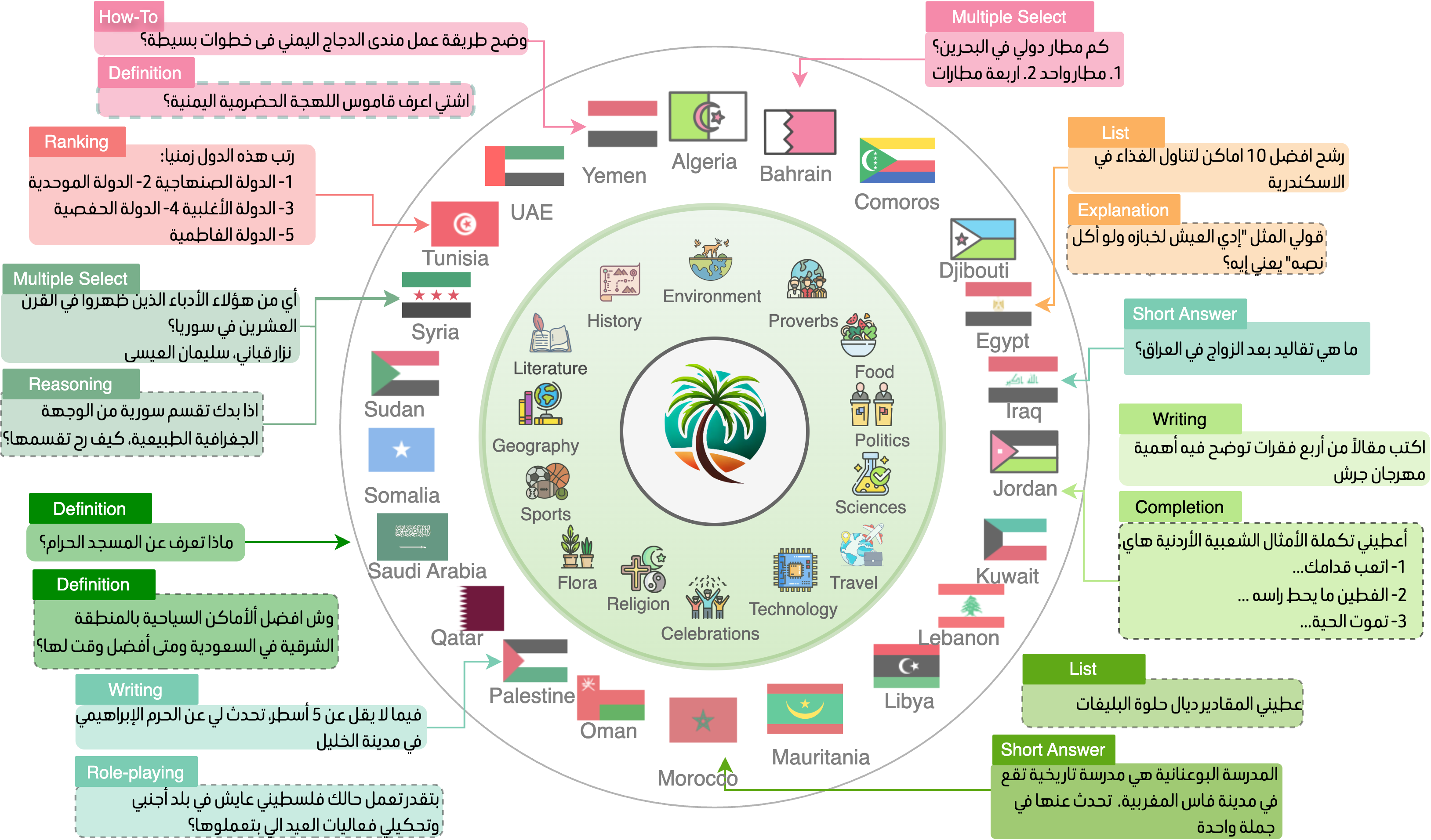}
\captionof{figure}{\ourdataset spans all $22$ Arab countries, each represented by its own flag, across $20$ diverse areas, including local \textit{celebrations}, \textit{geography}, and \textit{history}. Instructions (input–response pairs) are human-created at the country level. Dashed examples represent local dialects, whereas others use MSA. Only a subset of domains and instruction types is shown here, and we include only example inputs (not responses) due to space constraints. English translations of the Arabic instructions are in Appendix~\ref{apndx:examples}.}
\label{fig:main}
\end{strip}

% Force the figure to appear after the title content with enough space
% \clearpage
\twocolumn
\begin{abstract}

As large language models (LLMs) become increasingly integrated into daily life, ensuring their cultural sensitivity and inclusivity is paramount. We introduce \ourdataset, a year-long community-driven project covering \textit{all 22} Arab countries. The dataset includes instructions (input, response pairs) in both Modern Standard Arabic (MSA) and dialectal Arabic (DA), spanning 20 diverse topics. Built by a team of 44 researchers across the Arab world, all of whom are authors of this paper, \ourdataset~offers a broad, inclusive perspective. We use \ourdataset~to evaluate the cultural and dialectal capabilities of several frontier LLMs, revealing notable limitations. For instance, while closed-source LLMs generally exhibit strong performance, they are not without flaws, and smaller open-source models face greater challenges. Moreover, certain countries (e.g., Egypt, the UAE) appear better represented than others (e.g., Iraq, Mauritania, Yemen). Our annotation guidelines, code, and data for reproducibility are publicly available. More information about \ourdataset is available at our project page: \href{https://github.com/UBC-NLP/palm}{https://github.com/UBC-NLP/palm}.
\end{abstract}

\section{Introduction}

LLMs have become pervasive across a wide range of applications. These models are typically trained to predict the next token in a sequence~\cite{radford_improving_2018}, followed by a fine-tuning phase where they learn to respond to human prompts using instruction datasets \cite{ouyang2022traininglanguagemodelsfollow}. However, the responses generated by these models are often biased toward the data they were pre-trained or fine-tuned on, which may not reflect the values and cultures of diverse end users \cite{shankar2017classificationrepresentationassessinggeodiversity, xu2024surveymultilinguallargelanguage, naous-etal-2024-beer}. LLMs pre-trained on translated data from English often exhibit Western, Anglocentric, and American biases. For example, an Arabic LLM pre-trained on English-to-Arabic translated data suggested having a beer after prayer \cite{naous-etal-2024-beer}, a recommendation that starkly contradicts Arab cultural values, religious practices, and social norms. This example highlights the critical need for building LLMs that are culturally and linguistically aware, which requires the inclusion of more diverse global communities in their development \cite{adilazuarda2024measuringmodelingculturellms}. However, having a benchmarking tool for cultural and linguistic coverage for LLMs is a crucial phase.

In this work, we focus on the Arab world and its communities which span a vast geographical region across Africa and Asia, with a population exceeding $450$ million people in $22$ countries. It is home to a diverse array of local cultures, customs, traditions, political systems, and social practices. The linguistic landscape is equally rich: Arabic exists in three main forms—Classical Arabic, MSA, and DA, with MSA and DA being the most widely used today. MSA, the standardized form used in formal settings such as literature, media, and official documents, contrasts sharply with DA, which is employed in everyday conversation and varies significantly across regions, sometimes classified at the country level~\cite{abdul-mageed-etal-2024-nadi}. Several LLMs have been pre-trained for Arabic, including Jasmine~\cite{billah-nagoudi-etal-2023-jasmine}, JAIS~\cite{sengupta2023jaisjaischatarabiccentricfoundation}, AceGPT~\cite{huang2024acegptlocalizinglargelanguage}, ALLAM~\cite{bari2024allamlargelanguagemodels}, Fanar~\cite{fanarteam2025fanararabiccentricmultimodalgenerative}, and NileChat~\cite{mekki2025nilechatlinguisticallydiverseculturally}.
These models demonstrate powerful capabilities in generating Arabic across its different forms. However, when it comes to instruction tuning, the datasets used for some of these models (such as \texttt{JAIS} and \texttt{AceGPT}) are predominantly machine-generated or machine-translated, resulting in a set of instructions that are not related to Arab culture. In addition, most of these models lack evaluation on Arabic country-specific cultural awareness for all Arab countries, as most of them were evaluated on general NLP tasks but lack evaluation on specific Arabic countries' cultures and dialects. Our work addresses this need by providing a large dataset of Arabic instructions to ensure better cultural representation of Arab communities.

More specifically, we introduce \ourdataset, the first comprehensive fully human-created Arabic instruction dataset that is both culturally and linguistically diverse and inclusive. \ourdataset is the first dataset at the country level to cover \textit{all} $22$ Arab countries, spanning $20$ culturally relevant topics. What sets {\ourdataset} apart is its inclusion of instructions in both MSA and local dialects, all of which are human-annotated using reliable, country-specific sources. This dataset was developed through a large community-driven project, leveraging local expertise and collective knowledge. \ourdataset serves a dual purpose: it can be used for cultural and dialectal instruction tuning of LLMs, as well as for evaluating their cultural competence regarding the Arab world.

We offer the following contributions:

\begin{enumerate}
\item We present \textbf{\ourdataset}, a novel dataset developed through a year-long collaborative community effort. It includes culturally informed instructions from \textbf{all $22$ Arab countries} in both \textbf{MSA} and \textbf{local dialects}, spanning multiple linguistic forms and topics.
\item We benchmark several open-source and frontier LLMs on \textbf{\ourdataset}, providing a detailed analysis of model performance on both MSA and dialectal data across various dimensions.
\item We offer a comprehensive analysis using three models as evaluative judges, examining their alignment and highlighting the reliability of automated evaluations.
\item We conduct a human evaluation to validate the consistency between automated and human judgments, demonstrating the effectiveness of automated methods for assessing culturally aware and dialect-specific Arabic instructions.
\end{enumerate}

\section{Related Work}
\paragraph{Arabic Varieties.} Arabic, with its rich linguistic diversity, has attracted increasing attention in NLP. This focus has propelled the development of models for \textit{encoding}~\cite{antoun-etal-2020-arabert, abdul-mageed-etal-2021-arbert, inoue-etal-2021-interplay} and \textit{generating}~\cite{billah-nagoudi-etal-2023-jasmine, sengupta2023jaisjaischatarabiccentricfoundation, huang2024acegptlocalizinglargelanguage,fanarteam2025fanararabiccentricmultimodalgenerative} Arabic text, yielding powerful results in both understanding and generation tasks \cite{elmadany-etal-2023-orca, seelawi-etal-2021-alue, nagoudi-etal-2023-dolphin}.
Nonetheless, a critical gap persists: the underrepresentation of Arabic dialects in current language models, which affects both performance and cultural inclusion~\cite{alkhamissi-etal-2024-investigating}. Consequently, there is a pressing need for more comprehensive Arabic language models that can capture both Modern Standard Arabic (MSA) and diverse colloquial dialects, reflecting the linguistic and cultural richness of the Arab world.
\begin{figure*}[ht]
    \centering
    \includegraphics[width=0.98\linewidth]{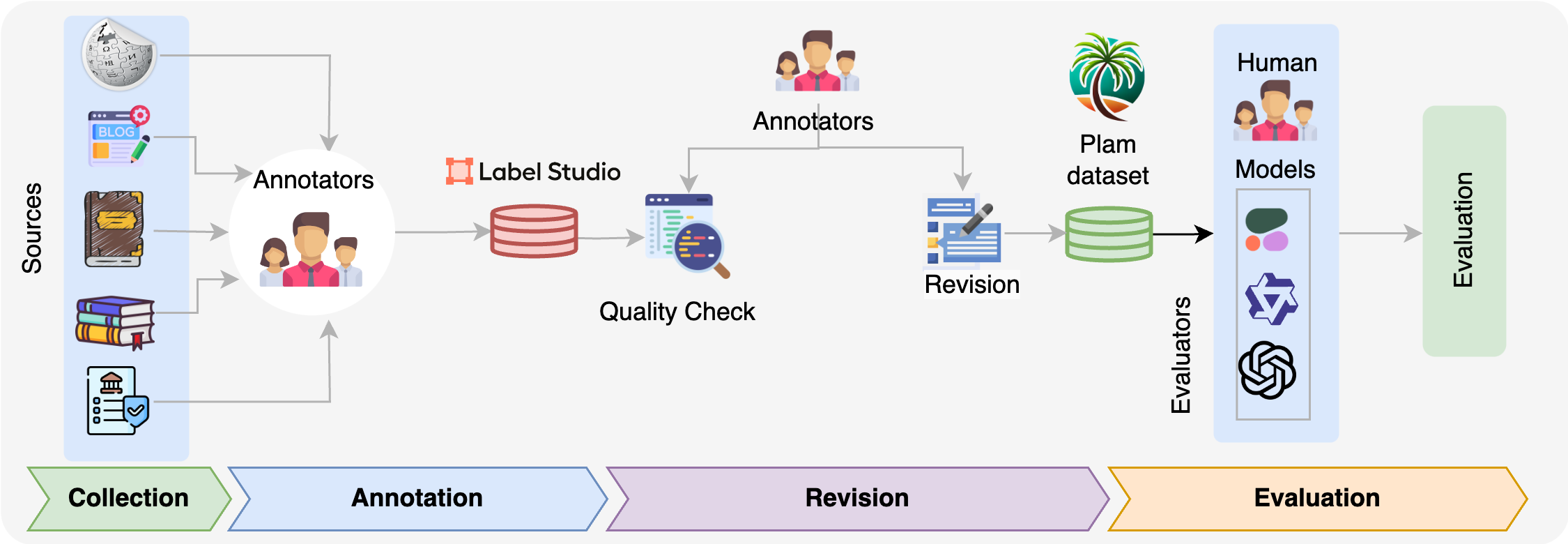}
\caption{The complete pipeline for \ourdataset creation, beginning with data collection from diverse sources, followed by annotation in Label Studio, quality checks, and subsequent revisions. Finally, \ourdataset undergoes both human and model evaluations, culminating in a final assessment phase.}
    \label{fig:pipeline}
\end{figure*}
\paragraph{Cultural Awareness in Language Models.} Recent research employs diverse methodologies to assess the cultural capabilities of language models \cite{adilazuarda2024measuringmodelingculturellms}. One approach constructs knowledge databases \cite{keleg-magdy-2023-dlama, shi2024culturebankonlinecommunitydrivenknowledge}, often drawing on online resources such as Wikipedia \cite{Nguyen_2023,fung2024massivelymulticulturalknowledgeacquisition} and web corpora \cite{Nguyen_2023}. However, these evaluations can be overly simplistic, since much of the web-scraped content might already appear in the models’ training data \cite{petroni-etal-2019-language}. Another methodology uses socio-cultural surveys like the World Value Survey \cite{ramezani-xu-2023-knowledge, alkhamissi-etal-2024-investigating}, which, while valuable, often cover only a narrow range of cultural concepts. To overcome these limitations, researchers have introduced new datasets and benchmarks tailored to evaluating the cultural capabilities of LLMs \cite{arora2024calmqaexploringculturallyspecific, myung2024blendbenchmarkllmseveryday, singh-etal-2024-aya}. 

\paragraph{Arabic Cultural Awareness in Language Models.} Recent studies have also increasingly focused on integrating Arabic language and culture into LLMs, alongside the development of relevant benchmarks~\cite{myung2024blendbenchmarkllmseveryday, singh-etal-2024-aya, alkhamissi-etal-2024-investigating, naous-etal-2024-beer, demidova-etal-2024-john, alwajih-etal-2024-peacock, alwajih2025pearlmultimodalculturallyawarearabic}. For instance,~\citet{alkhamissi-etal-2024-investigating} leveraged the World Values Survey to assess LLM alignment with Arab cultural values, revealing reduced alignment for underrepresented groups. Similarly, \citet{naous-etal-2024-beer} identified a significant Western bias in LLMs, attributing it to the prevalence of translated rather than original Arabic data in pre-training corpora. Researchers have also introduced new datasets. For example, \citet{mousi2024aradicebenchmarksdialectalcultural} proposed a benchmark comprising $180$ questions spanning nine topics and three Arabic regions, and \citet{alyafeai-etal-2024-cidar} introduced a localized dataset of $10{,}000$ instructions covering $17$ topics in MSA. However, much of this work relies on automatic annotation and remains centered on MSA, leaving gaps in dialectal and country-specific cultural representation. Moreover, most available benchmarks are limited in size and instruction formats, often focusing on multiple-choice questions \cite{mousi2024aradicebenchmarksdialectalcultural, fanarteam2025fanararabiccentricmultimodalgenerative}.

Our work,~\ourdataset, closes these gaps by presenting the first large-scale, fully human-curated collection of Arabic cultural input-output instruction pairs from all Arab countries. It encompasses multiple instruction types expressed in MSA and diverse Arabic varieties. Table~\ref{tab:arabic-datasets} compares \ourdataset to existing datasets involving any level of Arabic instructions.

\section{\ourdataset Dataset}
As stated earlier, \textit{\ourdataset} is a manually curated, culturally aligned dataset covering all $22$ Arab countries. It features a diverse set of instructions (input, output pairs) from both MSA and \textit{ten} different Arabic dialects, all entirely human-produced. \ourdataset comprises $20$ different \textbf{topical areas}, such as \textit{celebrations}, \textit{history}, \textit{geography}, \textit{literature}, \textit{politics}, \textit{proverbs}, and \textit{sports}, crafted at the local, country-specific, level or at the level of the whole Arab world (e.g., \textit{technology}). As such, \ourdataset represents a comprehensive view of the culture of Arabic local communities. Figure~\ref{fig:main} illustrates the composition of \ourdataset across different countries and areas. We now describe how we created~\ourdataset. 
% \usepackage{graphicx}

\begin{table*}[]
\centering
\resizebox{.98\textwidth}{!}{%
\begin{tabular}{lcccccc}
\toprule
                   & \multicolumn{2}{c}{\textbf{Multilingual}}   &                                                                            & \multicolumn{3}{c}{\textbf{Arabic Specific}}                                                              \\ \cmidrule{2-3} \cmidrule{5-7}

                                     & \textbf{AYA}   & \textbf{BLEnD}  &                                                               & \textbf{AraDiCE}         & \textbf{CIDAR} & \textbf{\colorbox{green!10}{\ourdataset (ours)}}   \\ \midrule

\# Arab countries     & -            &   $1$        &                                                                     & - & -             & $22$ \\
\# Arabic dialects    & -             & -  & &  $6$                 & -             & $10$            \\
\midrule 
Cultural coverage?           & limited        &   \bluecheck&                                                                           &    limited               & \bluecheck            & \bluecheck                    \\ 
Human collected?            & mixed&    \bluecheck   &                                                                              & \redxmark                & \redxmark             & \bluecheck                    \\
Human revised?               & \bluecheck            &     \redxmark  &                                                                          & \bluecheck                    & \bluecheck            & \bluecheck      \\ 
From scratch?               & mixed&     \bluecheck&                                                                          & \redxmark& \redxmark& \bluecheck      \\
Open classes?               & \bluecheck            &     \redxmark&                                                                          & \redxmark & \redxmark             & \bluecheck      \\

\midrule

\# Arabic instructions & $5$K out of $204$K &  $3.6$K out of $55$K &                                                                     & $180$ out of $45$k& $10$K ($100$\%)    & $18$K ($100$\%)            \\
\bottomrule             
\end{tabular}%
}
\caption{\ourdataset in comparison. \ourdataset is specifically designed to capture country-specific knowledge from all Arab countries, exceeding existing datasets in both geographic scope and dialectal diversity. Collected entirely by human annotators from scratch (unlike AraDice, which involves translation and data retargeting, and CIDAR, which relies on localization), \ourdataset is also the only Arabic dataset based on open-ended instructional prompts (e.g., writing instructions, role-playing, reasoning) rather than solely QA. Further details on how \ourdataset compares to other datasets are provided in Appendix~\ref{apndx:comparative_analysis}.}
\label{tab:arabic-datasets}
\end{table*}

\subsection{Team Structure}
\ourdataset is a community project involving 44 trained native speakers, all of whom are authors of this work. We aimed to incorporate local knowledge from every Arab country and succeeded for 15 out of 22. For each of these 15 countries, we assigned at least two annotators. For the remaining seven, two annotators from neighboring countries were chosen to ensure cultural familiarity.\footnote{Countries without local team members are Bahrain, Comoros, Djibouti, Iraq, Libya, Qatar, and Somalia.} Each team member holds at least a bachelor’s degree, with most having advanced degrees. Within each country, members hail from different regions, promoting inclusive cultural coverage and broad dialectal variety. To our knowledge, \ourdataset stands among the most comprehensive datasets of its kind in the Arab world, both culturally and linguistically.

\subsection{Annotation Guidelines} \label{sec:guidelines}
We developed our annotation guidelines iteratively over a period of about three months. The first version of the guidelines, created by four senior team members in consultation with a wider pool of participants, introduced the main objectives of the project, the topical areas from which the data will be created, and several categories of instruction types specific to each country (e.g., various types of open-ended requests and questions) illustrated with rich sets of examples. We also included samples from trustworthy \textbf{information sources} (see Appendix \ref{apndx:guidelines}) that can be treated as references while creating the data. This initial version of the guidelines was then shared with the team members who were asked to build a pilot dataset based on these guidelines. After a series of meetings, we further improved the guidelines, reaching an extensive version totaling $100$ pages. This final version was then used to train all team members in order to ensure consistency across all aspects of the project.

In our guidelines, we asked the annotators to create instructions for two main categories: \textit{\textbf{general category}}, which covers MSA instructions for general knowledge such as \textit{science} and \textit{technology}; and \textit{\textbf{country-specific category}}, where the annotators provide instructions reflecting their country's culture in multiple topics as mentioned earlier, including  \textit{local celebrations}, \textit{customs}, \textit{local geography, national history, proverbs,} and \textit{food}. The country-specific instructions could be expressed in either MSA or the dialect corresponding to the respective country. For more details about our annotation guidelines, refer to Appendix~\ref{apndx:guidelines}.\footnote{Our full annotation guidelines manual is available at \href{https://github.com/UBC-NLP/palm/blob/main/guidelines.md}{this link}.}

\subsection{Platform and Quality Assurance} We used Label Studio~\cite{tkachenko2020label} as our annotation platform, forming country-specific sub-teams. Each annotator accessed their respective country’s sections and created instructions in relevant topical areas, following our carefully designed annotation guidelines. We also implemented a structured review process to ensure data quality and annotation consistency. Weekly meetings addressed annotation accuracy, source reliability, instruction diversity, and progress. A dedicated Slack channel enabled real-time collaboration. Figure~\ref{fig:pipeline} illustrates the pipeline for constructing {\ourdataset}, from data collection to final instruction revision. After completing the dataset, we conducted a comprehensive review in which team members cross-reviewed each other’s contributions, ensuring each sample was examined by at least two reviewers. Section~\ref{apndx:revision_porcess} in the appendix analyzes the impact of this revision process and highlights the resulting improvements in data quality.
\subsection{Dataset Analysis}

Unlike other Arabic instruction datasets listed in Table \ref{tab:arabic-datasets}, {\ourdataset} offers several unique features: it covers \textit{all} $22$ Arab countries, a larger number of dialects, and is entirely created and reviewed by humans.
Totaling $17,411$ instruction pairs,~\ourdataset is meticulously designed to reflect the cultural and linguistic richness of Arab countries. In order to further characterize the dataset and showcase the various types of instructions developed, we provide a detailed quantitative analysis here.

\paragraph{Overall Statistics.} 
To facilitate analysis, we divide the countries in {\ourdataset} into two categories: \textit{high-resource} (more than $500$ instructions) and \textit{low-resource} (around $100$ instructions). High-resource countries include Egypt, Jordan, Mauritania, Morocco, Palestine, Saudi Arabia, Sudan, Syria, Tunisia, UAE, and Yemen. We also incorporate a \textit{General} category containing $1,109$ instructions that are not tied to a single country but rather pertain to the Arab world at large. In total, the high-resource countries and the General category account for $16,066$ instructions, representing roughly $92\%$ of the dataset.

A key feature of our dataset is its inclusion of dialects. For the ten out of the eleven high-resource countries, we asked annotators to provide around $380$ examples in their respective local dialects, resulting in $4,211$ dialectal instruction pairs. This addition substantially broadens the dataset’s linguistic diversity and enhances its cultural authenticity. Details of the dialectal distribution are presented in Table \ref{tab:country_data_summary} in Appendix~\ref{apndx:statistics}.

\begin{figure}

    \centering
    \includegraphics[width=1\linewidth]{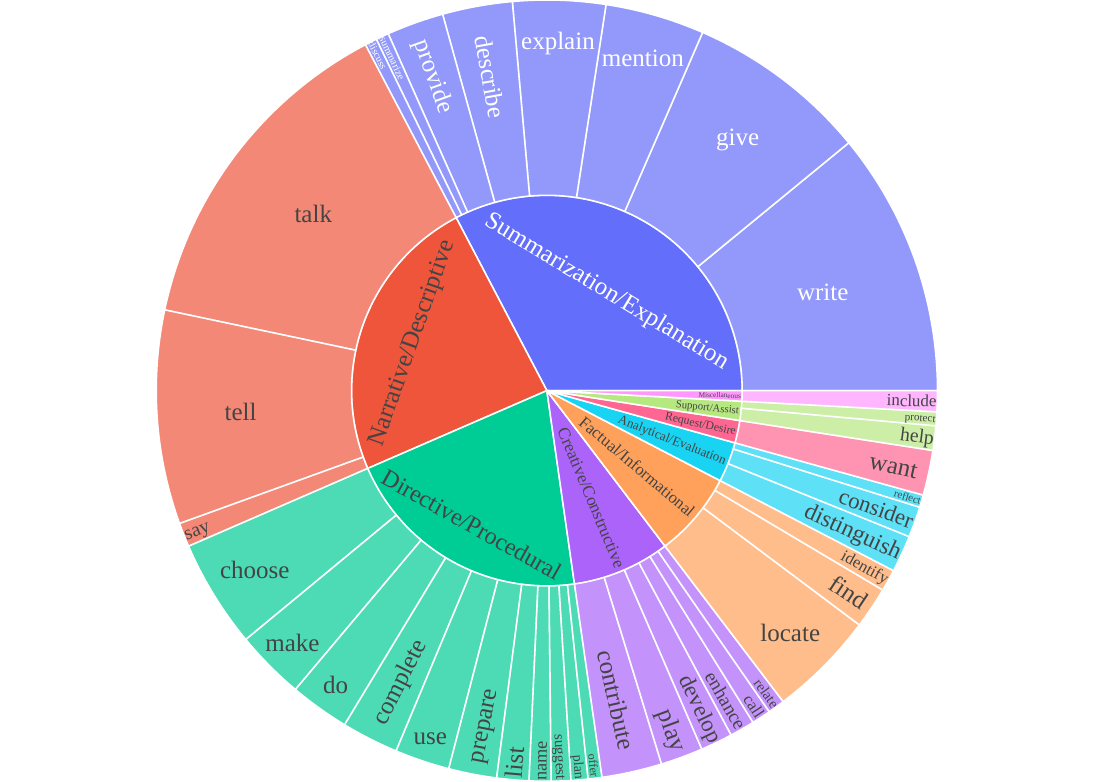}
    \caption{Distribution of instruction types based on verb usage in \ourdataset, illustrating the frequency and categorization of verb–noun pairs and offering insights into the dataset’s instructional diversity. Appendix~\ref{fig:all_sunbursts} presents a drill-down of this analysis. }
     \label{fig:instruction_distribution}
\end{figure}

In contrast, the low-resource countries—Algeria, Bahrain, Comoros, Djibouti, Iraq, Kuwait, Lebanon, Libya, Oman, Qatar, and Somalia—have fewer instructions, largely due to limited annotator availability and resources. Collectively, these countries contribute $1,345$ instructions, amounting to approximately $8\%$ of the dataset.
%%%%%%%%%%%%%%%%%%%%%%%%%%%%%%%%
\paragraph{Instruction Types.}
Instructions in~\ourdataset span a range of diverse categories. During our meetings, we frequently discussed ways to introduce richer components such as role-playing and reasoning, in addition to various information requests and question prompts. One method for characterizing the data after its creation is to extract the \textit{verbs} and subsequent \textit{nouns} from each instruction, allowing us to identify examples such as "\textit{summarize} the following \textit{article}". To implement this, we first translate all instructions into English and then use \texttt{GPT4-o1} to extract the relevant verbs and nouns. We retain only those verbs appearing at least $25$ times and nouns that co-occur with each verb at least five times. We next employ \texttt{GPT-o1} again to cluster these verb-noun pairs according to different instruction types and assign a representative name to each cluster, yielding the following: \textit{(1) Summarization and Explanation} (requests for summaries or definitions), (2) \textit{Directive and Procedural} (guidance on specific actions), \textit{(3) Factual and Informational Queries} (requests for factual information), \textit{(4) Creative and Constructive Generation} (content creation tasks), \textit{(5) Analytical and Evaluation-Based} (critical thinking tasks), and \textit{(6) Narrative and Descriptive Tasks} (storytelling or descriptive content). Figure~\ref{fig:instruction_distribution} depicts the distribution of these instruction types in \textit{Palm}. For a more detailed breakdown of each theme and its corresponding verbs and nouns, refer to Figures~\ref{fig:all_sunbursts} in Appendix~\ref{apndx:instruction_types}.

\paragraph{Dataset Splits.} The {\ourdataset} dataset is organized into three distinct splits: training, public test, and private test sets, each designed to serve a specific purpose. \textbf{\textit{Training Set:}} This set, comprising $13,559$ instructions, will be publicly released as training data for models designed to achieve cultural and linguistic alignment with Arabic communities. It provides researchers with a valuable resource for developing models that are more attuned to Arabic cultural contexts. \textbf{\textit{Public Test:}} This set, hereinafter referred to as the \textit{test} set, will be publicly released as a benchmarking dataset. It enables researchers to evaluate their models’ performance on culturally-specific instructions from the Arab world during the development phase. The test set comprises $1,926$ instructions. \textbf{\textit{Private Test:}} This set will remain private and be accessible exclusively through a leaderboard, ensuring a fair comparison of different models and approaches by preventing data leakage. As with the public test set, it consists of $1,926$ instructions.

\section{Evaluation}
\subsection{Evaluation Setup}
To demonstrate the efficacy of {\ourdataset}, we employ the test set as an evaluation benchmark. This evaluation serves dual purposes: (i) assessing current LLMs' performance across individual countries, topics, and dialects to provide a nuanced measure of Arabic cultural awareness, and (ii) offering a robust methodology for future researchers to evaluate model proficiency in handling Arabic culture and dialects. We evaluate $18$ Arabic-aware LLMs\footnote{The full list of models is in Table~\ref{tab:bench_llms} in Appendix.}, including \texttt{GPT-4o}, \texttt{Claude 3.5 Sonnet}, \texttt{Command R+ (CMDR+)}, \texttt{QWEN 2.5}, \texttt{JAIS}, \texttt{AceGPT}, and \texttt{LLaMA 3.1}, by generating responses \textit{(with greedy decoding)} for the test set of {\ourdataset} instructions and assessing the outputs.\footnote{Models such as Fanar \cite{fanarteam2025fanararabiccentricmultimodalgenerative} and Allam \cite{bari2024allamlargelanguagemodels} are reported to provide encouraging performance but were not available for evaluation at the time of submission.}
Evaluation metrics can be categorized into \textit{surface-level} and \textit{LLM-as-Judge} metrics \cite{zheng2023judgingllmasajudgemtbenchchatbot}.

\subsection{Surface-Level Evaluation}
Following \citet{arora2024calmqaexploringculturallyspecific}, we employ surface-level attributes to automatically evaluate the generated answers. While these metrics do not assess the \textit{correctness} of responses, they enable us to measure three key aspects: \textit{(i) language consistency} between the instruction and the generated answer; \textit{(ii) preservation of the prompt’s dialect (dialectal consistency)} in the generated answer; and \textit{(iii) presence of sequence repetitions} within the generated answer. This approach allows us to examine the model’s capacity to maintain linguistic and cultural fidelity without directly assessing factual accuracy.

\subsection{LLM-as-Judge Evaluation}
For the LLM-as-Judge assessment \cite{zheng2023judgingllmasajudgemtbenchchatbot}, we focus on the \textit{correctness} metric. We select three models with strong performance in Arabic tasks—\texttt{GPT-4o}, \texttt{CMDR+}, and \texttt{QWEN $2.5$ $72$B}—and prompt them to rate answer correctness on a scale of 1 to 10, using LangChain’s built-in evaluation pipeline.\footnote{\url{https://langchain.com/}} Each rating considers both the instruction and the ground truth, as shown in Figure~\ref{fig:prompt}. We then compute a mean correctness score for each generated response, capturing factual accuracy relative to the provided ground truth and ensuring no major errors. To gauge the reliability of this method, we calculate the Intraclass Correlation Coefficient (ICC) among the three models, yielding an \textbf{ICC of 0.68}, which indicates good scoring consistency. Table~\ref{tab:automatic_data_stats_pre_country} summarizes the \textbf{1,926} samples in the test set used for this evaluation.\footnote{We also apply the same methodology to evaluate \textit{coherence}, \textit{detail}, and \textit{helpfulness}, with descriptions provided in Appendix~\ref{apndx:evaluation_prompt}.}

\subsection{Human Evaluation}
\begin{figure*}[h!]
    \centering
    \begin{subfigure}[b]{0.42\textwidth}
        \centering
        \includegraphics[width=\textwidth]{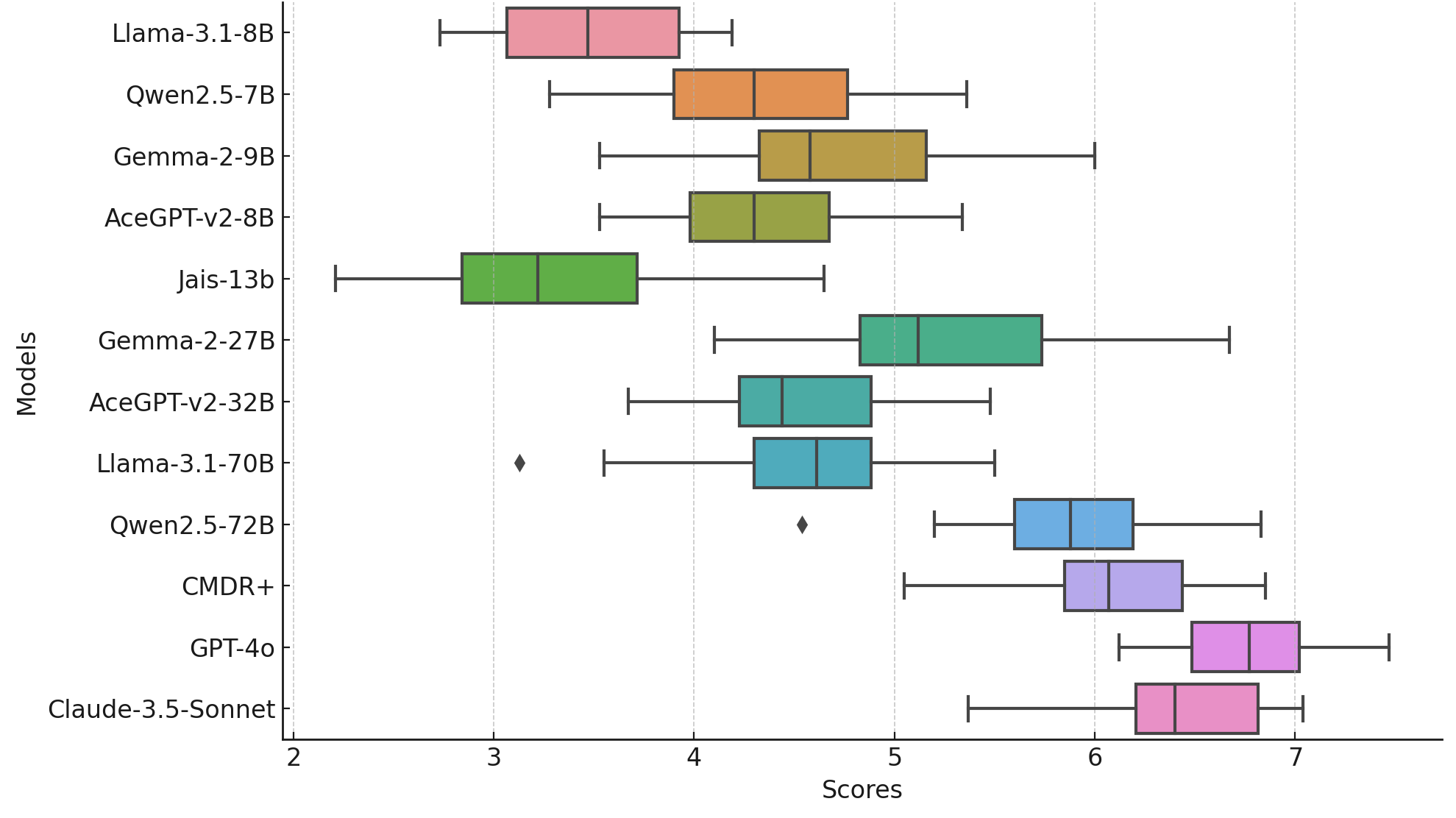}
        \caption{ }
        \label{fig:Average_correctness_country_boxplot}
    \end{subfigure}
    \hfill
    \begin{subfigure}[b]{0.42\textwidth}
        \centering
        \includegraphics[width=\textwidth]{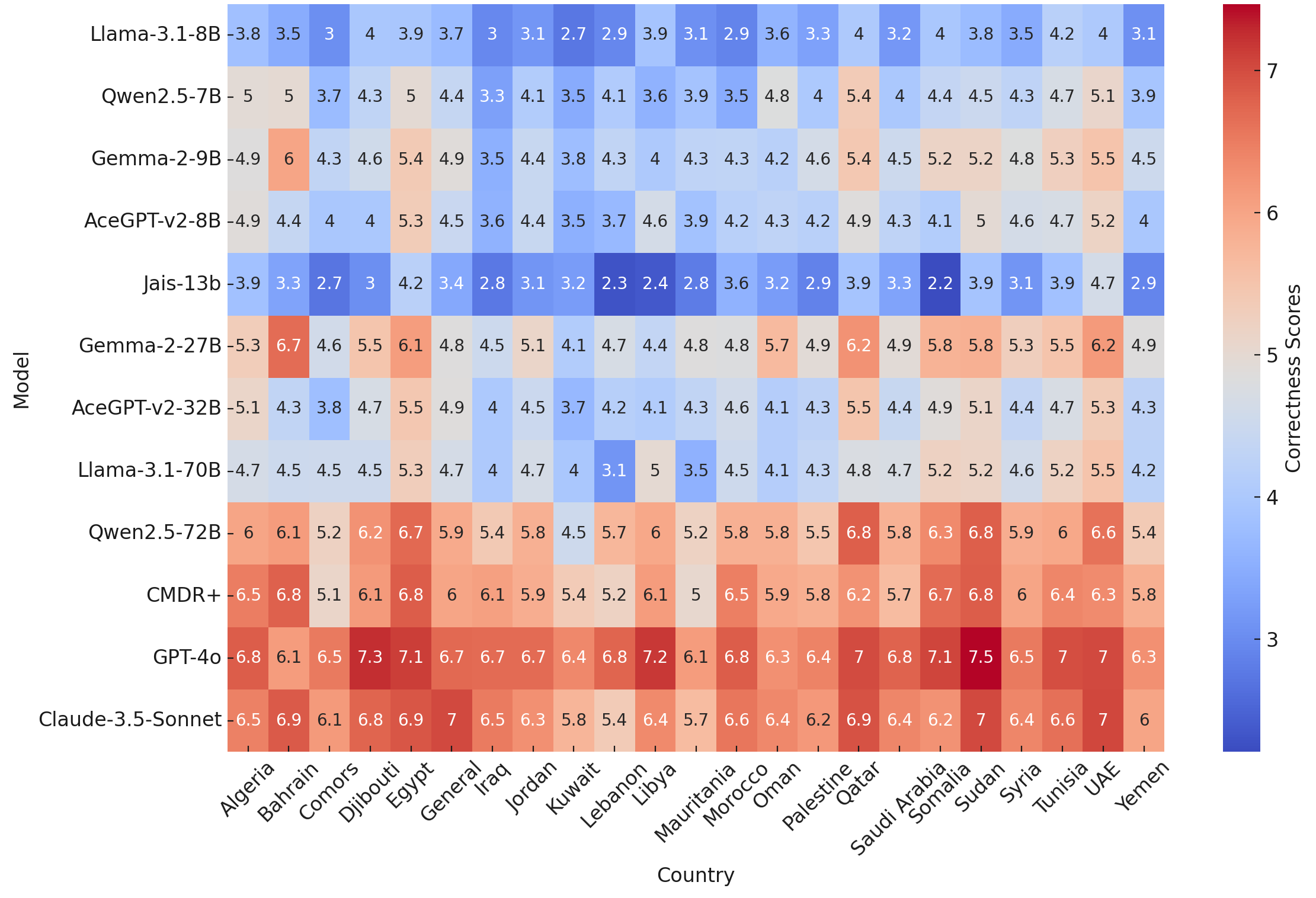}
        \caption{}
        \label{fig:Average_correctness_country_heatmap}
    \end{subfigure}
    \caption{LLM-as-judge average correctness scores across Arabic countries.}
    \label{fig:combined}
\end{figure*}

To validate our automatic evaluation, we conducted a human evaluation on a subset of generated responses, encompassing diverse examples from multiple countries, topics, and both MSA and dialects. We enlisted evaluators from the same data creation team, representing various Arab countries. These evaluators assessed responses from five LLMs based solely on the \textit{correctness} criterion. Each response was evaluated by at least three independent human evaluators. We then computed the intraclass correlation coefficient (ICC) across all evaluators, yielding an \textbf{average ICC of 0.67}, indicating high inter-rater agreement.

The evaluation subset was sampled from the same test set used in the LLM-as-Judge evaluation and included \textbf{92 MSA samples} plus \textbf{20 dialectal samples} each for Egypt, Morocco, Syria, and Yemen. We further computed the correlation between the human evaluation scores and the LLM-as-Judge \textit{correctness} scores, observing a strong relationship. The \textbf{Pearson correlation coefficient is 0.76 (p-value < 0.05)} and the \textbf{ICC is 0.78}, lending strong credibility to the LLM-as-Judge results presented in the following section.

\section{Results and Discussion}

\begin{figure}[]
  \centering
  \includegraphics[width=\columnwidth]{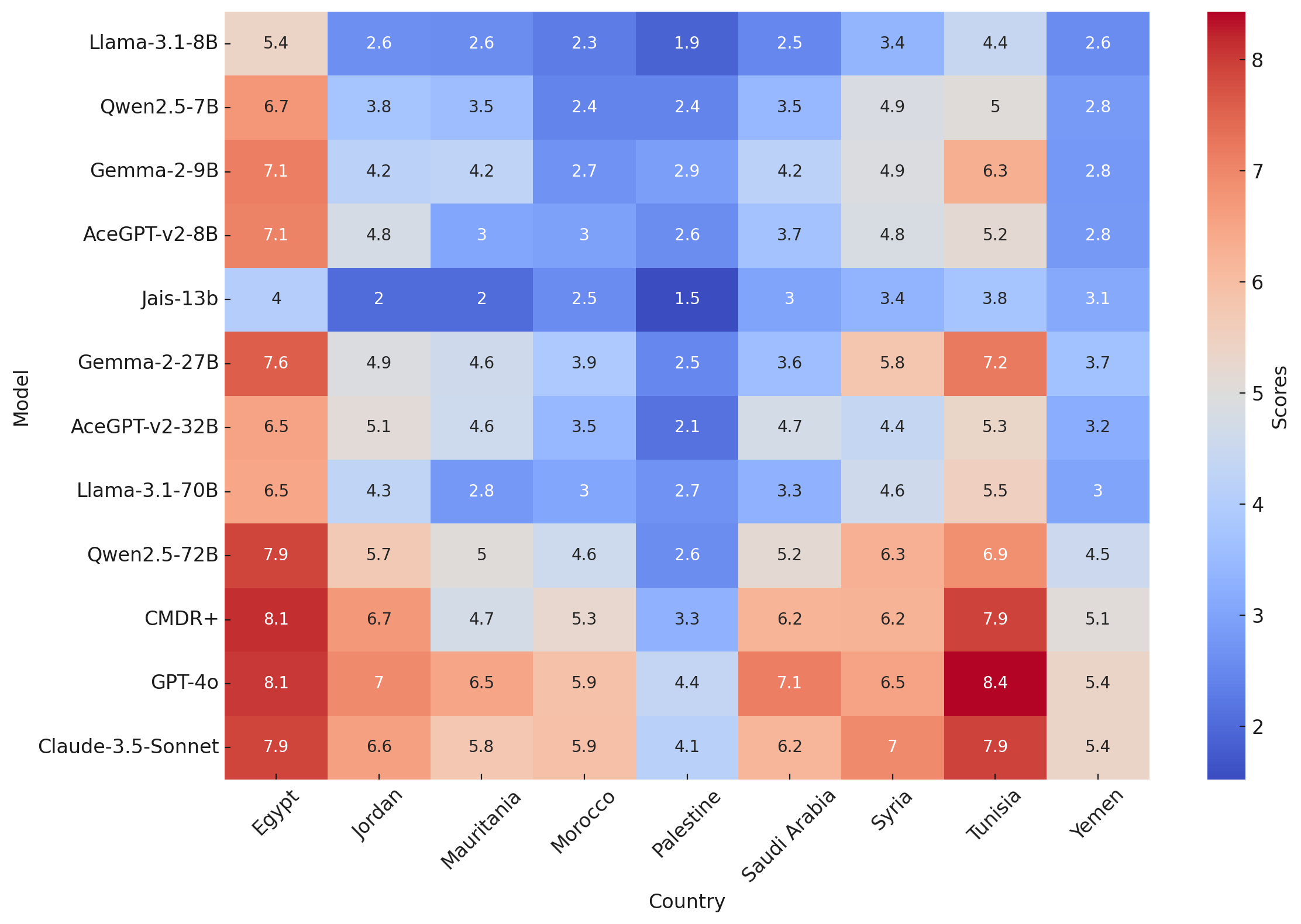}
  \caption{Correctness scores across Arabic dialects.
  }   
  \label{fig:avg_correctness_country_da}
\end{figure}

Here, we analyze results of a subset of 12 LLMs that yield notable insights.\footnote{Complete results are available in Appendix \ref{append:llm_judge_res}, with Section \ref{append:small_model_ablation} showing that smaller models generally underperform in providing correct answers to the Arabic cultural questions.}

\subsection{Surface-Level Results}Table~\ref{tab:surface_attributes} shows the results for the surface-level attributes, namely repetitions, language consistency, and dialectal consistency. The top-performing models (e.g., \texttt{Qwen 2.5}, \texttt{GPT-4o}, and \texttt{Claude-3.5-Sonnet}) display negligible repetition rates when prompted with Arabic instructions, whereas \texttt{Gemma-2} and \texttt{Llama 3.1} exhibit higher repetition rates. For instance, $42\%$ of responses generated by \texttt{Gemma-2-9b} contain repetitions. Regarding language consistency, all models follow a similar trend. However, dialectal consistency remains below roughly $10\%$ across all LLMs, indicating that even when these Arabic-aware models are prompted with dialectal Arabic instructions, they tend to produce answers in MSA about $90\%$ of the time.

\begin{table}[ht]
\centering
\resizebox{.36\textwidth}{!}{
\begin{tabular}{lcccc}
\toprule
\textbf{Model}             & \textbf{A1} & \textbf{A2} & \textbf{A3} \\
\midrule
AceGPT-v2-32B     & 0.36             & 90.45           & 2.78            \\
AceGPT-v2-8B      & 0.57             & 90.71           & 3.33            \\
Llama-3.1-70B     & 2.75             & 91.17           & 10.56           \\
Llama-3.1-8B      & 8.20             & 91.28           & 10.56           \\
Qwen2.5-72B       & 0.00             & 91.33           & 8.33            \\
Qwen2.5-7B        & 0.93             & 90.81           & 3.89            \\
CMDR+        & 1.04             & 90.81           & 6.67            \\
gemma-2-27b       & 30.22            & 90.29           & 7.78            \\
gemma-2-9b        & 42.47            & 90.76           & 6.11            \\
GPT-4o            & 0.00             & 91.07           & 8.33            \\
jais-13b          & 0.73             & 90.71           & 2.22            \\
Claude-3.5-Sonnet & 0.00             & 91.02           & 9.44      \\
\bottomrule
\end{tabular}}
\caption{Results of surface-level attributes (\%). \textbf{A1:} repetitions, \textbf{A2:} lang consistency, \textbf{A3:} dialect consistency.}
\label{tab:surface_attributes}
\end{table}
\subsection{LLM-as-Judge Results}

\textbf{Overall Results.} Figure~\ref{fig:Average_correctness_country_boxplot} presents the results of our \textit{correctness} evaluation using LLM-as-Judge. Overall, \textit{GPT-4o and Claude-3.5-Sonnet exhibit the highest performance}, achieving median scores above $6.0$. These larger models are followed closely by \texttt{CMDR+} and \texttt{Qwen2.5-72B}, which display robust performance with median scores ranging from $5.8$ to $6.0$. \textit{Mid-range models}—including \texttt{AceGPT-v2-32B}, \texttt{Gemma-2-27B}, and \texttt{Llama-3.1-70B}—cluster around median scores of $4.5$ to $5.0$, indicating moderate competence in Arabic evaluation. Conversely, \textit{lower-capacity models} such as \texttt{Jais-13b} and \texttt{Llama-3.1-8B} demonstrate comparatively weaker performance, with median scores between $3.0$ and $4.0$. Notably, the boxplot whiskers suggest considerable variance across all models, indicating that performance can be inconsistent depending on the specific test cases. An interesting, albeit unsurprising, finding is the overall correlation between model size and performance, with larger models (>70B parameters) consistently outperforming their smaller counterparts. We report the results for the \textit{coherence}, \textit{detail}, and \textit{helpfulness} metrics in Figure~\ref{fig:avg_four_metrics}, located in Appendix~\ref{append:llm_judge_res}.

\paragraph{Per-Country Results.} Figure~\ref{fig:Average_correctness_country_heatmap} presents the \textit{correctness} scores by country. The figure shows that more advanced models like \texttt{GPT-4o} and \texttt{Claude-3.5-Sonnet} consistently exhibit higher scores—often exceeding $6.0$—indicating better cultural understanding. For example, \texttt{GPT-4o} achieves scores of $7.5$ for Syria and $7.3$ for Djibouti. In contrast, earlier or smaller models, such as \texttt{Llama-3.1-8B} and \texttt{Jais-13b}, display lower \textit{correctness} scores, frequently below $4.0$, with \texttt{Jais-13b} scoring as low as $2.2$ for Saudi Arabia. The variation in scores across different countries underscores each model’s varying degree of cultural competence, highlighting the complexity of capturing diverse cultural nuances within the Arabic-speaking world. For instance, \texttt{Claude-3.5-Sonnet} scores $7.0$ for Yemen but $5.8$ for Lebanon, suggesting a stronger grasp of Yemeni contexts relative to Lebanese ones.

\paragraph{Per-Dialect Results.}
Figure~\ref{fig:avg_correctness_country_da} shows the \textit{correctness} scores of various models when prompted with dialectal instructions across multiple Arabic dialects. Larger models, such as \texttt{GPT-4o}, \texttt{Claude-3.5-Sonnet}, and \texttt{CMDR+}, consistently exhibit higher scores—frequently above $7.0$—in most countries. For example, \texttt{GPT-4o} achieves $8.1$ for Egypt and $8.4$ for Tunisia. In contrast, smaller models (e.g., \texttt{Llama-3.1-8B} and \texttt{Jais-13b}) generally perform below $4.0$. Performance also varies by country, with some dialects presenting more difficulty than others. Morocco and Palestine often yield lower scores for multiple models, whereas Egypt and Tunisia tend to yield higher ones.\footnote{Results per topic are presented in Appendix~\ref{append:llm_judge_res}. LLMs demonstrated higher performance in topics such as celebrations, history, and travel, while achieving lower scores in categories like sports and food.}

\subsection{Human Evaluation Results}

Table \ref{tab:human_eval_dialect} presents the average human-evaluation \textit{correctness} scores for dialectal Arabic instructions in four Arab countries. \texttt{AceGPT-v2-32B} consistently performed well, achieving the highest scores for Egypt (6.47) and Morocco (4.55). \texttt{Claude-3-5-Sonnet} showed strong performance for Morocco (6.23) and Syria (4.65). \texttt{Llama-3.1-8B} and \texttt{Qwen2.5-72B} had middle-point results, while \texttt{Jais-13b} generally scored lowest except for Yemen. Notably, scores varied substantially between countries for each model, suggesting that \textit{performance on dialectal Arabic instructions is highly dependent on the specific country and dialect being evaluated}.

\begin{table}[h]
\centering
\resizebox{.38\textwidth}{!}{
\begin{tabular}{lccccc}
\toprule
\textbf{Country} & \textbf{M1} & \textbf{M2} & \textbf{M3} & \textbf{M4} & \textbf{M5} \\ \midrule
Egypt   & 6.47 & 4.26 & 4.71 & 4.15 & 4.08 \\ 
Moroc & 4.55 & 2.87 & 4.44 & 6.23 & 3.10 \\ 
Syria   & 3.27 & 3.40 & 4.03 & 4.65 & 2.27 \\ 
Yemen   & 2.13 & 1.85 & 2.58 & 4.28 & 2.90 \\ \bottomrule
\end{tabular}}
\caption{Avg human eval correctness for dialect instructions per country. \textbf{M1:} AceGPT-v2-32B, \textbf{M2:} Llama-3.1-8B, \textbf{M3:} Qwen2.5-72B, \textbf{M4:} Claude-3-5-Sonnet, \textbf{M5:} Jais-13b.}
\label{tab:human_eval_dialect}
\end{table}

A similar trend is noticed for Table~\ref{tab:human_eval_msa}, where \texttt{Claude-3-5-Sonnet} achieves the highest score for several countries, notably Syria ($7.25$) and Egypt ($6.83$). \texttt{AceGPT-v2-32B} and \texttt{Qwen 2.5-72B} also perform well in multiple regions. Performance varies significantly by country.

\section{Conclusion}
In this work, we introduced \ourdataset, a culturally inclusive and linguistically diverse dataset that covers all $22$ Arab countries. \ourdataset is designed to enhance the cultural capabilities and facilitate benchmarking of Arabic LLMs. Through a year-long, community-driven effort involving $44$ researchers from across $15$ different Arab countries, \ourdataset offers a comprehensive set of instructions that cover both MSA and various regional dialects. Our evaluations using \ourdataset demonstrate the importance of culturally tailored datasets in assessing LLMs, highlighting the gaps in existing models when it comes to understanding and generating culturally relevant and dialect-specific responses. \ourdataset not only improves the representation of diverse Arab cultures in technology but also provides a benchmark for future work in culturally sensitive and inclusive NLP. By making \ourdataset publicly available, we aim to foster continued research and development in the field, ultimately contributing to the creation of more culturally aware language technologies.

\section*{Limitations}

While \ourdataset serves as a valuable resource for training and benchmarking culturally aware, dialectally diverse Arabic LLMs, it has some limitations. In low-resource countries, content was often contributed by annotators from neighboring countries rather than local speakers. Although these annotators share certain cultural similarities, the resulting instructions may lack the depth and nuances that native, local speakers would bring.

Moreover, many Arab countries have multiple regional dialects, which require larger, geographically diverse teams to fully represent. Due to the scale of the project, some dialectal variations may not be covered in detail, limiting the dataset’s ability to capture every linguistic nuance.

Lastly, although automatic evaluations using LLMs facilitate scalable assessment, they can fall short when dealing with dialects and subtle cultural elements. These models may misjudge culturally specific or dialectal content, introducing biases. Consequently, human evaluation remains essential alongside automated methods to ensure reliable results.

\section*{Ethics Statement}
In developing \ourdataset, we emphasized cultural sensitivity, inclusivity, and ethical responsibility. All annotations were created by informed participants, each of whom is a co-author on this paper, ensuring that every contributor receives full credit for their work. We adhered strictly to publicly available and reputable sources, refraining from using any private or sensitive data. Clear guidelines were also provided to respect local norms, maintain data privacy, and secure participant consent.

Although \ourdataset aims to mitigate biases in Arabic LLMs, unintentional cultural bias may still occur—particularly in regions lacking direct local representation. We encourage ongoing community involvement to address these gaps, ensuring continual refinement and improvement of the dataset.

%\noindent
\textbf{Reproducibility.}
 Our test data, prompts, and code necessary to produce all results reported in this work are publicly available at  \href{https://github.com/UBC-NLP/palm}{\ourdataset}. Our private test set will also be available through a leaderboard.

\section*{Acknowledgments}\label{sec:acknow}
We acknowledge support from Canada Research Chairs (CRC), the Natural Sciences and Engineering Research Council of Canada (NSERC; RGPIN-2018-04267), the Social Sciences and Humanities Research Council of Canada (SSHRC; 895-2020-1004; 895-2021-1008), Canadian Foundation for Innovation (CFI; 37771), Digital Research Alliance of Canada,\footnote{\href{https://alliancecan.ca}{https://alliancecan.ca}} and UBC Advanced Research Computing-Sockeye.\footnote{\href{https://arc.ubc.ca/ubc-arc-sockeye}{https://arc.ubc.ca/ubc-arc-sockeye}}

\bibliography{custom}

\begin{thebibliography}{40}
\providecommand{\natexlab}[1]{#1}

\bibitem[{Abdul-Mageed et~al.(2021)Abdul-Mageed, Elmadany, and Nagoudi}]{abdul-mageed-etal-2021-arbert}
Muhammad Abdul-Mageed, AbdelRahim Elmadany, and El~Moatez~Billah Nagoudi. 2021.
\newblock \href {https://doi.org/10.18653/v1/2021.acl-long.551} {{ARBERT} {\&} {MARBERT}: Deep bidirectional transformers for {A}rabic}.
\newblock In \emph{Proceedings of the 59th Annual Meeting of the Association for Computational Linguistics and the 11th International Joint Conference on Natural Language Processing (Volume 1: Long Papers)}, pages 7088--7105, Online. Association for Computational Linguistics.

\bibitem[{Abdul-Mageed et~al.(2024)Abdul-Mageed, Keleg, Elmadany, Zhang, Hamed, Magdy, Bouamor, and Habash}]{abdul-mageed-etal-2024-nadi}
Muhammad Abdul-Mageed, Amr Keleg, AbdelRahim Elmadany, Chiyu Zhang, Injy Hamed, Walid Magdy, Houda Bouamor, and Nizar Habash. 2024.
\newblock \href {https://doi.org/10.18653/v1/2024.arabicnlp-1.79} {{NADI} 2024: The fifth nuanced {A}rabic dialect identification shared task}.
\newblock In \emph{Proceedings of The Second Arabic Natural Language Processing Conference}, pages 709--728, Bangkok, Thailand. Association for Computational Linguistics.

\bibitem[{Adilazuarda et~al.(2024)Adilazuarda, Mukherjee, Lavania, Singh, Aji, O'Neill, Modi, and Choudhury}]{adilazuarda2024measuringmodelingculturellms}
Muhammad~Farid Adilazuarda, Sagnik Mukherjee, Pradhyumna Lavania, Siddhant Singh, Alham~Fikri Aji, Jacki O'Neill, Ashutosh Modi, and Monojit Choudhury. 2024.
\newblock \href {https://arxiv.org/abs/2403.15412} {Towards measuring and modeling "culture" in llms: A survey}.
\newblock \emph{Preprint}, arXiv:2403.15412.

\bibitem[{AlKhamissi et~al.(2024)AlKhamissi, ElNokrashy, Alkhamissi, and Diab}]{alkhamissi-etal-2024-investigating}
Badr AlKhamissi, Muhammad ElNokrashy, Mai Alkhamissi, and Mona Diab. 2024.
\newblock \href {https://aclanthology.org/2024.acl-long.671} {Investigating cultural alignment of large language models}.
\newblock In \emph{Proceedings of the 62nd Annual Meeting of the Association for Computational Linguistics (Volume 1: Long Papers)}, pages 12404--12422, Bangkok, Thailand. Association for Computational Linguistics.

\bibitem[{Alwajih et~al.(2025)Alwajih, Magdy, Mekki, Nacar, Nafea, Abdelfadil, Yahya, Luqman, Almarwani, Aloufi, Qawasmeh, Atou, Sibaee, Alsayadi, Al-Dhabyani, Al-shaibani, aatar, Qandos, Alhamouri, Ahmad, Khassib, Hamad, AL-Ghrawi, Alshamari, Malainine, Qawasmeh, Yacoub, moilid, AbuHweidi, Aboeitta, Lemin, Abdel-Salam, Bashiti, Ammar, Alansari, Ashraf, Alturayeif, Shatnawi, Inciarte, Elmadany, cheikh tourad, Berrada, Jarrar, Shehata, and Abdul-Mageed}]{alwajih2025pearlmultimodalculturallyawarearabic}
Fakhraddin Alwajih, Samar~Mohamed Magdy, Abdellah~El Mekki, Omer Nacar, Youssef Nafea, Safaa~Taher Abdelfadil, Abdulfattah~Mohammed Yahya, Hamzah Luqman, Nada Almarwani, Samah Aloufi, Baraah Qawasmeh, Houdaifa Atou, Serry Sibaee, Hamzah~A. Alsayadi, Walid Al-Dhabyani, Maged~S. Al-shaibani, Aya~El aatar, Nour Qandos, Rahaf Alhamouri, Samar Ahmad, Razan Khassib, Lina Hamad, Mohammed~Anwar AL-Ghrawi, Fatimah Alshamari, Cheikh Malainine, Doaa Qawasmeh, Aminetou Yacoub, Tfeil moilid, Ruwa AbuHweidi, Ahmed Aboeitta, Vatimetou~Mohamed Lemin, Reem Abdel-Salam, Ahlam Bashiti, Adel Ammar, Aisha Alansari, Ahmed Ashraf, Nora Alturayeif, Sara Shatnawi, Alcides~Alcoba Inciarte, AbdelRahim~A. Elmadany, Mohamedou cheikh tourad, Ismail Berrada, Mustafa Jarrar, Shady Shehata, and Muhammad Abdul-Mageed. 2025.
\newblock \href {https://arxiv.org/abs/2505.21979} {Pearl: A multimodal culturally-aware arabic instruction dataset}.
\newblock \emph{Preprint}, arXiv:2505.21979.

\bibitem[{Alwajih et~al.(2024)Alwajih, Nagoudi, Bhatia, Mohamed, and Abdul-Mageed}]{alwajih-etal-2024-peacock}
Fakhraddin Alwajih, El~Moatez~Billah Nagoudi, Gagan Bhatia, Abdelrahman Mohamed, and Muhammad Abdul-Mageed. 2024.
\newblock \href {https://aclanthology.org/2024.acl-long.689} {Peacock: A family of {A}rabic multimodal large language models and benchmarks}.
\newblock In \emph{Proceedings of the 62nd Annual Meeting of the Association for Computational Linguistics (Volume 1: Long Papers)}, pages 12753--12776, Bangkok, Thailand. Association for Computational Linguistics.

\bibitem[{Alyafeai et~al.(2024{\natexlab{a}})Alyafeai, Almubarak, Ashraf, Alnuhait, Alshahrani, Abdulrahman, Ahmed, Gawah, Saleh, Ghaleb, Ali, and Al-shaibani}]{alyafeai-etal-2024-cidar}
Zaid Alyafeai, Khalid Almubarak, Ahmed Ashraf, Deema Alnuhait, Saied Alshahrani, Gubran Abdulrahman, Gamil Ahmed, Qais Gawah, Zead Saleh, Mustafa Ghaleb, Yousef Ali, and Maged Al-shaibani. 2024{\natexlab{a}}.
\newblock \href {https://doi.org/10.18653/v1/2024.findings-acl.764} {{CIDAR}: Culturally relevant instruction dataset for {A}rabic}.
\newblock In \emph{Findings of the Association for Computational Linguistics ACL 2024}, pages 12878--12901, Bangkok, Thailand and virtual meeting. Association for Computational Linguistics.

\bibitem[{Alyafeai et~al.(2024{\natexlab{b}})Alyafeai, Almubarak, Ashraf, Alnuhait, Alshahrani, Abdulrahman, Ahmed, Gawah, Saleh, Ghaleb et~al.}]{alyafeai2024cidar}
Zaid Alyafeai, Khalid Almubarak, Ahmed Ashraf, Deema Alnuhait, Saied Alshahrani, Gubran~AQ Abdulrahman, Gamil Ahmed, Qais Gawah, Zead Saleh, Mustafa Ghaleb, et~al. 2024{\natexlab{b}}.
\newblock Cidar: Culturally relevant instruction dataset for arabic.
\newblock \emph{arXiv preprint arXiv:2402.03177}.

\bibitem[{Antoun et~al.(2020)Antoun, Baly, and Hajj}]{antoun-etal-2020-arabert}
Wissam Antoun, Fady Baly, and Hazem Hajj. 2020.
\newblock \href {https://aclanthology.org/2020.osact-1.2} {{A}ra{BERT}: Transformer-based model for {A}rabic language understanding}.
\newblock In \emph{Proceedings of the 4th Workshop on Open-Source Arabic Corpora and Processing Tools, with a Shared Task on Offensive Language Detection}, pages 9--15, Marseille, France. European Language Resource Association.

\bibitem[{Arora et~al.(2024)Arora, Karpinska, Chen, Bhattacharjee, Iyyer, and Choi}]{arora2024calmqaexploringculturallyspecific}
Shane Arora, Marzena Karpinska, Hung-Ting Chen, Ipsita Bhattacharjee, Mohit Iyyer, and Eunsol Choi. 2024.
\newblock \href {https://arxiv.org/abs/2406.17761} {Calmqa: Exploring culturally specific long-form question answering across 23 languages}.
\newblock \emph{Preprint}, arXiv:2406.17761.

\bibitem[{Bari et~al.(2024)Bari, Alnumay, Alzahrani, Alotaibi, Alyahya, AlRashed, Mirza, Alsubaie, Alahmed, Alabduljabbar, Alkhathran, Almushayqih, Alnajim, Alsubaihi, Mansour, Alrubaian, Alammari, Alawami, Al-Thubaity, Abdelali, Kuriakose, Abujabal, Al-Twairesh, Alowisheq, and Khan}]{bari2024allamlargelanguagemodels}
M~Saiful Bari, Yazeed Alnumay, Norah~A. Alzahrani, Nouf~M. Alotaibi, Hisham~A. Alyahya, Sultan AlRashed, Faisal~A. Mirza, Shaykhah~Z. Alsubaie, Hassan~A. Alahmed, Ghadah Alabduljabbar, Raghad Alkhathran, Yousef Almushayqih, Raneem Alnajim, Salman Alsubaihi, Maryam~Al Mansour, Majed Alrubaian, Ali Alammari, Zaki Alawami, Abdulmohsen Al-Thubaity, Ahmed Abdelali, Jeril Kuriakose, Abdalghani Abujabal, Nora Al-Twairesh, Areeb Alowisheq, and Haidar Khan. 2024.
\newblock \href {https://arxiv.org/abs/2407.15390} {Allam: Large language models for arabic and english}.
\newblock \emph{Preprint}, arXiv:2407.15390.

\bibitem[{Billah~Nagoudi et~al.(2023)Billah~Nagoudi, Abdul-Mageed, Elmadany, Inciarte, and Islam~Khondaker}]{billah-nagoudi-etal-2023-jasmine}
El~Moatez Billah~Nagoudi, Muhammad Abdul-Mageed, AbdelRahim Elmadany, Alcides Inciarte, and Md~Tawkat Islam~Khondaker. 2023.
\newblock \href {https://doi.org/10.18653/v1/2023.emnlp-main.1040} {{JASMINE}: {A}rabic {GPT} models for few-shot learning}.
\newblock In \emph{Proceedings of the 2023 Conference on Empirical Methods in Natural Language Processing}, pages 16721--16744, Singapore. Association for Computational Linguistics.

\bibitem[{Demidova et~al.(2024)Demidova, Atwany, Rabih, Sha{'}ban, and Abdul-Mageed}]{demidova-etal-2024-john}
Anastasiia Demidova, Hanin Atwany, Nour Rabih, Sanad Sha{'}ban, and Muhammad Abdul-Mageed. 2024.
\newblock \href {https://doi.org/10.18653/v1/2024.arabicnlp-1.18} {John vs. ahmed: Debate-induced bias in multilingual {LLM}s}.
\newblock In \emph{Proceedings of The Second Arabic Natural Language Processing Conference}, pages 193--209, Bangkok, Thailand. Association for Computational Linguistics.

\bibitem[{{El Mekki} et~al.(2025){El Mekki}, Atou, Nacar, Shehata, and Abdul-Mageed}]{mekki2025nilechatlinguisticallydiverseculturally}
Abdellah {El Mekki}, Houdaifa Atou, Omer Nacar, Shady Shehata, and Muhammad Abdul-Mageed. 2025.
\newblock \href {https://arxiv.org/abs/2505.18383} {Nilechat: Towards linguistically diverse and culturally aware llms for local communities}.
\newblock \emph{Preprint}, arXiv:2505.18383.

\bibitem[{Elmadany et~al.(2023)Elmadany, Nagoudi, and Abdul-Mageed}]{elmadany-etal-2023-orca}
AbdelRahim Elmadany, ElMoatez~Billah Nagoudi, and Muhammad Abdul-Mageed. 2023.
\newblock \href {https://doi.org/10.18653/v1/2023.findings-acl.609} {{ORCA}: A challenging benchmark for {A}rabic language understanding}.
\newblock In \emph{Findings of the Association for Computational Linguistics: ACL 2023}, pages 9559--9586, Toronto, Canada. Association for Computational Linguistics.

\bibitem[{Fung et~al.(2024)Fung, Zhao, Doo, Sun, and Ji}]{fung2024massivelymulticulturalknowledgeacquisition}
Yi~Fung, Ruining Zhao, Jae Doo, Chenkai Sun, and Heng Ji. 2024.
\newblock \href {https://arxiv.org/abs/2402.09369} {Massively multi-cultural knowledge acquisition \& lm benchmarking}.
\newblock \emph{Preprint}, arXiv:2402.09369.

\bibitem[{Huang et~al.(2024)Huang, Yu, Zhu, Sun, Cheng, Song, Chen, Alharthi, An, He, Liu, Zhang, Chen, Li, Wang, Zhang, Sun, Wan, Li, and Xu}]{huang2024acegptlocalizinglargelanguage}
Huang Huang, Fei Yu, Jianqing Zhu, Xuening Sun, Hao Cheng, Dingjie Song, Zhihong Chen, Abdulmohsen Alharthi, Bang An, Juncai He, Ziche Liu, Zhiyi Zhang, Junying Chen, Jianquan Li, Benyou Wang, Lian Zhang, Ruoyu Sun, Xiang Wan, Haizhou Li, and Jinchao Xu. 2024.
\newblock \href {https://arxiv.org/abs/2309.12053} {Acegpt, localizing large language models in arabic}.
\newblock \emph{Preprint}, arXiv:2309.12053.

\bibitem[{Inoue et~al.(2021)Inoue, Alhafni, Baimukan, Bouamor, and Habash}]{inoue-etal-2021-interplay}
Go~Inoue, Bashar Alhafni, Nurpeiis Baimukan, Houda Bouamor, and Nizar Habash. 2021.
\newblock \href {https://aclanthology.org/2021.wanlp-1.10} {The interplay of variant, size, and task type in {A}rabic pre-trained language models}.
\newblock In \emph{Proceedings of the Sixth Arabic Natural Language Processing Workshop}, pages 92--104, Kyiv, Ukraine (Virtual). Association for Computational Linguistics.

\bibitem[{Keleg and Magdy(2023)}]{keleg-magdy-2023-dlama}
Amr Keleg and Walid Magdy. 2023.
\newblock \href {https://doi.org/10.18653/v1/2023.findings-acl.389} {{DLAMA}: A framework for curating culturally diverse facts for probing the knowledge of pretrained language models}.
\newblock In \emph{Findings of the Association for Computational Linguistics: ACL 2023}, pages 6245--6266, Toronto, Canada. Association for Computational Linguistics.

\bibitem[{Mousi et~al.(2024{\natexlab{a}})Mousi, Durrani, Ahmad, Hasan, Hasanain, Kabbani, Dalvi, Chowdhury, and Alam}]{mousi2024aradicebenchmarksdialectalcultural}
Basel Mousi, Nadir Durrani, Fatema Ahmad, Md.~Arid Hasan, Maram Hasanain, Tameem Kabbani, Fahim Dalvi, Shammur~Absar Chowdhury, and Firoj Alam. 2024{\natexlab{a}}.
\newblock \href {https://arxiv.org/abs/2409.11404} {{AraDiCE}: Benchmarks for dialectal and cultural capabilities in llms}.

\bibitem[{Mousi et~al.(2024{\natexlab{b}})Mousi, Durrani, Ahmad, Hasan, Hasanain, Kabbani, Dalvi, Chowdhury, and Alam}]{mousi2024aradice}
Basel Mousi, Nadir Durrani, Fatema Ahmad, Md~Arid Hasan, Maram Hasanain, Tameem Kabbani, Fahim Dalvi, Shammur~Absar Chowdhury, and Firoj Alam. 2024{\natexlab{b}}.
\newblock Aradice: Benchmarks for dialectal and cultural capabilities in llms.
\newblock \emph{arXiv preprint arXiv:2409.11404}.

\bibitem[{Myung et~al.(2024{\natexlab{a}})Myung, Lee, Zhou, Jin, Putri, Antypas, Borkakoty, Kim, Perez-Almendros, Ayele, Gutiérrez-Basulto, Ibáñez-García, Lee, Muhammad, Park, Rzayev, White, Yimam, Pilehvar, Ousidhoum, Camacho-Collados, and Oh}]{myung2024blendbenchmarkllmseveryday}
Junho Myung, Nayeon Lee, Yi~Zhou, Jiho Jin, Rifki~Afina Putri, Dimosthenis Antypas, Hsuvas Borkakoty, Eunsu Kim, Carla Perez-Almendros, Abinew~Ali Ayele, Víctor Gutiérrez-Basulto, Yazmín Ibáñez-García, Hwaran Lee, Shamsuddeen~Hassan Muhammad, Kiwoong Park, Anar~Sabuhi Rzayev, Nina White, Seid~Muhie Yimam, Mohammad~Taher Pilehvar, Nedjma Ousidhoum, Jose Camacho-Collados, and Alice Oh. 2024{\natexlab{a}}.
\newblock \href {https://arxiv.org/abs/2406.09948} {Blend: A benchmark for llms on everyday knowledge in diverse cultures and languages}.
\newblock \emph{Preprint}, arXiv:2406.09948.

\bibitem[{Myung et~al.(2024{\natexlab{b}})Myung, Lee, Zhou, Jin, Putri, Antypas, Borkakoty, Kim, Perez-Almendros, Ayele et~al.}]{myung2024blend}
Junho Myung, Nayeon Lee, Yi~Zhou, Jiho Jin, Rifki~Afina Putri, Dimosthenis Antypas, Hsuvas Borkakoty, Eunsu Kim, Carla Perez-Almendros, Abinew~Ali Ayele, et~al. 2024{\natexlab{b}}.
\newblock Blend: A benchmark for llms on everyday knowledge in diverse cultures and languages.
\newblock \emph{arXiv preprint arXiv:2406.09948}.

\bibitem[{Nagoudi et~al.(2023)Nagoudi, Elmadany, El-Shangiti, and Abdul-Mageed}]{nagoudi-etal-2023-dolphin}
El~Moatez~Billah Nagoudi, AbdelRahim Elmadany, Ahmed El-Shangiti, and Muhammad Abdul-Mageed. 2023.
\newblock \href {https://doi.org/10.18653/v1/2023.findings-emnlp.98} {Dolphin: A challenging and diverse benchmark for {A}rabic {NLG}}.
\newblock In \emph{Findings of the Association for Computational Linguistics: EMNLP 2023}, pages 1404--1422, Singapore. Association for Computational Linguistics.

\bibitem[{Naous et~al.(2024)Naous, Ryan, Ritter, and Xu}]{naous-etal-2024-beer}
Tarek Naous, Michael Ryan, Alan Ritter, and Wei Xu. 2024.
\newblock \href {https://aclanthology.org/2024.acl-long.862} {Having beer after prayer? measuring cultural bias in large language models}.
\newblock In \emph{Proceedings of the 62nd Annual Meeting of the Association for Computational Linguistics (Volume 1: Long Papers)}, pages 16366--16393, Bangkok, Thailand. Association for Computational Linguistics.

\bibitem[{Nguyen et~al.(2023)Nguyen, Razniewski, Varde, and Weikum}]{Nguyen_2023}
Tuan-Phong Nguyen, Simon Razniewski, Aparna Varde, and Gerhard Weikum. 2023.
\newblock \href {https://doi.org/10.1145/3543507.3583535} {Extracting cultural commonsense knowledge at scale}.
\newblock In \emph{Proceedings of the ACM Web Conference 2023}, volume~21 of \emph{WWW ’23}, page 1907–1917. ACM.

\bibitem[{Ouyang et~al.(2022)Ouyang, Wu, Jiang, Almeida, Wainwright, Mishkin, Zhang, Agarwal, Slama, Ray, Schulman, Hilton, Kelton, Miller, Simens, Askell, Welinder, Christiano, Leike, and Lowe}]{ouyang2022traininglanguagemodelsfollow}
Long Ouyang, Jeff Wu, Xu~Jiang, Diogo Almeida, Carroll~L. Wainwright, Pamela Mishkin, Chong Zhang, Sandhini Agarwal, Katarina Slama, Alex Ray, John Schulman, Jacob Hilton, Fraser Kelton, Luke Miller, Maddie Simens, Amanda Askell, Peter Welinder, Paul Christiano, Jan Leike, and Ryan Lowe. 2022.
\newblock \href {https://arxiv.org/abs/2203.02155} {Training language models to follow instructions with human feedback}.
\newblock \emph{Preprint}, arXiv:2203.02155.

\bibitem[{Petroni et~al.(2019)Petroni, Rockt{\"a}schel, Riedel, Lewis, Bakhtin, Wu, and Miller}]{petroni-etal-2019-language}
Fabio Petroni, Tim Rockt{\"a}schel, Sebastian Riedel, Patrick Lewis, Anton Bakhtin, Yuxiang Wu, and Alexander Miller. 2019.
\newblock \href {https://doi.org/10.18653/v1/D19-1250} {Language models as knowledge bases?}
\newblock In \emph{Proceedings of the 2019 Conference on Empirical Methods in Natural Language Processing and the 9th International Joint Conference on Natural Language Processing (EMNLP-IJCNLP)}, pages 2463--2473, Hong Kong, China. Association for Computational Linguistics.

\bibitem[{Radford et~al.(2018)Radford, Narasimhan, Salimans, and Sutskever}]{radford_improving_2018}
Alec Radford, Karthik Narasimhan, Tim Salimans, and Ilya Sutskever. 2018.
\newblock Improving language understanding by generative pre-training.

\bibitem[{Ramezani and Xu(2023)}]{ramezani-xu-2023-knowledge}
Aida Ramezani and Yang Xu. 2023.
\newblock \href {https://doi.org/10.18653/v1/2023.acl-long.26} {Knowledge of cultural moral norms in large language models}.
\newblock In \emph{Proceedings of the 61st Annual Meeting of the Association for Computational Linguistics (Volume 1: Long Papers)}, pages 428--446, Toronto, Canada. Association for Computational Linguistics.

\bibitem[{Seelawi et~al.(2021)Seelawi, Tuffaha, Gzawi, Farhan, Talafha, Badawi, Sober, Al-Dweik, Freihat, and Al-Natsheh}]{seelawi-etal-2021-alue}
Haitham Seelawi, Ibraheem Tuffaha, Mahmoud Gzawi, Wael Farhan, Bashar Talafha, Riham Badawi, Zyad Sober, Oday Al-Dweik, Abed~Alhakim Freihat, and Hussein Al-Natsheh. 2021.
\newblock \href {https://aclanthology.org/2021.wanlp-1.18} {{ALUE}: {A}rabic language understanding evaluation}.
\newblock In \emph{Proceedings of the Sixth Arabic Natural Language Processing Workshop}, pages 173--184, Kyiv, Ukraine (Virtual). Association for Computational Linguistics.

\bibitem[{Sengupta et~al.(2023)Sengupta, Sahu, Jia, Katipomu, Li, Koto, Marshall, Gosal, Liu, Chen, Afzal, Kamboj, Pandit, Pal, Pradhan, Mujahid, Baali, Han, Bsharat, Aji, Shen, Liu, Vassilieva, Hestness, Hock, Feldman, Lee, Jackson, Ren, Nakov, Baldwin, and Xing}]{sengupta2023jaisjaischatarabiccentricfoundation}
Neha Sengupta, Sunil~Kumar Sahu, Bokang Jia, Satheesh Katipomu, Haonan Li, Fajri Koto, William Marshall, Gurpreet Gosal, Cynthia Liu, Zhiming Chen, Osama~Mohammed Afzal, Samta Kamboj, Onkar Pandit, Rahul Pal, Lalit Pradhan, Zain~Muhammad Mujahid, Massa Baali, Xudong Han, Sondos~Mahmoud Bsharat, Alham~Fikri Aji, Zhiqiang Shen, Zhengzhong Liu, Natalia Vassilieva, Joel Hestness, Andy Hock, Andrew Feldman, Jonathan Lee, Andrew Jackson, Hector~Xuguang Ren, Preslav Nakov, Timothy Baldwin, and Eric Xing. 2023.
\newblock \href {https://arxiv.org/abs/2308.16149} {Jais and jais-chat: Arabic-centric foundation and instruction-tuned open generative large language models}.
\newblock \emph{Preprint}, arXiv:2308.16149.

\bibitem[{Shankar et~al.(2017)Shankar, Halpern, Breck, Atwood, Wilson, and Sculley}]{shankar2017classificationrepresentationassessinggeodiversity}
Shreya Shankar, Yoni Halpern, Eric Breck, James Atwood, Jimbo Wilson, and D.~Sculley. 2017.
\newblock \href {https://arxiv.org/abs/1711.08536} {No classification without representation: Assessing geodiversity issues in open data sets for the developing world}.
\newblock \emph{Preprint}, arXiv:1711.08536.

\bibitem[{Shi et~al.(2024)Shi, Li, Zhang, Ziems, yu, Horesh, de~Paula, and Yang}]{shi2024culturebankonlinecommunitydrivenknowledge}
Weiyan Shi, Ryan Li, Yutong Zhang, Caleb Ziems, Chunhua yu, Raya Horesh, Rogério~Abreu de~Paula, and Diyi Yang. 2024.
\newblock \href {https://arxiv.org/abs/2404.15238} {Culturebank: An online community-driven knowledge base towards culturally aware language technologies}.
\newblock \emph{Preprint}, arXiv:2404.15238.

\bibitem[{Singh et~al.(2024{\natexlab{a}})Singh, Vargus, D{'}souza, Karlsson, Mahendiran, Ko, Shandilya, Patel, Mataciunas, O{'}Mahony, Zhang, Hettiarachchi, Wilson, Machado, Moura, Krzemi{\'n}ski, Fadaei, Ergun, Okoh, Alaagib, Mudannayake, Alyafeai, Chien, Ruder, Guthikonda, Alghamdi, Gehrmann, Muennighoff, Bartolo, Kreutzer, {\"U}st{\"u}n, Fadaee, and Hooker}]{singh-etal-2024-aya}
Shivalika Singh, Freddie Vargus, Daniel D{'}souza, B{\"o}rje Karlsson, Abinaya Mahendiran, Wei-Yin Ko, Herumb Shandilya, Jay Patel, Deividas Mataciunas, Laura O{'}Mahony, Mike Zhang, Ramith Hettiarachchi, Joseph Wilson, Marina Machado, Luisa Moura, Dominik Krzemi{\'n}ski, Hakimeh Fadaei, Irem Ergun, Ifeoma Okoh, Aisha Alaagib, Oshan Mudannayake, Zaid Alyafeai, Vu~Chien, Sebastian Ruder, Surya Guthikonda, Emad Alghamdi, Sebastian Gehrmann, Niklas Muennighoff, Max Bartolo, Julia Kreutzer, Ahmet {\"U}st{\"u}n, Marzieh Fadaee, and Sara Hooker. 2024{\natexlab{a}}.
\newblock \href {https://doi.org/10.18653/v1/2024.acl-long.620} {Aya dataset: An open-access collection for multilingual instruction tuning}.
\newblock In \emph{Proceedings of the 62nd Annual Meeting of the Association for Computational Linguistics (Volume 1: Long Papers)}, pages 11521--11567, Bangkok, Thailand. Association for Computational Linguistics.

\bibitem[{Singh et~al.(2024{\natexlab{b}})Singh, Vargus, Dsouza, Karlsson, Mahendiran, Ko, Shandilya, Patel, Mataciunas, OMahony, Zhang, Hettiarachchi, Wilson, Machado, Moura, Krzemiński, Fadaei, Ergün, Okoh, Alaagib, Mudannayake, Alyafeai, Chien, Ruder, Guthikonda, Alghamdi, Gehrmann, Muennighoff, Bartolo, Kreutzer, Üstün, Fadaee, and Hooker}]{singh2024aya}
Shivalika Singh, Freddie Vargus, Daniel Dsouza, Börje~F. Karlsson, Abinaya Mahendiran, Wei-Yin Ko, Herumb Shandilya, Jay Patel, Deividas Mataciunas, Laura OMahony, Mike Zhang, Ramith Hettiarachchi, Joseph Wilson, Marina Machado, Luisa~Souza Moura, Dominik Krzemiński, Hakimeh Fadaei, Irem Ergün, Ifeoma Okoh, Aisha Alaagib, Oshan Mudannayake, Zaid Alyafeai, Vu~Minh Chien, Sebastian Ruder, Surya Guthikonda, Emad~A. Alghamdi, Sebastian Gehrmann, Niklas Muennighoff, Max Bartolo, Julia Kreutzer, Ahmet Üstün, Marzieh Fadaee, and Sara Hooker. 2024{\natexlab{b}}.
\newblock \href {https://arxiv.org/abs/2402.06619} {Aya dataset: An open-access collection for multilingual instruction tuning}.
\newblock \emph{Preprint}, arXiv:2402.06619.

\bibitem[{Team et~al.(2025)Team, Abbas, Ahmad, Alam, Altinisik, Asgari, Boshmaf, Boughorbel, Chawla, Chowdhury, Dalvi, Darwish, Durrani, Elfeky, Elmagarmid, Eltabakh, Fatehkia, Fragkopoulos, Hasanain, Hawasly, Husaini, Jung, Lucas, Magdy, Messaoud, Mohamed, Mohiuddin, Mousi, Mubarak, Musleh, Naeem, Ouzzani, Popovic, Sadeghi, Sencar, Shinoy, Sinan, Zhang, Ali, Kheir, Ma, and Ruan}]{fanarteam2025fanararabiccentricmultimodalgenerative}
Fanar Team, Ummar Abbas, Mohammad~Shahmeer Ahmad, Firoj Alam, Enes Altinisik, Ehsannedin Asgari, Yazan Boshmaf, Sabri Boughorbel, Sanjay Chawla, Shammur Chowdhury, Fahim Dalvi, Kareem Darwish, Nadir Durrani, Mohamed Elfeky, Ahmed Elmagarmid, Mohamed Eltabakh, Masoomali Fatehkia, Anastasios Fragkopoulos, Maram Hasanain, Majd Hawasly, Mus'ab Husaini, Soon-Gyo Jung, Ji~Kim Lucas, Walid Magdy, Safa Messaoud, Abubakr Mohamed, Tasnim Mohiuddin, Basel Mousi, Hamdy Mubarak, Ahmad Musleh, Zan Naeem, Mourad Ouzzani, Dorde Popovic, Amin Sadeghi, Husrev~Taha Sencar, Mohammed Shinoy, Omar Sinan, Yifan Zhang, Ahmed Ali, Yassine~El Kheir, Xiaosong Ma, and Chaoyi Ruan. 2025.
\newblock \href {https://arxiv.org/abs/2501.13944} {Fanar: An arabic-centric multimodal generative ai platform}.
\newblock \emph{Preprint}, arXiv:2501.13944.

\bibitem[{Tkachenko et~al.(2020)Tkachenko, Malyuk, Holmanyuk, and Liubimov}]{tkachenko2020label}
Maxim Tkachenko, Mikhail Malyuk, Andrey Holmanyuk, and Nikolai Liubimov. 2020.
\newblock \href {https://github. com/heartexlabs/label-studio} {Label studio: Data labeling software}.
\newblock \emph{Open source software available from https://github. com/heartexlabs/label-studio}, 2022.

\bibitem[{Xu et~al.(2024)Xu, Hu, Zhao, Qiu, Ye, and Gu}]{xu2024surveymultilinguallargelanguage}
Yuemei Xu, Ling Hu, Jiayi Zhao, Zihan Qiu, Yuqi Ye, and Hanwen Gu. 2024.
\newblock \href {https://arxiv.org/abs/2404.00929} {A survey on multilingual large language models: Corpora, alignment, and bias}.
\newblock \emph{Preprint}, arXiv:2404.00929.

\bibitem[{Zheng et~al.(2023)Zheng, Chiang, Sheng, Zhuang, Wu, Zhuang, Lin, Li, Li, Xing, Zhang, Gonzalez, and Stoica}]{zheng2023judgingllmasajudgemtbenchchatbot}
Lianmin Zheng, Wei-Lin Chiang, Ying Sheng, Siyuan Zhuang, Zhanghao Wu, Yonghao Zhuang, Zi~Lin, Zhuohan Li, Dacheng Li, Eric~P. Xing, Hao Zhang, Joseph~E. Gonzalez, and Ion Stoica. 2023.
\newblock \href {https://arxiv.org/abs/2306.05685} {Judging llm-as-a-judge with mt-bench and chatbot arena}.
\newblock \emph{Preprint}, arXiv:2306.05685.

\end{thebibliography}

\appendix

% #############################################################
\clearpage
\appendixpage
\addappheadtotoc
\numberwithin{figure}{section}
\numberwithin{table}{section}
%%%%%%%%%%%%%%%%%%%%%%%%%%%%%%%%%%%%%%%%%%%%%%%%%%%
% ############################################################

We provide an addition organized as follows:

\begin{itemize}
\item Annotation Guidelines \ref{apndx:guidelines}.
\item Details of Instruction Dataset Topics and Categories \ref{apndx:topics_categories}.
\item Diverse Instruction Formats and Linguistic Variations \ref{apndx:instruction_types}.
\item Statistics \ref{apndx:statistics}.
\item Revision Process \ref{apndx:revision_porcess}.

\item Comparative Analysis of Arabic Instructional Datasets \ref{apndx:comparative_analysis}.
\item Evaluation Prompt \ref{apndx:evaluation_prompt}.

\item Selected Examples \ref{apndx:examples}.

\end{itemize}

\section{Annotation Guidelines}
\label{apndx:guidelines}

The key guidelines include the following criteria:
\begin{itemize}
  \item Use Trustworthy Sources: Annotators are instructed to use only reliable sources, which may include but are not limited to:
Wikipedia,
Online encyclopedias,
Books,
Governmental websites,
and Specialized websites
    \item Maintain Objectivity: For certain topics and domains, annotators are asked to provide answers that are objective and based on factual information and established knowledge.
  \item Avoid Personal Opinions: In domains such as politics and religion, annotators should focus solely on presenting information without incorporating personal beliefs or interpretations.
  \item Encourage Creativity: Annotators are encouraged to be creative by generating a diverse range of instructions across all domains.
\end{itemize}

{\bf Information Sources.} To ensure high data quality, we underscore the importance of consulting reliable and authoritative sources during instruction creation. Annotators were consistently advised to perform thorough verification of these sources. Exemplary sources include Wikipedia and other reputable online encyclopedias, academic books, governmental websites, and specialized platforms (e.g., health organization websites offering medical information). We explicitly cautioned against relying on single-individual sources, such as personal posts or social media content, unless the individual is a widely recognized expert in the respective field. Notably, for domains such as travel, culinary arts, and culturally specific celebrations, the most valuable insights often derive from online discussions and forums. In these cases, we leveraged annotators’ local cultural knowledge and judgment to ensure the trustworthiness and relevance of the data.

Our full annotation guidelines manual is available at \href{https://github.com/UBC-NLP/palm/blob/main/guidelines.md}{https://github.com/UBC-NLP/palm/blob/main/guidelines.md}.

\section{\textit{Palm} Topics and Domains}
\label{apndx:topics_categories}
\noindent \textit{Palm} categorizes its instructions into three main domains: \textbf{General}, \textbf{Hybrid}, and \textbf{Country-Specific}. The \textbf{General Domain} covers topics with universally applicable knowledge, such as science, sports, and technology. The \textbf{Hybrid Domain} consists of topics that include both general and country-specific knowledge, bridging regional cultural insights with broader themes. Lastly, the \textbf{Country-Specific Domain} focuses exclusively on Arab nations, highlighting their traditions, social norms, and linguistic nuances. Table~\ref{tab:domains} provides a breakdown of the key instruction domains within \textit{Palm}.

\begin{table*}[ht]
\centering
\definecolor{headercolor}{HTML}{E0F2E9} % Very light green for header
\definecolor{rowcolor}{HTML}{F2FAF4}   % Very light green for alternating rows   % Very light blue for alternating rows

\renewcommand{\arraystretch}{1.2}      % Adjust row height for readability
\setlength{\tabcolsep}{8pt}           % Adjust column spacing

% Scale the entire table down to 70%
\scalebox{0.8}{%

% Apply alternating row colors starting from the first row below the header
\rowcolors{2}{rowcolor}{white}

\begin{tabular}{p{3cm} p{13cm}}
    % Header row with custom background color
    \rowcolor{headercolor}
    \toprule
    \textbf{Domain} & \textbf{Definition} \\
    \midrule
    
    \textbf{Science} & Covers various scientific fields, including biology, physics, chemistry, mathematics, and astronomy. Instructions range from fundamental concepts to applied sciences and technological advancements. \\
    \textbf{Food} & Covers general knowledge about ingredients, nutrition, and food safety, as well as country-specific dishes, traditional recipes, and meal customs in Arab countries. \\
    \textbf{Sports} & Includes general sports rules and history, as well as country-specific sporting traditions, major tournaments, and notable athletes in the Arab world. \\
    \textbf{Politics} & Covers both general political concepts (e.g., voting systems, ideologies) and country-specific topics like political parties, government structures, and notable leaders. \\
    \textbf{Religion} & Explores the major monotheistic religions (Islam, Christianity, Judaism), focusing on historical sites, religious figures, and institutions while avoiding specific rituals. \\
    \textbf{History} & Encompasses ancient civilizations, historical events, wars, and influential leaders, highlighting their impact on Arab culture and heritage. \\
    \textbf{Travel} & Provides information on notable historical landmarks, best travel destinations, itineraries, and cultural tourism across Arab countries. \\
    \textbf{Flora \& Environment} & Discusses wildlife, national parks, climate change, agricultural practices, and native plant species in different Arab regions. \\
    \textbf{Local Geography} & Focuses on terrain diversity, water resources, economic impact, and geographical landmarks of specific Arab countries. \\
    \textbf{Celebrations} & Highlights national, historical, and religious festivals, their cultural significance, associated traditions, and unique practices in different Arab communities. \\
    \textbf{Language} & Examines Modern Standard Arabic (MSA) and dialectal variations, including translation tasks, word usage, and sentence restructuring between dialects and MSA. \\
    \textbf{Proverbs} & Captures the cultural relevance of Arabic proverbs, their meanings, usage, and context in everyday conversations. \\
    \bottomrule
\end{tabular}
}

\caption{Instruction areas/domains in \textit{Palm} categorized by their relevance to general, hybrid, and country-specific knowledge.}
\label{tab:domains}
\end{table*}

\section{Diverse Instruction Formats and Linguistic Variations}
\label{apndx:instruction_types}
\begin{figure*}[ht]
  \centering
  
  % Row 1
  \begin{subfigure}{0.40\linewidth}
    \centering
    \includegraphics[width=\linewidth]{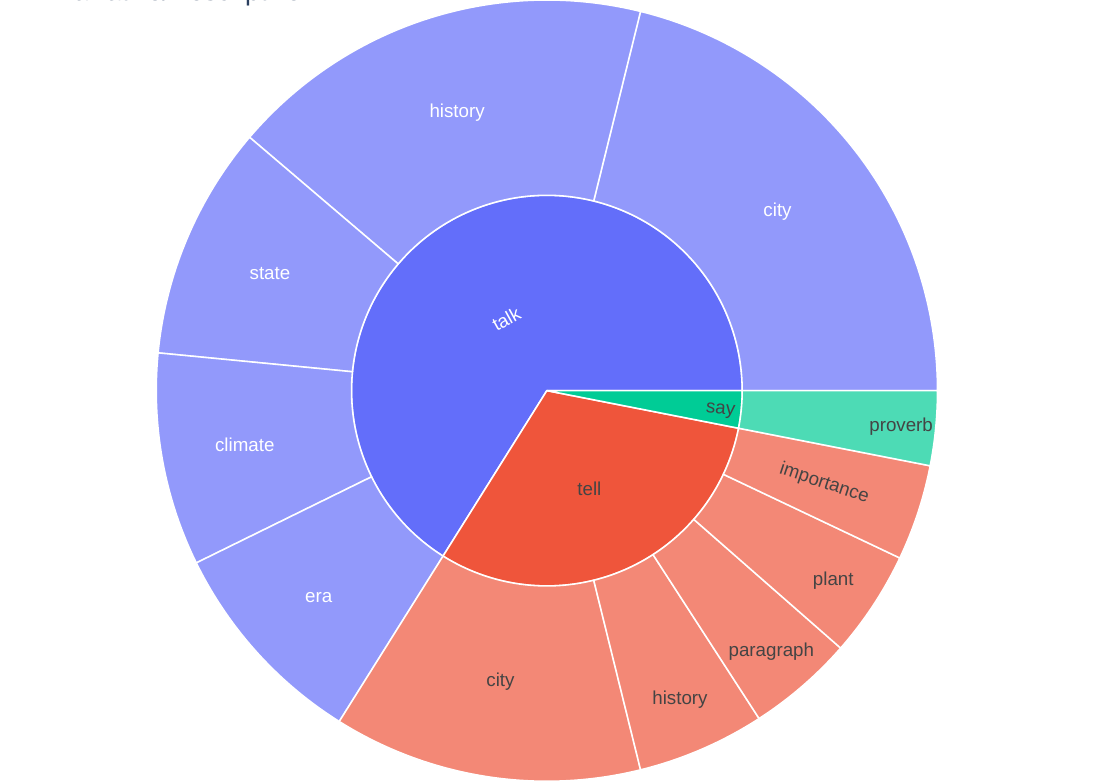}
    \caption{Narrative/Descriptive}
    \label{fig:sub:narrative_descriptive}
  \end{subfigure}
  \hfill
  \begin{subfigure}{0.40\linewidth}
    \centering
    \includegraphics[width=\linewidth]{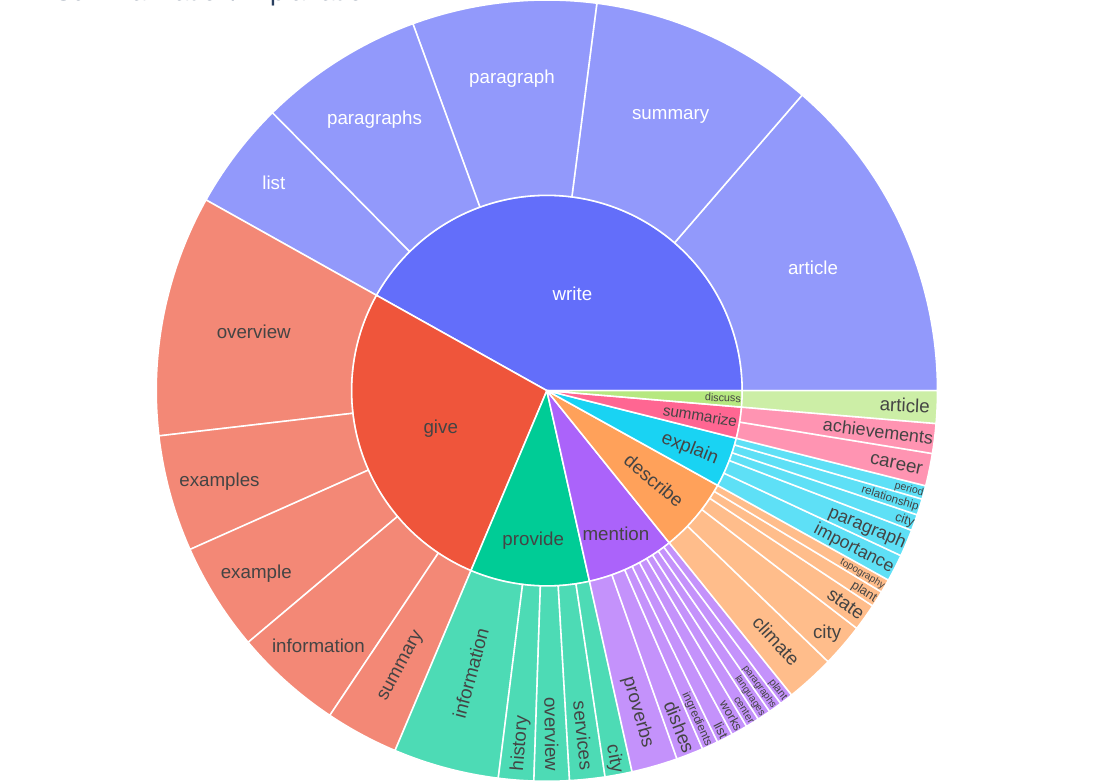}
    \caption{Summarization/Explanation}
    \label{fig:sub:summarization_explanation}
  \end{subfigure}
  
  \vspace{1em} % Adjust vertical spacing as needed
  
  % Row 2
  \begin{subfigure}{0.40\linewidth}
    \centering
    \includegraphics[width=\linewidth]{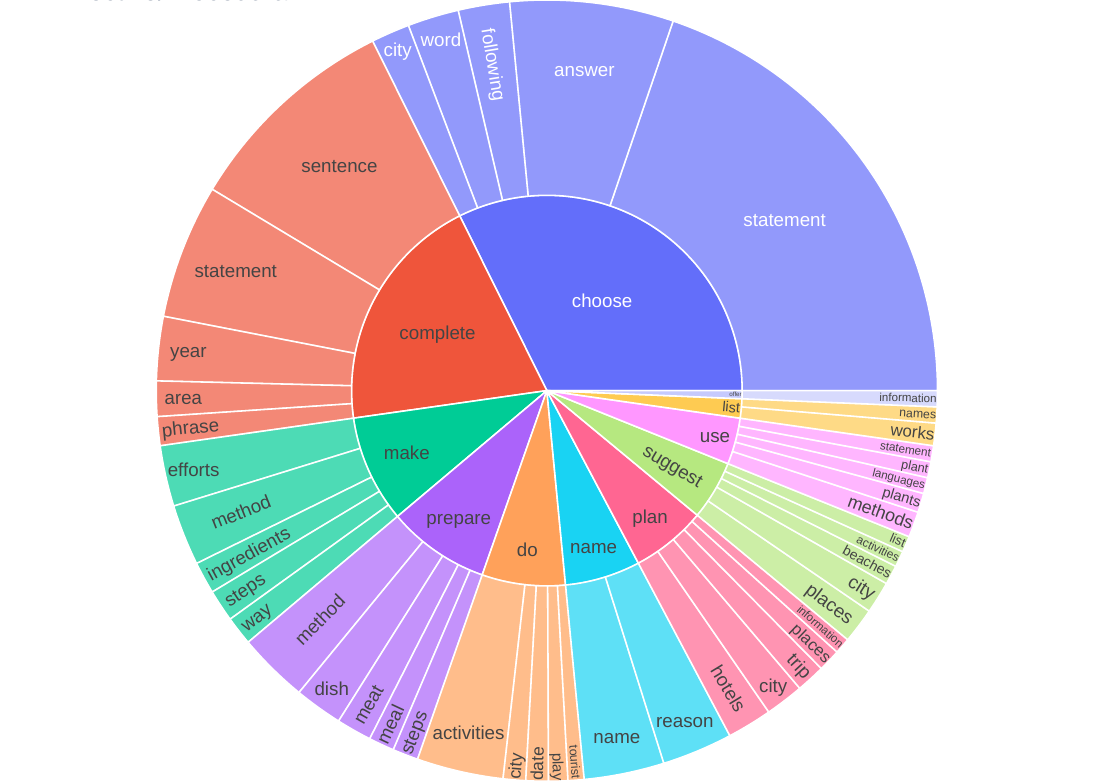}
    \caption{Directive/Procedural}
    \label{fig:sub:directive_procedural}
  \end{subfigure}
  \hfill
  \begin{subfigure}{0.40\linewidth}
    \centering
    \includegraphics[width=\linewidth]{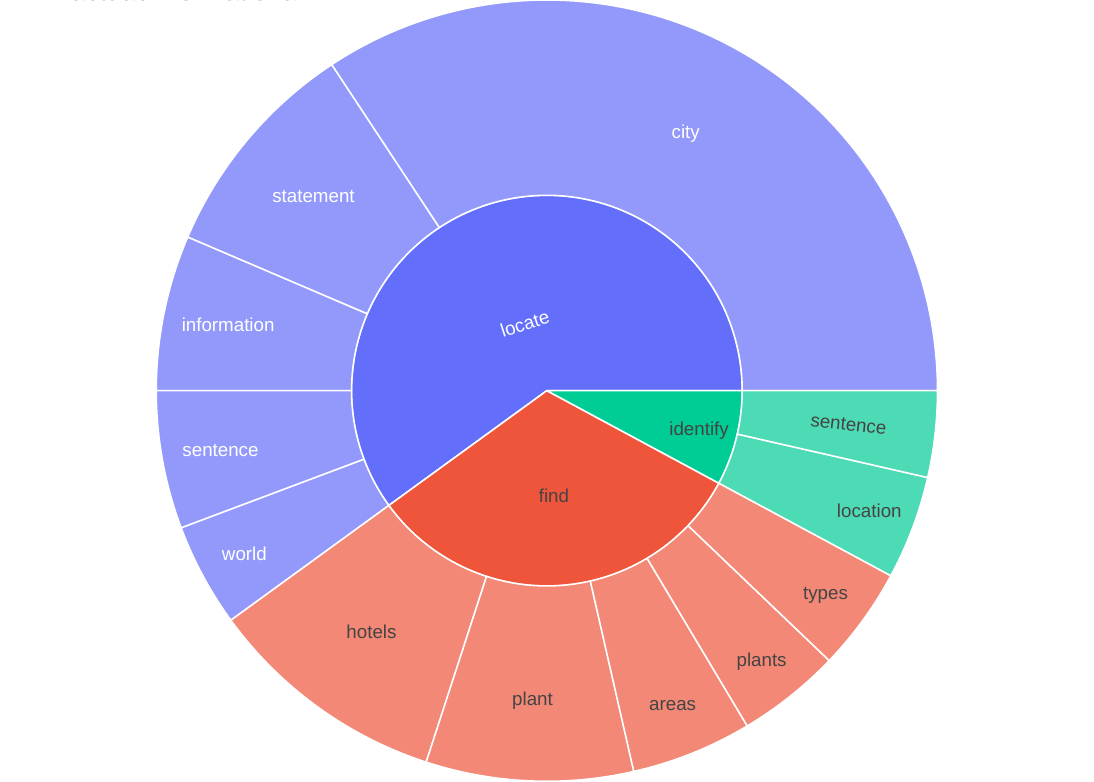}
    \caption{Factual/Informational}
    \label{fig:sub:factual_informational}
  \end{subfigure}
  
  \vspace{1em} % Adjust vertical spacing as needed
  
  % Row 3
  \begin{subfigure}{0.40\linewidth}
    \centering
    \includegraphics[width=\linewidth]{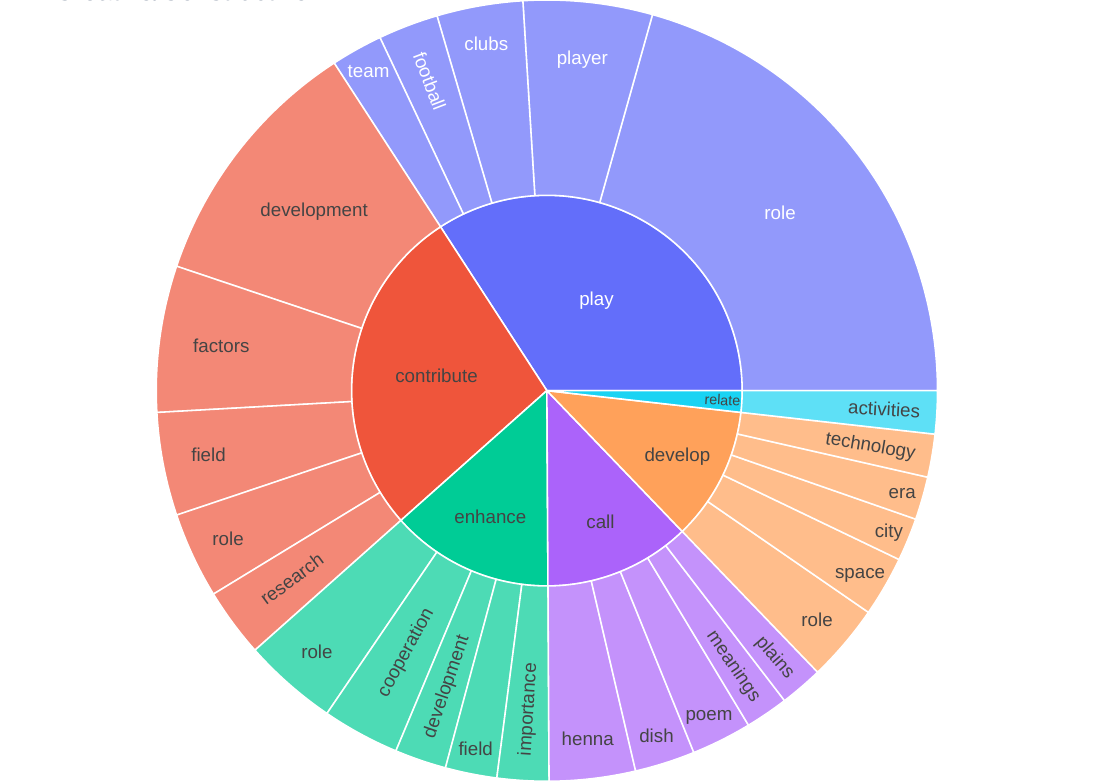}
    \caption{Creative/Constructive}
    \label{fig:sub:creative_constructive}
  \end{subfigure}
  \hfill
  \begin{subfigure}{0.40\linewidth}
    \centering
    \includegraphics[width=\linewidth]{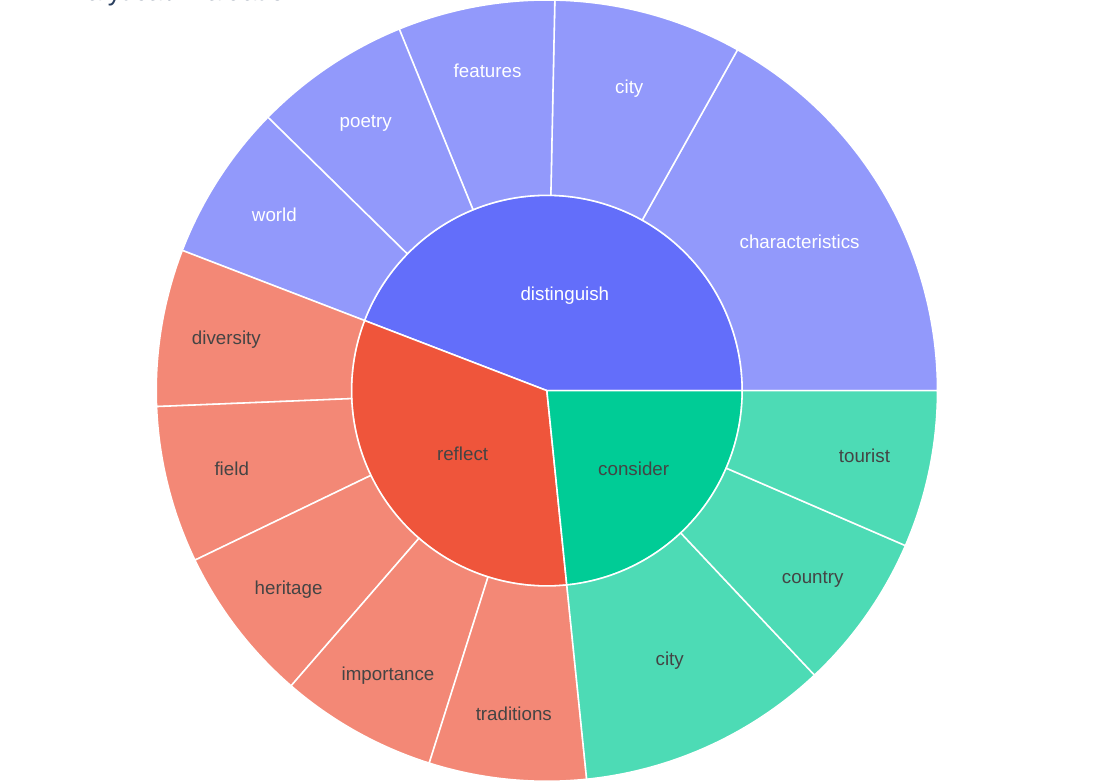}
    \caption{Analytical/Evaluation}
    \label{fig:sub:analytical_evaluation}
  \end{subfigure}
  
  \vspace{1em} % Adjust vertical spacing as needed
  
  % Row 4
  \begin{subfigure}{0.40\linewidth}
    \centering
    \includegraphics[width=\linewidth]{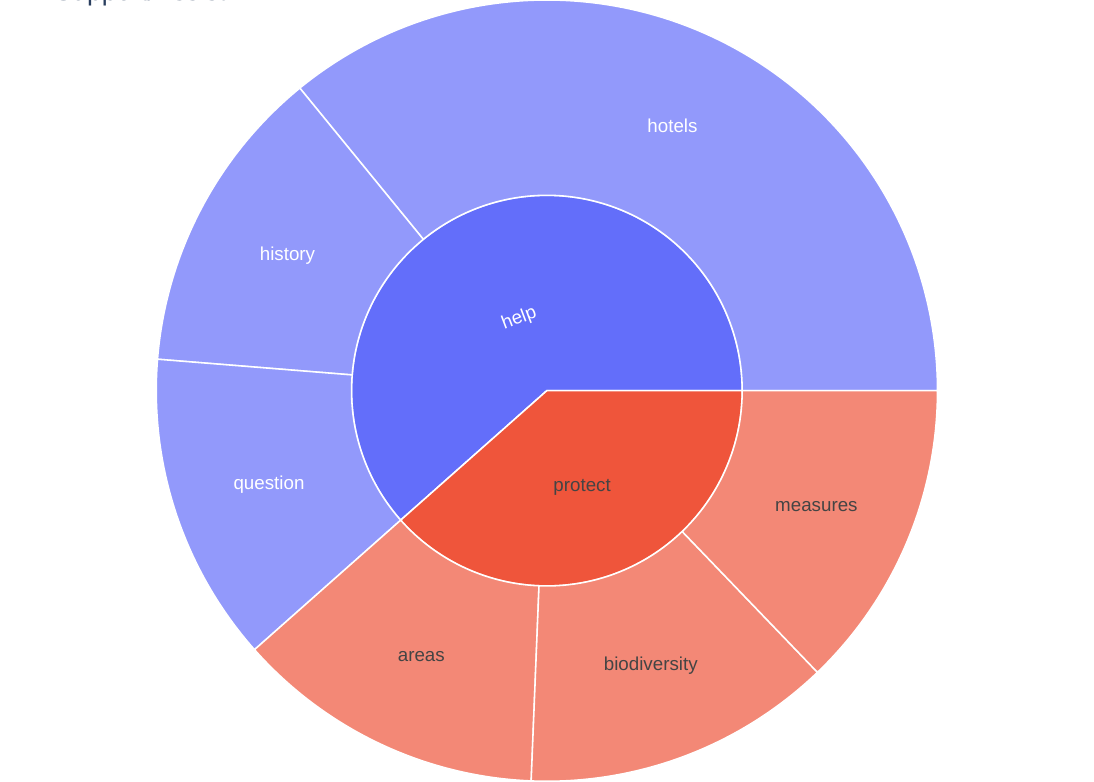}
    \caption{Support/Assist}
    \label{fig:sub:support_assist}
  \end{subfigure}
  \hfill
  \begin{subfigure}{0.40\linewidth}
    \centering
    \includegraphics[width=\linewidth]{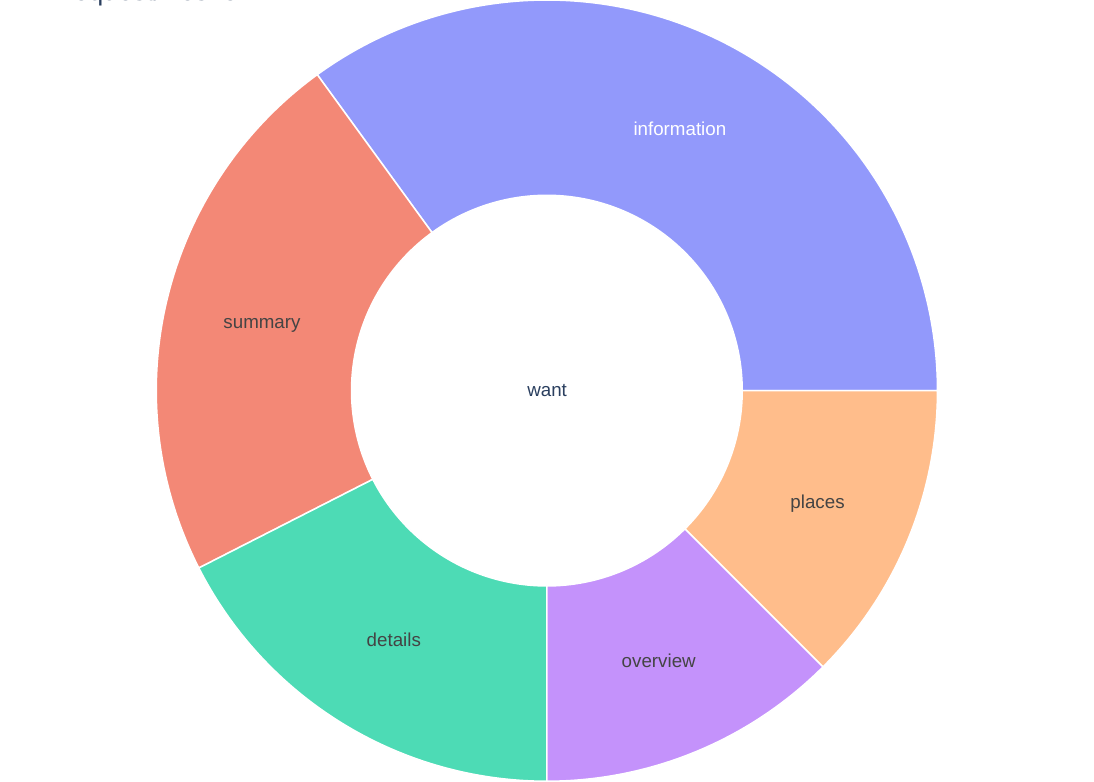}
    \caption{Request/Desire}
    \label{fig:sub:request_desire}
  \end{subfigure}
  
  \caption{Sunburst charts for each instruction type category in \textit{Palm}. Each subfigure represents a distinct theme derived from verb usage, providing insights into the dataset's instructional diversity.}
  \label{fig:all_sunbursts}
\end{figure*}

Figure~\ref{fig:all_sunbursts} presents sunburst charts for every instruction type category in \textit{Palm}. Each subfigure highlights a unique theme, derived from verb usage and the subsequent noun, that sheds light on the diversity of instructional approaches within the dataset. All of these subcharts are drilldowns from Figure~\ref{fig:instruction_distribution}.

\section{Statistics}
\label{apndx:statistics}

\begin{table*}[t]
\centering
\resizebox{.95\textwidth}{!}{%
\begin{tabular}{lrrrrrrrrrrrrrrrr|r|r}
\toprule
Topic & Celeb. & Env. & Flora & Food & General & Hist. & Lang. & Lit. & Geog. & Politics & Proverbs & Religion & Sports & Tech & Travel & Science & Dialect  & Total  \\
\midrule
Egypt        & 118          & 90               & 116   & 118  & ---       & 105     & 22       & 140        & 177             & 71       & 109      & 109      & 117    & 5          & 135    & 24      & 438     & 1480  \\
Jordan       & 109          & 122              & 100   & 112  & ---       & 74      & 125      & 113        & 98              & 83       & 100      & 98       & 186    & 79         & 114    & ---       & 500     & 1513  \\
Mauritania   & 69           & 28               & 13    & 109  & ---       & 285     & 105      & 78         & 343             & 40       & 112      & 39       & 37     & ---          & 38     & 1       & 294     & 1298  \\
Morocco      & 32           & 70               & 33    & 256  & ---       & 247     & 139      & 45         & 87              & 206      & 220      & 111      & 245    & 42         & 188    & 9       & 717     & 1939  \\
Palestine    & 46           & 102              & 143   & 181  & ---       & 103     & 202      & 130        & 137             & 60       & 178      & 114      & 109    & ---          & 13     & ---       & 525     & 1518  \\
Saudi Arabia & 109          & 200              & 61    & 42   & ---       & 104     & 10       & 140        & 163             & 17       & 39       & 61       & 142    & 100        & 111    & ---       & 296     & 1299  \\
Sudan        & 18           & ---                & ---     & 28   & ---       & 450     & ---        & 40         & 285             & 124      & ---        & 27       & 17     & ---          & 8      & ---       & ---       & 997   \\
Syria        & 31           & 72               & 201   & 131  & ---       & 202     & ---        & 198        & 65              & 55       & 240      & 100      & 38     & 4          & 202    & 26      & 551     & 1591  \\
Tunisia      & 12           & 16               & 13    & 40   & ---       & 48      & 51       & 13         & 48              & 11       & 16       & 43       & 50     & 92         & 14     & 160     & 29      & 787   \\
UAE          & 50           & 21               & 17    & 80   & ---       & 127     & ---        & 132        & 152             & 133      & ---        & 23       & 35     & 34         & 191    & 27      & 26      & 1049  \\
Yemen        & 40           & 16               & 53    & 50   & ---       & 172     & 251      & 119        & 114             & 92       & 262      & 43       & 235    & 25         & 46     & 114     & 592     & 1746  \\
\midrule
General      & 10           & 46               & 42    & 42   & ---       & 52      & 363      & 45         & 60              & 32       & ---        & 12       & 80     & 13         & 12     & 300     & ---       & 1409  \\
\midrule
Algeria      & 6            & ---                & ---     & 20   & ---       & 50      & ---        & 1          & 48              & 17       & ---        & 2        & 25     & ---          & 3      & ---       & ---       & 172   \\
Bahrain      & ---            & ---                & ---     & ---    & 100     & ---       & ---        & ---          & ---               & ---        & ---        & ---        & ---      & ---          & ---      & ---       & ---       & 100   \\
Comoros      & 7            & 4                & 10    & 8    & 13      & 6       & 5        & 7          & 6               & 3        & ---        & 5        & 8      & 4          & 15     & ---       & ---       & 101   \\
Djibouti     & 2            & 14               & ---     & 8    & 18      & 23      & 3        & ---          & 20              & 7        & ---        & 3        & ---      & 2          & ---      & ---       & ---       & 100   \\
Iraq         & 13           & 11               & ---     & 17   & ---       & 11      & 21       & 5          & ---               & 8        & ---        & ---        & 8      & ---          & 19     & ---       & ---       & 113   \\
Kuwait       & 15           & ---                & ---     & 21   & ---       & 16      & ---        & 16         & 16              & ---        & ---        & 16       & 16     & 16         & 15     & ---       & ---       & 147   \\
Lebanon      & ---            & ---                & ---     & ---    & 100     & ---       & ---        & ---          & ---               & ---        & ---        & ---        & ---      & ---          & ---      & ---       & ---       & 100   \\
Libya        & ---            & ---                & ---     & 1    & 100     & ---       & ---        & ---          & ---               & ---        & ---        & ---        & ---      & ---          & ---      & ---       & ---       & 101   \\
Oman         & ---            & ---                & ---     & ---    & 100     & ---       & ---        & ---          & ---               & ---        & ---        & ---        & ---      & ---          & ---      & ---       & ---       & 100   \\
Qatar        & ---            & ---                & 10    & 10   & 38      & 61      & ---        & 12         & 11              & ---        & ---        & ---        & 28     & ---          & 40     & ---       & ---       & 210   \\
Somalia      & 2            & 7                & ---     & 9    & 31      & 6       & ---        & ---          & 13              & 26       & ---        & ---        & 7      & ---          & ---      & ---       & ---       & 101  \\
\bottomrule
\end{tabular}
}
\caption{The overall statistics of instructions number per country, per topic, and per dialect. \textsuperscript{$\star$}\texttt{Celb.}: Celebrations, \textsuperscript{$\star$}\texttt{Env:} Environment, \textsuperscript{$\star$}\texttt{Hist.}: History, \textsuperscript{$\star$}\texttt{Lit.}: Literature, \textsuperscript{$\star$}\texttt{Geog.}: Geography.}
\label{tab:country_data_summary}
\end{table*}

\subsection{Comparative Analysis of Token Length Distributions Across Models}

\begin{figure*}[t]
  \centering
  \includegraphics[width=\linewidth]{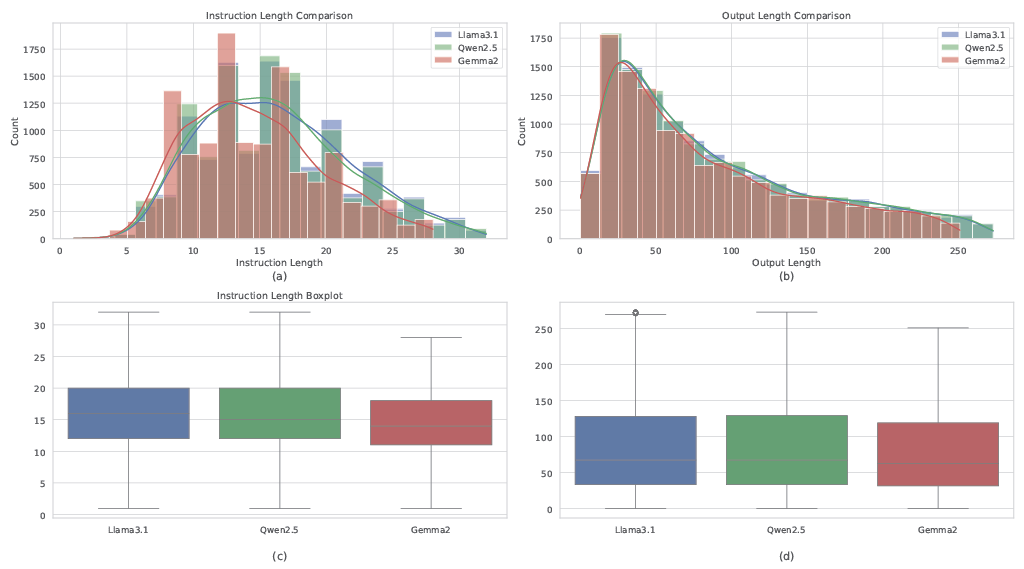}
  \caption{Token Lengths Comparison.}
  \label{fig:token-lengths} 
\end{figure*}

Figure~\ref{fig:token-lengths} presents a comparative analysis of instruction and output token lengths for three LLMs tokenizers: Llama 3.1, Qwen 2.5, and Gemma 2. The figure includes histograms with density curves for instruction and output lengths (top panels) and boxplots for the same data (bottom panels).

In the Instruction Lengths Figure~\ref{fig:token-lengths}(a), most instructions across all models range between 10 and 20 tokens. Notably, Gemma 2 exhibits a higher concentration of shorter instructions (5–15 tokens), while Llama 3.1 and Qwen 2.5 tend toward slightly longer instructions, with frequencies gradually declining beyond 20 tokens.

For the Output Lengths Figure~\ref{fig:token-lengths}(b), all models display a peak around 50 tokens, with distributions extending up to 250 tokens. Llama 3.1 tends to generate longer outputs overall, evident from a more pronounced tail toward higher token counts compared to the other models.

The Instruction Length Boxplots Figure~\ref{fig:token-lengths}(c) show that Llama 3.1 and Qwen 2.5 have similar distributions, with median instruction lengths around 15 tokens and comparable variability. Gemma 2 has a slightly shorter median length and a narrower spread, indicating less variation in instruction lengths.

In the Output Length Boxplots Figure~\ref{fig:token-lengths}(d), Llama 3.1 again produces the longest outputs, with a median around 90 tokens and outliers extending beyond 250 tokens. Qwen 2.5 and Gemma 2 have median output lengths around 70–80 tokens, with fewer extreme outliers.

Overall, this analysis demonstrates that Llama 3.1 generates longer outputs compared to Qwen 2.5 and Gemma 2, while Gemma 2 often produces shorter instructions. These variations highlight differences in how the models handle input-output lengths.

Additionally, Table~\ref{tab:avg_char_lengths} presents the average character lengths for instructions and outputs across various countries in the dataset. Countries like Tunisia, UAE, and Jordan have longer average instruction lengths, while Lebanon and Bahrain feature shorter instructions. For output lengths, Egypt and Somalia have the highest averages, while Qatar and Syria have shorter outputs. This table provides a detailed view of the character length variations across the dataset.

\begin{table}[h]
\centering
\resizebox{.30\textwidth}{!}{%
\begin{tabular}{lcc}
\toprule
\textbf{Country} & \textbf{Instruction} & \textbf{Response} \\ \midrule
Egypt & 52.34 & 1,444.13 \\ 
Jordan & 58.26 & 334.34 \\ 
Mauritania & 48.00 & 505.51 \\ 
Morocco & 56.10 & 816.22 \\ 
Palestine & 54.78 & 460.43 \\ 
Saudi Arabia & 52.55 & 444.23 \\ 
Somalia & 34.61 & 1,639.81 \\ 
Sudan & 39.85 & 528.33 \\ 
Syria & 45.76 & 246.92 \\ 
Tunisia & 73.26 & 957.76 \\ 
UAE & 72.27 & 280.12 \\ 
Yemen & 48.55 & 624.52 \\ 
\midrule
Algeria & 53.13 & 1,152.94 \\ 
Bahrain & 33.65 & 1,039.91 \\ 
Comoros & 51.95 & 479.75 \\ 
Djibouti & 39.43 & 760.79 \\ 
Iraq & 34.38 & 1,010.91 \\ 
Kuwait & 56.39 & 479.41 \\ 
Lebanon & 27.94 & 336.42 \\ 
Libya & 30.11 & 613.85 \\ 
Qatar & 59.86 & 122.10 \\ 
Oman & 31.70 & 381.61 \\ \bottomrule
\end{tabular}
}
\caption{Average character length for instructions and responses by country.}
\label{tab:avg_char_lengths}
\end{table}

\subsection{Lexical Onset Analysis of Prompts}

\begin{figure}[ht]

  \centering
  \includegraphics[width=\linewidth]{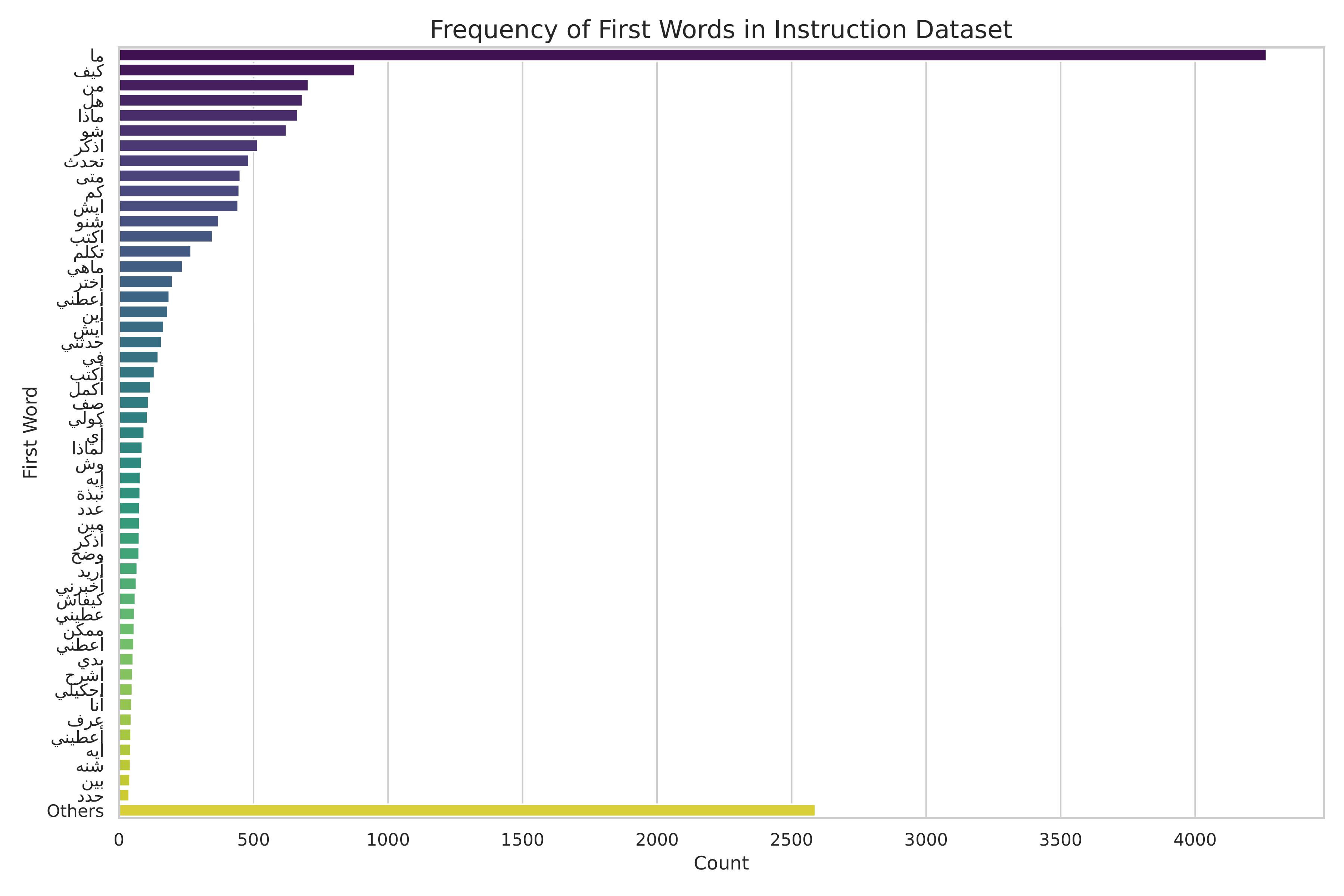}
  \caption{Frequency of first words.}
  \label{fig:first-word-count} 
\end{figure}

Figure~\ref{fig:first-word-count} presents a bar plot illustrating the distribution of first words, defined as space-delimited strings, in instructions within the \textbf{Palm} dataset. The most frequent initial word is "$\RL{ما}$" ("what"), appearing \textbf{4,265} times, followed by "$\RL{كيف}$" ("how") with \textbf{877} occurrences and "$\RL{من}$" ("who/from") with \textbf{704} instances. This indicates a strong emphasis on informational and definitional queries. Additionally, verbs such as "$\RL{اذكر}"$ ("provide"/"list") and "$\RL{اكتب}$" ("write") are prevalent, appearing \textbf{516} and \textbf{348} times respectively, suggesting a focus on task-oriented instructions. The presence of dialectal variations like "$\RL{شو}$" ("what") and "$\RL{شنه}$" ("what") alongside standard forms underscores the dataset's comprehensive coverage of both MSA and colloquial dialects. The 'Others' category, comprising \textbf{2,589} instances, reflects the dataset's diversity in addressing various user queries. Overall, the distribution reveals that \ourdataset facilitates information retrieval and explanatory responses, essential for training LLMs to handle a wide range of culturally nuanced and linguistically diverse inquiries effectively.

\section{Revision Process}
\label{apndx:revision_porcess}

To quantitatively assess the changes made during the review process, we employed Python's \texttt{difflib.SequenceMatcher} to compute similarity ratios between the original and revised versions of the instructions, inputs, and outputs. Specifically, we defined the difference score as \(1 - \text{similarity}\) and observed mean differences of 0.012 for instructions, 0.001 for inputs, and 0.017 for outputs. Notably, only 22\% of the samples underwent any modifications, reflecting the effectiveness of our rigorous annotator training and the weekly meetings held to address emerging issues early in the project. This proactive approach ensured clarity and consistency throughout the annotation workflow.

We also examined a sample of 200 instructions to characterize the types of revisions made during the second phase of data review. These revisions primarily fell into three categories. First, \textbf{grammar and mechanics} revisions included punctuation adjustments (such as adding missing marks or removing extraneous ones) and grammatical corrections (e.g., fixing subject–verb agreement, verb tenses, and pronoun usage). Second, \textbf{question revisions} involved rephrasing or clarifying questions to preserve their core meaning. Third, \textbf{answer revisions} comprised either summarizing responses to highlight key points or expanding them by adding further details and context.

\section{Comparative Analysis of Arabic Instructional Datasets}
\label{apndx:comparative_analysis}

Recent efforts in developing instructional datasets for Arabic language processing have produced a variety of resources, each with distinct strengths and limitations. Multilingual datasets such as AYA~\cite{singh2024aya} and BLEnD~\cite{myung2024blend} have contributed valuable resources by including Arabic instructions; however, their focus on multiple languages means that Arabic-specific nuances are underrepresented. For example, while AYA provides 5K Arabic instructions out of a total of 204K, BLEnD offers only 3.6K Arabic instructions among 55K entries, and BLEnD’s coverage is limited to just one Arab country. In contrast, Arabic-specific datasets like AraDiCE~\cite{mousi2024aradice} and CIDAR~\cite{alyafeai2024cidar} have been developed to capture more localized content. AraDiCE, which spans six dialects, often relies on translation and data retargeting methods, and its limited number of native instructions (180 out of 45K) may not fully capture the linguistic diversity. Similarly, although CIDAR contains a full set of 10K Arabic instructions with human revisions, it lacks the breadth in geographic and dialectal diversity.

Our dataset, \ourdataset, addresses these gaps by providing a more comprehensive resource tailored to the Arabic language. It uniquely covers 22 Arab countries and incorporates 10 Arabic dialects, ensuring broader cultural and regional representation. Importantly, \ourdataset is the only dataset in this comparison that is built entirely from scratch through human collection and revision, rather than relying on machine translation or localization. Moreover, by focusing on open-ended instructional prompts—including tasks such as writing, role-playing, and reasoning—\ourdataset offers richer linguistic expressions and a more authentic reflection of native language use. This meticulous design aims to better support the development and evaluation of Arabic language models in a variety of real-world applications.

\section{Evaluation Prompt and Metrics}
\label{apndx:evaluation_prompt}
The evaluation metrics are defined as follows (on a scale of $1$ to $10$):
\begin{enumerate}
    \item \textbf{Correctness}: Measures the factual accuracy of the response in relation to the instruction. A correct response should provide accurate information without errors or misconceptions.
    \item \textbf{Coherence}: Evaluates the logical consistency and clarity of the response. A coherent response should be well-structured, logically organized, and easy to understand.
    \item \textbf{Helpfulness}: Determines the utility of the response to the user. A helpful response should provide valuable information that satisfies the user's needs.
    \item \textbf{Details}: Measures the depth and comprehensiveness of the response. A detailed response should provide sufficient elaboration and cover relevant aspects of the topic.
\end{enumerate}

\begin{figure}
    \centering
    \includegraphics[width=0.99\linewidth]{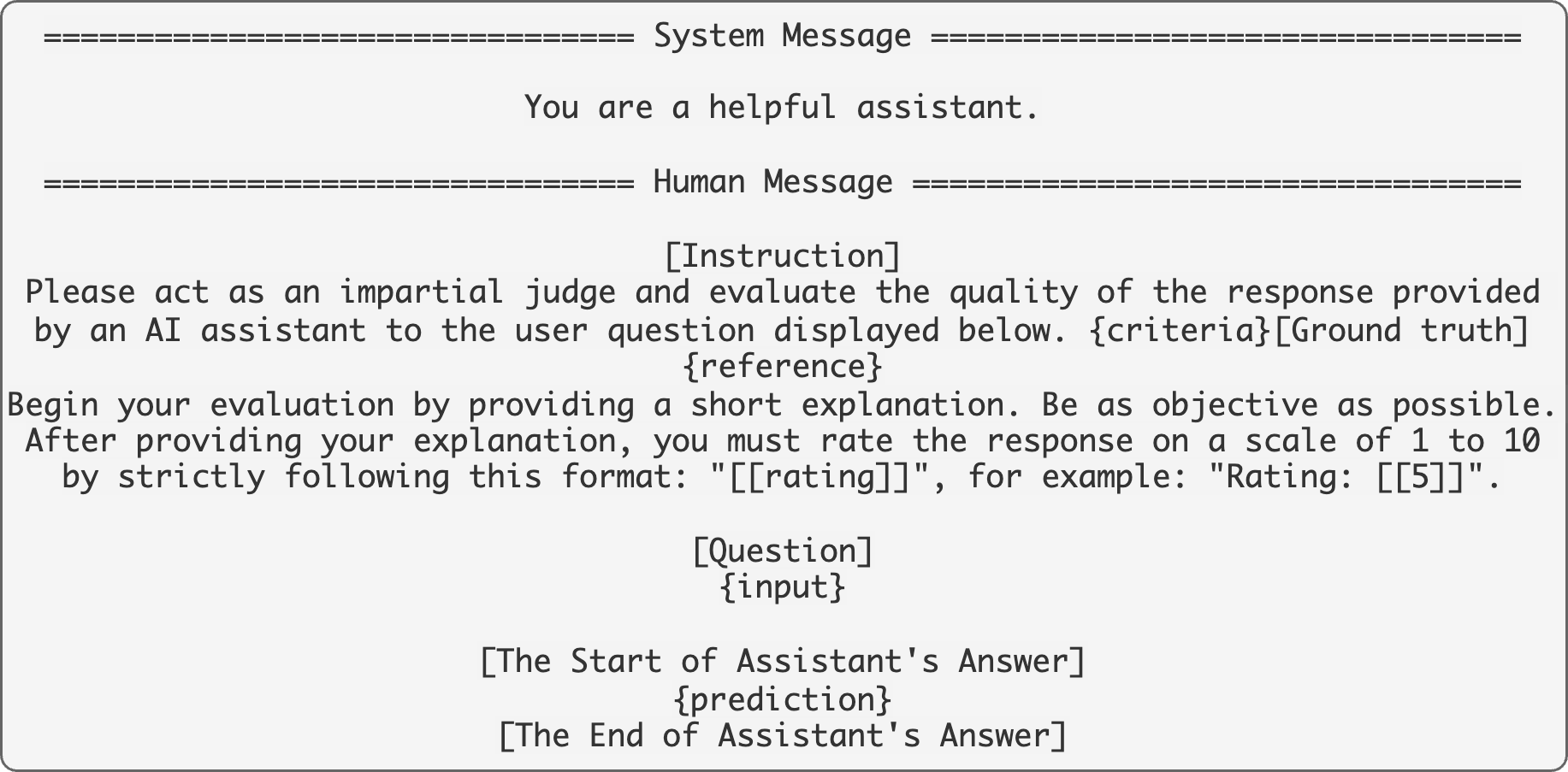}
    \caption{Prompt used in our evaluation.}
    \label{fig:prompt}
\end{figure}

\section{Selected Examples}
\label{apndx:examples}

% Increase row spacing
\renewcommand{\arraystretch}{1.2}
% Activate ArabTeX
\setarab

\begin{table*}[htbp]
\centering

% Define custom colors to match the image (light green header and very light green row)
\definecolor{headercolor}{HTML}{DFF0D8}
\definecolor{rowcolor}{HTML}{F9FFF3}

% Set the table border color to a neutral gray or black
\arrayrulecolor{black}

% Scale the entire table down to 70%
\scalebox{0.7}{%
  % Apply alternating row colors, starting with the second row
  \rowcolors{2}{rowcolor}{white}
  \begin{tabular}{|p{2cm}|p{2cm}|p{8cm}|p{8cm}|}
    \hline
    % Header row with custom background
    \rowcolor{headercolor}
    \textbf{Country} & \textbf{Topic} & \textbf{Instruction} & \textbf{Output} \\
    \hline
    Syria & literature &
    \begin{minipage}[t]{\linewidth}
      \begin{RLtext}
      أي من هؤلاء الأدباء الذين ظهروا في القرن العشرين في سوريا ؟\\
      نزار قباني، سليمان العيسى
      \end{RLtext}
      \vspace{0.5em}
      Which of these writers emerged in 20th-century Syria?\\
      Nizar Qabbani, Suleiman al-Issa
    \end{minipage}
    &
    \begin{minipage}[t]{\linewidth}
      \begin{RLtext}
      نزار قباني و سليمان العيسى
      \end{RLtext}
      \vspace{0.5em}
      Nizar Qabbani and Suleiman al-Issa
    \end{minipage} \\
    \hline
    Tunisia & history &
    \begin{minipage}[t]{\linewidth}
      \begin{RLtext}
      رتب هذه الدول زمنيا:\\
      - الدولة الصنهاجية\\
      - الدولة الموحدية\\
      - الدولة الأغلبية\\
      - الدولة الحفصية\\
      - الدولة الفاطمية
      \end{RLtext}
      \vspace{0.5em}
      Arrange these states in chronological order:\\
      - The Sanhaji state\\
      - The Almohad state\\
      - The Aghlabid state\\
      - The Hafsid state\\
      - The Fatimid state
    \end{minipage}
    &
    \begin{minipage}[t]{\linewidth}
      \begin{RLtext}
      بالتأكيد. هذا هو الترتيب الزمني الصحيح للدول المذكورة التي تعاقبت...
      \end{RLtext}
      \vspace{0.5em}
      Certainly. This is the correct chronological order of the mentioned states that succeeded each other...
    \end{minipage} \\
    \hline
    Saudi Arabia & travel &
    \begin{minipage}[t]{\linewidth}
      \begin{RLtext}
      ماذا تعرف عن المسجد الحرام؟
      \end{RLtext}
      \vspace{0.5em}
      What do you know about Masjid al-Haram?
    \end{minipage}
    &
    \begin{minipage}[t]{\linewidth}
      \begin{RLtext}
      على الرحب والسعة، المسجد الحرام أحد أهم وأشهر معالم السياحة...
      \end{RLtext}
      \vspace{0.5em}
      You're very welcome, Masjid al-Haram is one of the most important and renowned tourist landmarks...
    \end{minipage} \\
    \hline
    Palestine & travel &
    \begin{minipage}[t]{\linewidth}
      \begin{RLtext}
      فيما لا يقل عن 5 أسطر، تحدث لي عن الحرم الإبراهيمي في مدينة الخليل.
      \end{RLtext}
      \vspace{0.5em}
      In at least 5 lines, tell me about the Ibrahimi Mosque in the city of Hebron.
    \end{minipage}
    &
    \begin{minipage}[t]{\linewidth}
      \begin{RLtext}
      المسجد الإبراهيميّ، أو الحرم الإبراهيميّ الشريف، وهو عند اليهود باسم...
      \end{RLtext}
      \vspace{0.5em}
      The Ibrahimi Mosque, or the Noble Ibrahimi Sanctuary, is known among the Jews as...
    \end{minipage} \\
    \hline
    Morocco & travel &
    \begin{minipage}[t]{\linewidth}
      \begin{RLtext}
      المدرسة البوعنانية هي مدرسة تاريخية تقع في مدينة فاس المغربية. تحدث عنها في جملة واحدة
      \end{RLtext}
      \vspace{0.5em}
      The Bouanania School is a historic school located in the city of Fez, Morocco. Talk about it in one sentence.
    \end{minipage}
    &
    \begin{minipage}[t]{\linewidth}
      \begin{RLtext}
      المدرسة البوعنانية هي مدرسة تاريخية تقع في مدينة فاس المغربية....
      \end{RLtext}
      \vspace{0.5em}
      
    \end{minipage} \\
    \hline
    Jordan & celebrations &
    \begin{minipage}[t]{\linewidth}
      \begin{RLtext}
      اكتب مقالاً من أربع فقرات توضح فيه أهمية مهرجان جرش
      \end{RLtext}
      \vspace{0.5em}
      Write an essay consisting of four paragraphs that explains the significance of the Jerash Festival.
    \end{minipage}
    &
    \begin{minipage}[t]{\linewidth}
      \begin{RLtext}
      يلعب المهرجان الذي يستند على الإرث التاريخي والإبداعي العريق دورًا...
      \end{RLtext}
      \vspace{0.5em}
      The festival, which is based on a rich historical and creative heritage, plays a role...
    \end{minipage} \\
    \hline
    Egypt & travel &
    \begin{minipage}[t]{\linewidth}
      \begin{RLtext}
      رشح أفضل 10 أماكن لتناول الغذاء في الإسكندرية.
      \end{RLtext}
      \vspace{0.5em}
      Recommend the top 10 places to eat in Alexandria.
    \end{minipage}
    &
    \begin{minipage}[t]{\linewidth}
      \begin{RLtext}
      بكل تأكيد!\\
      إليك تقرير عن أفضل 10 مطاعم في الإسكندرية\\[0.5em]
      أفضل مطاعم...
      \end{RLtext}
      \vspace{0.5em}
      Certainly!\\
      Here is a report on the top 10 restaurants in Alexandria\\[0.5em]
      Top restaurants...
    \end{minipage} \\
    \hline
    Algeria & travel &
    \begin{minipage}[t]{\linewidth}
      \begin{RLtext}
      اقترح افضل الشواطئ في تلمسان التي يجب زيارتها.
      \end{RLtext}
      \vspace{0.5em}
      Suggest the best beaches in Tlemcen to visit.
    \end{minipage}
    &
    \begin{minipage}[t]{\linewidth}
      \begin{RLtext}
      بما أن تلمسان مدينة ساحلية تطل على البحر الابيض المتوسط،...
      \end{RLtext}
      \vspace{0.5em}
      Since Tlemcen is a coastal city overlooking the Mediterranean Sea,...
    \end{minipage} \\
    \hline
    Yemen & food &
    \begin{minipage}[t]{\linewidth}
      \begin{RLtext}
      وضح طريقة عمل مندى الدجاج فى خطوات بسيطة؟
      \end{RLtext}
      \vspace{0.5em}
      Explain how to prepare chicken mandi in simple steps?
    \end{minipage}
    &
    \begin{minipage}[t]{\linewidth}
      \begin{RLtext}
      طريقة عمل مندى الدجاج فى خطوات بسيطة، لتقديم وجبة مغذية...
      \end{RLtext}
      \vspace{0.5em}
      The method for preparing chicken mandi in simple steps, to serve a nutritious meal...
    \end{minipage} \\
    \hline
  \end{tabular}%
} % end scalebox
\caption{Examples of Arabic and English instructions and outputs. All examples are derived from Figure~\ref{fig:main}, translated to English, and the answers have been abbreviated for space.}
\label{tab:palm_examples}
\end{table*}

Table~\ref{tab:palm_examples} presents a collection of combined Arabic and English examples of instructions and outputs. These examples, originally taken from Figure~\ref{fig:main}, have been translated into English with the answers shortened to save space.

\section{Evaluation}

\subsection{Evaluated LLMs} \label{append:evaluated_llms}

Table~\ref{tab:bench_llms} enumerates the LLMs employed to generate evaluation answers for \ourdataset. These models were selected from a curated list of Arabic-aware systems, with each entry including its size (in billions of parameters) and release date. Note that we used the instruct version for all LLMs.

% \usepackage{multirow}
\begin{table}[]
\centering
\resizebox{0.48\textwidth}{!}{%
\begin{tabular}{llrr}
\toprule
                                      & \textbf{LLM}      & \textbf{Size} & \textbf{Release Date} \\ \midrule
\multirow{2}{*}{\textbf{Closed LLMs}} & Claude-3.5-Sonnet & -             & Jun. 2024             \\
                                      & GPT-4o            & -             & Aug. 2024             \\ \midrule
\multirow{16}{*}{\textbf{Open LLMs}}  & Command R+        & 104B          & Aug. 2024             \\
                                      & Qwen2.5-72B       & 72B           & Sep. 2024             \\
                                      & Llama-3.1-70B     & 70B           & Jul. 2024             \\
                                      & AceGPT-v2-32B     & 32B           & Jun. 2024             \\
                                      & gemma-2-27b       & 27B           & Jul. 2024             \\
                                      & gemma-2-9b        & 8B            & Jul. 2024             \\
                                      & Llama-3.1-8B      & 8B            & Jul. 2024             \\
                                      & AceGPT-v2-8B      & 8B            & Jun. 2024             \\
                                      & Qwen2.5-7B        & 7B            & Sep. 2024             \\
                                      & jais-13b          & 13B           & Aug. 2023             \\
                                      & Phi-3.5-mini      & 3.8B          & Aug. 2024             \\
                                      & Qwen2.5-3B        & 3B            & Sep. 2024             \\
                                      & Llama-3.2-3B      & 3B            & Sep. 2024             \\
                                      & gemma-2-2b        & 2B            & Jul. 2024             \\
                                      & Qwen2.5-1.5B      & 1.5B          & Sep. 2024             \\
                                      & Llama-3.2-1B      & 1B            & Sep. 2024             \\ \bottomrule
\end{tabular}}
\caption{The LLMs used to generate answers for evaluation of \ourdataset were selected from a list of Arabic-aware models. Each LLM with its corresponding size in Billion parameters and release date. We used the instruct version for all LLMs.
}
\label{tab:bench_llms}
\end{table}

\subsection{LLM-as-Judge Results}\label{append:llm_judge_res}
\begin{figure}[]
  \centering
  \includegraphics[width=\columnwidth]{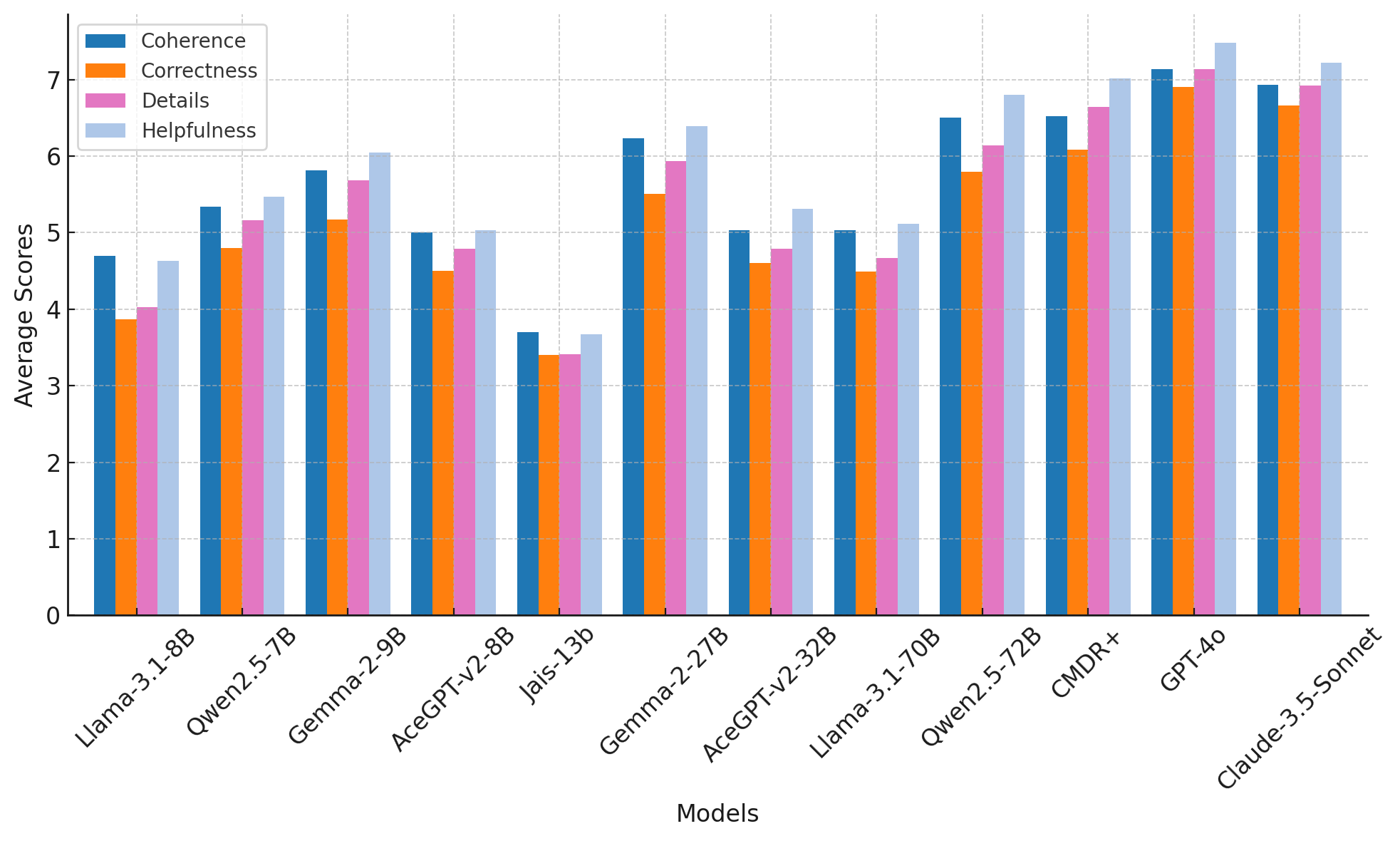}
  \caption{Performance comparison of 11 evaluated LLMs across the four metrics: \textit{correctness}, \textit{coherence}, \textit{helpfulness}, and \textit{details}.} 
  \label{fig:avg_four_metrics}
\end{figure}

\paragraph{Overall Results for Different Metrics.} Figure~\ref{fig:avg_four_metrics} presents a comparison of the results for the four evaluated metrics using LLM-as-Judge: \textit{coherence}, \textit{correctness}, \textit{details}, and \textit{helpfulness}. Across all models, there is a general trend of higher scores in \textit{coherence} and \textit{helpfulness} compared to \textit{correctness} and \textit{details}. Claude-3.5-sonnet, GPT-4o, and Command R+ consistently achieve the highest scores across all four metrics. Specifically, GPT-4o achieves an average score of $7.14$ in \textit{coherence} and $7.48$ in \textit{helpfulness}, while Claude-3.5-Sonnet follows closely with scores of $6.93$ and $7.22$, respectively. Their performance in \textit{correctness} and \textit{details} is similarly strong, highlighting their well-rounded capabilities. Interestingly, even smaller models like Gemma-2-7B show competitive performance in \textit{coherence} and \textit{helpfulness}, though they tend to lag in \textit{correctness} and details. The chart also reveals that as model size increases, there is typically an improvement across all metrics, with the most pronounced gains observed in \textit{correctness} and \textit{details}. This trend underscores the impact of model scale on performance across various aspects of language understanding and generation.

\paragraph{Per-Topic Results.}
Figure~\ref{fig:avg_correctness_topics} in Appendix~\ref{append:llm_judge_res} presents the performance of various models across different topics. \texttt{GPT-4o} and \texttt{Claude-3.5-Sonnet} consistently exhibit superior performance, with scores frequently above $6.0$. For instance, \texttt{GPT-4o} achieves a top score of $7.4$ in the \textit{History} category, while \texttt{Claude-3.5-Sonnet} scores $7.0$ in both \textit{History} and \textit{Proverbs}. In contrast, models such as \texttt{Llama-3.1-8B} and \texttt{Jais-13b} generally perform worse, often scoring below $4.0$ in multiple topics. The \textit{Food} category appears particularly challenging, displaying lower scores compared to other areas. Some models show particular strengths in specific domains. For example, \texttt{Qwen2.5-72B} scores $6.3$ in \textit{Celebrations} and $6.4$ in \textit{Science}, while \texttt{Gemma-2-27B} earns $5.9$ in both \textit{Flora} and \textit{Science}.

\begin{figure*}[h!]
    \centering

    \begin{subfigure}[b]{0.5\textwidth}
        \centering
        \includegraphics[width=\textwidth]{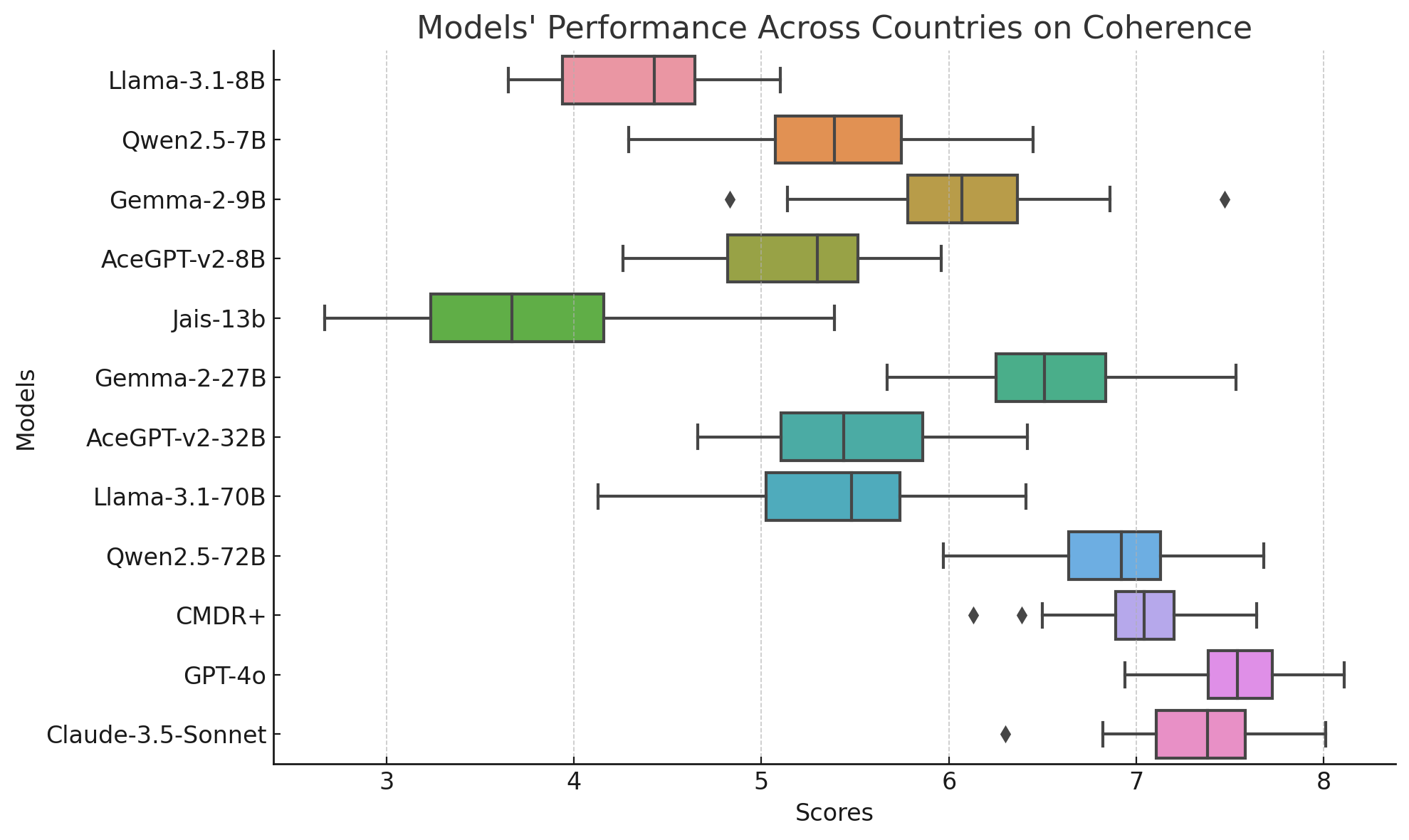}
        \caption{ }
        \label{fig:Average_coherence_coherence_boxplot}
    \end{subfigure}
    \hfill
    \begin{subfigure}[b]{0.49\textwidth}
        \centering
        \includegraphics[width=\textwidth]{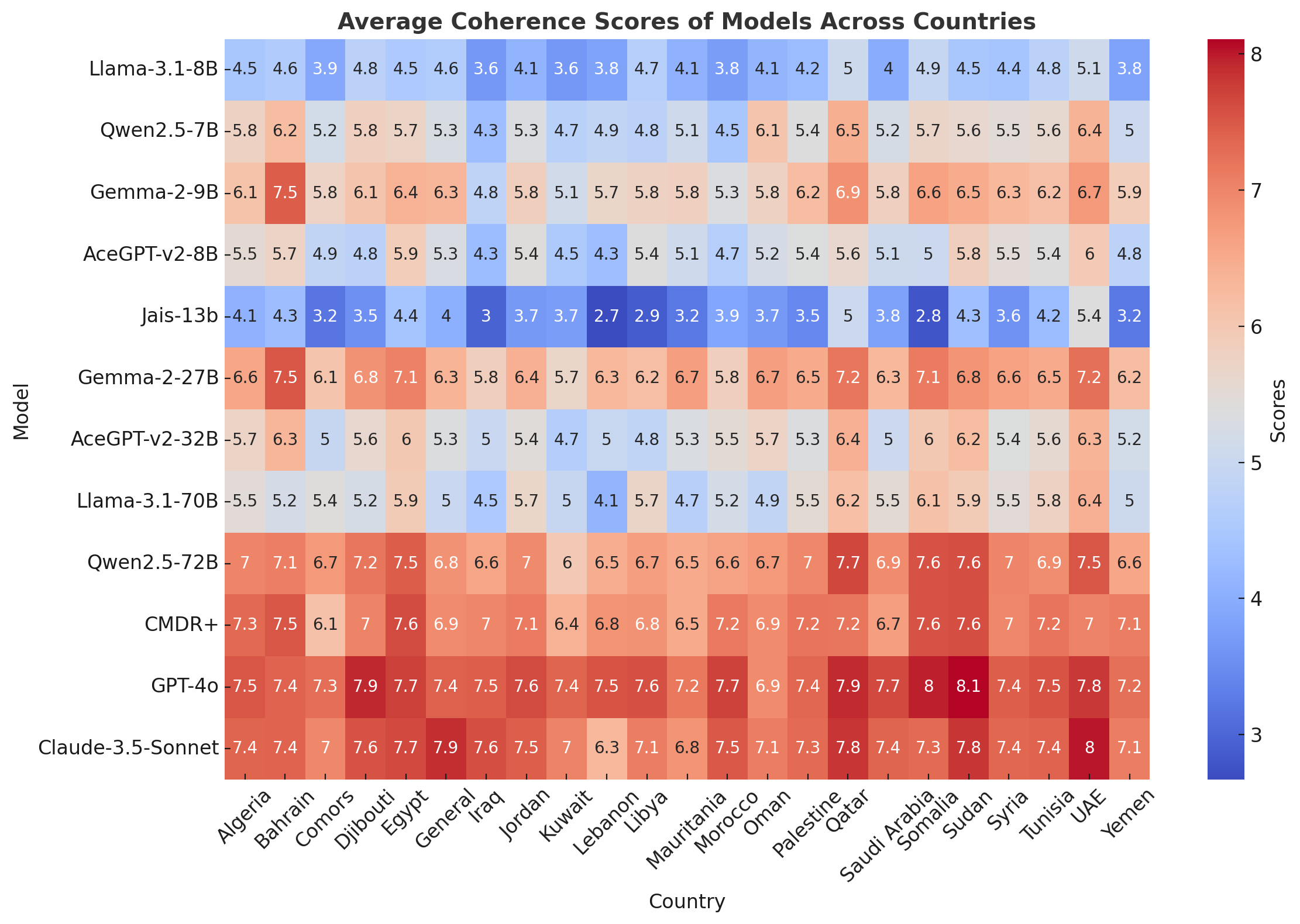}
        \caption{}
        \label{fig:Average_coherence_country_heatmap}
    \end{subfigure}
    \vskip\baselineskip
    \begin{subfigure}[b]{0.5\textwidth}
        \centering
        \includegraphics[width=\textwidth]{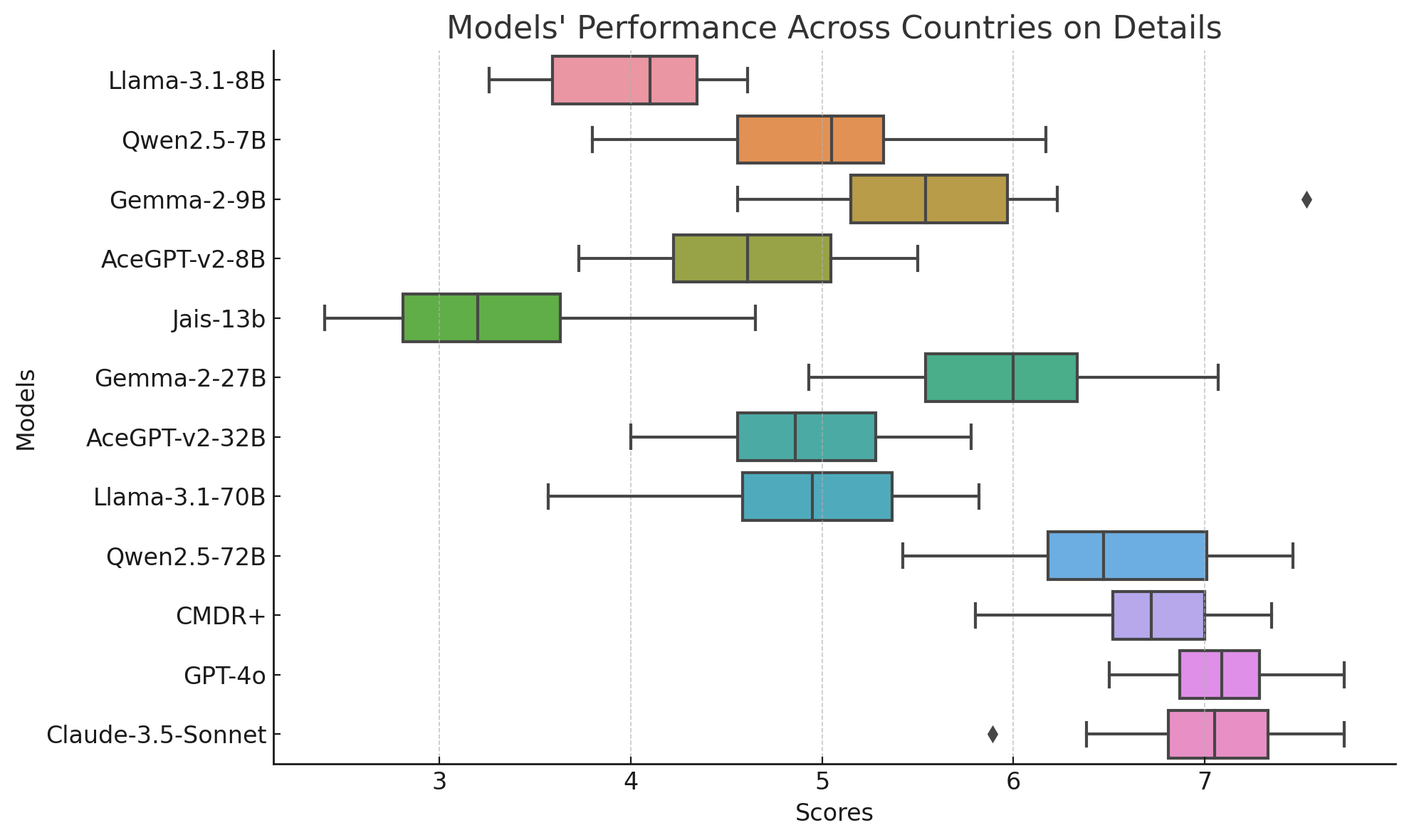}
        \caption{ }
        \label{fig:Average_details_country_boxplot}
    \end{subfigure}
    \hfill
    \begin{subfigure}[b]{0.49\textwidth}
        \centering
        \includegraphics[width=\textwidth]{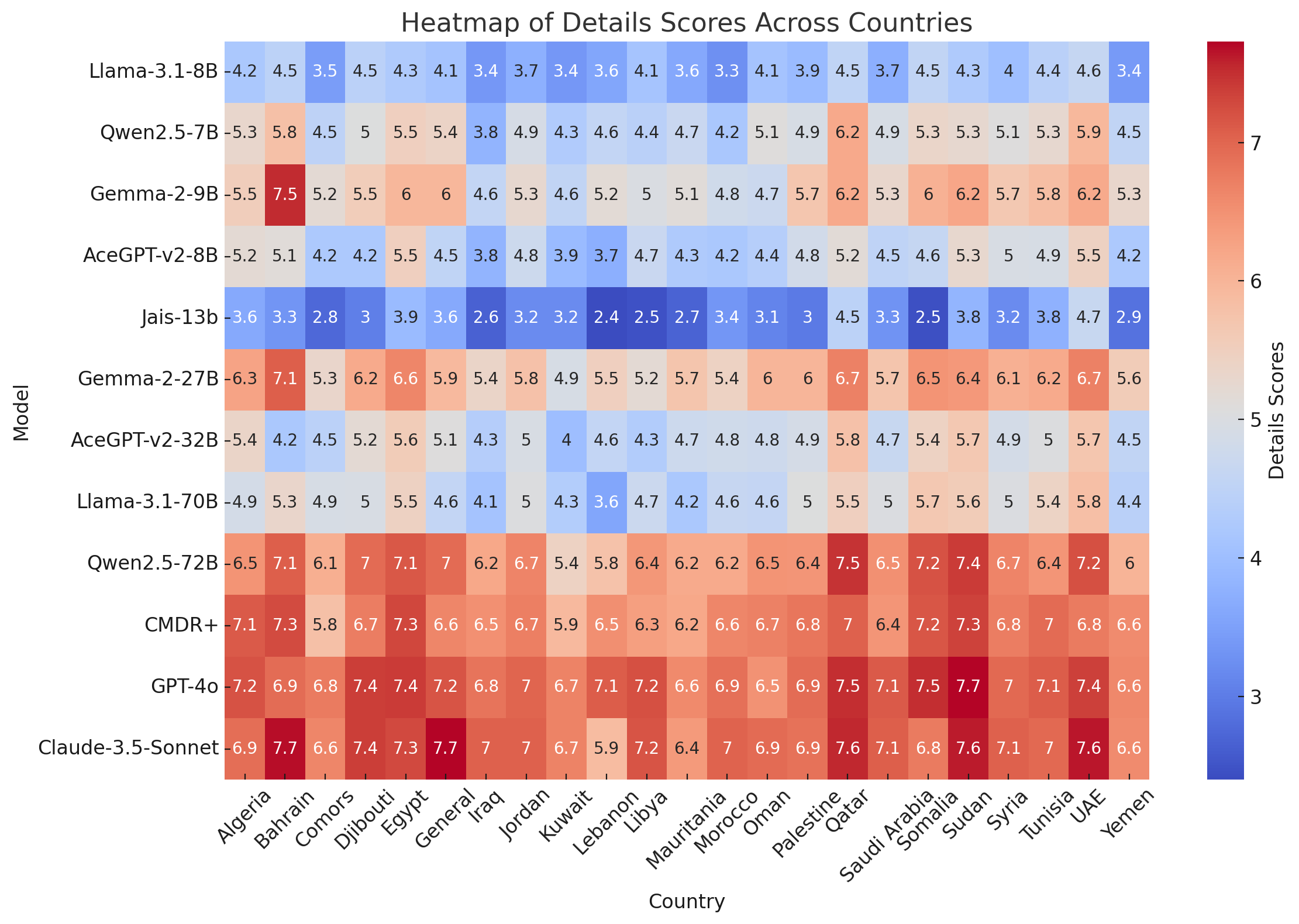}
        \caption{}
        \label{fig:Average_details_heatmap}
    \end{subfigure}
    \vskip\baselineskip

     \begin{subfigure}[b]{0.5\textwidth}
        \centering
        \includegraphics[width=\textwidth]{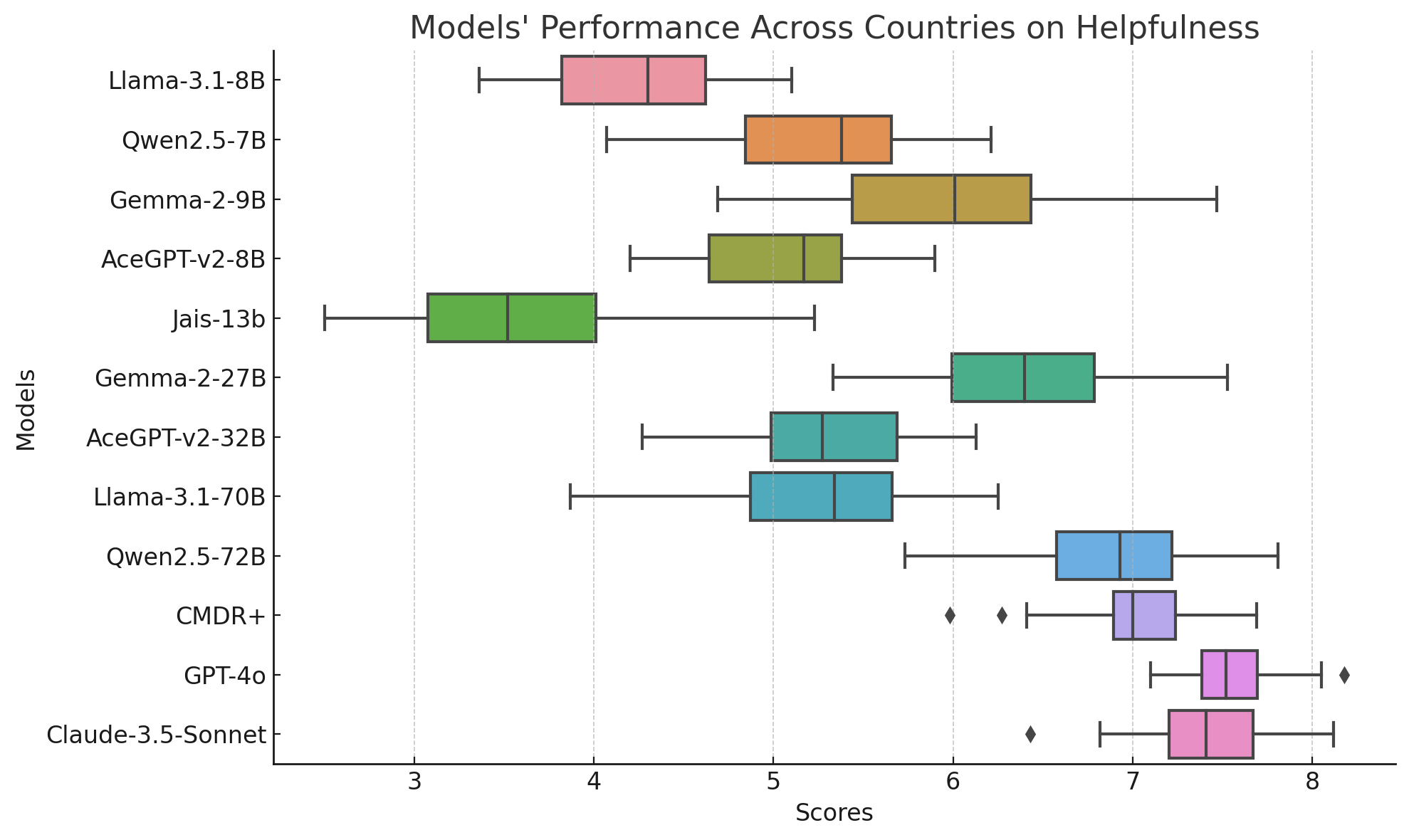}
        \caption{ }
        \label{fig:Average_helpfulness_boxplot}
    \end{subfigure}
    \hfill
    \begin{subfigure}[b]{0.49\textwidth}
        \centering
        \includegraphics[width=\textwidth]{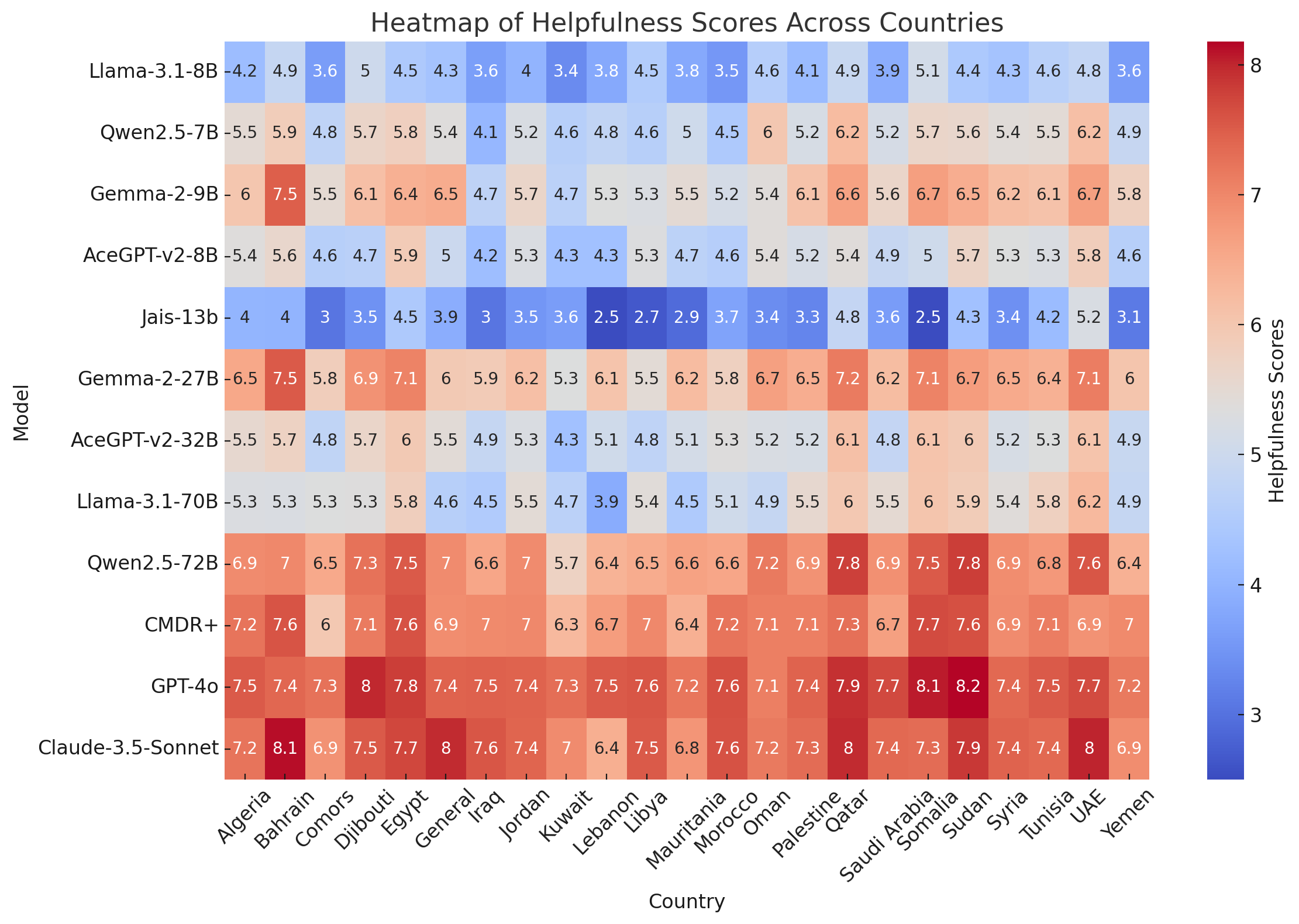}
        \caption{}
        \label{fig:Average_helpfulness_country_heatmap}
    \end{subfigure}

    \caption{Comparative analysis of the models across evaluation metrics: Coherence, Details, and Helpfulness.}
    \label{fig:three_metrics}
\end{figure*}

Results for the other metrics are presented in Appendix \ref{append:llm_judge_res}, namely \textit{coherence} (Figure~\ref{fig:Average_coherence_coherence_boxplot}, Figure~\ref{fig:Average_coherence_country_heatmap}), \textit{details} (Figure~\ref{fig:Average_details_country_boxplot}, Figure\ref{fig:Average_details_heatmap}), and \textit{helpfulness} (Figure~\ref{fig:Average_helpfulness_boxplot}, Figure~\ref{fig:Average_helpfulness_country_heatmap}).

\begin{figure}[]
  \centering
  \includegraphics[width=\columnwidth]{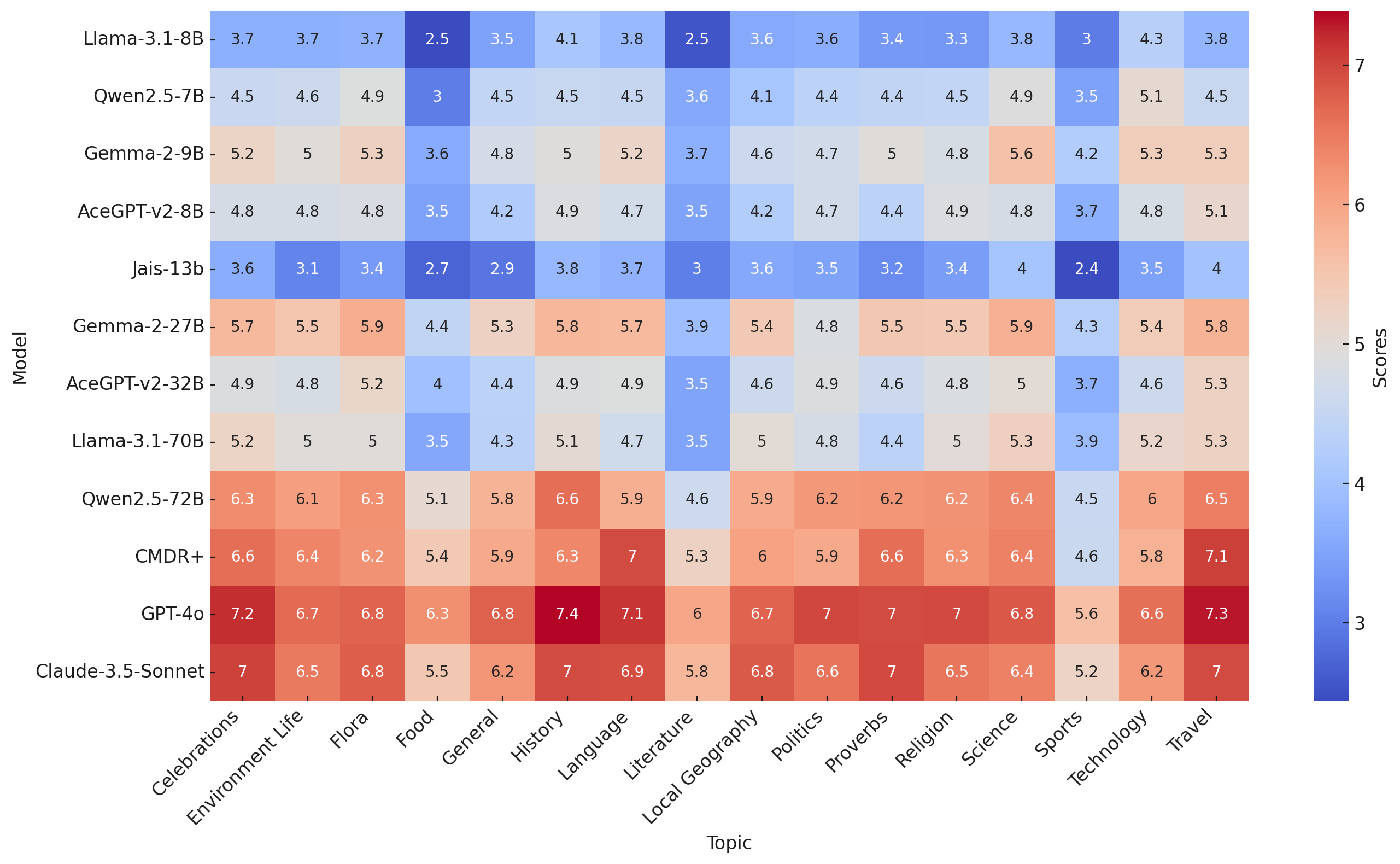}
  \caption{Performance of various models using Correctness score across different topics.
  }   
  \label{fig:avg_correctness_topics}
\end{figure}

\begin{table}[ht]
\centering
\resizebox{.24\textwidth}{!}{%

\begin{tabular}{lr}
\toprule
\textbf{Country}       & \textbf{Count} \\
\midrule
Egypt           & 146 \\
Jordan          & 140 \\
Mauritania      & 59  \\
Morocco         & 75  \\
Palestine       & 121 \\
Saudi Arabia    & 133 \\
Sudan           & 98  \\
Syria           & 164 \\
Tunisia         & 191 \\
UAE             & 137 \\
Yemen           & 250 \\
\midrule
General         & 20  \\
\midrule
Algeria         & 36  \\
Bahrain         & 5   \\
Comoros         & 66  \\
Djibouti        & 56  \\
Iraq            & 46  \\
Kuwait          & 75  \\
Lebanon         & 10  \\
Libya           & 11  \\
Oman            & 10  \\
Qatar           & 40  \\
Somalia         & 37  \\
\bottomrule
\end{tabular}
}
\caption{Number of samples per country for automatic evaluations.}
\label{tab:automatic_data_stats_pre_country}
\end{table}

\begin{figure}[h!]
    \centering
    \begin{subfigure}[b]{\columnwidth}
        \centering
        \includegraphics[width=\columnwidth]{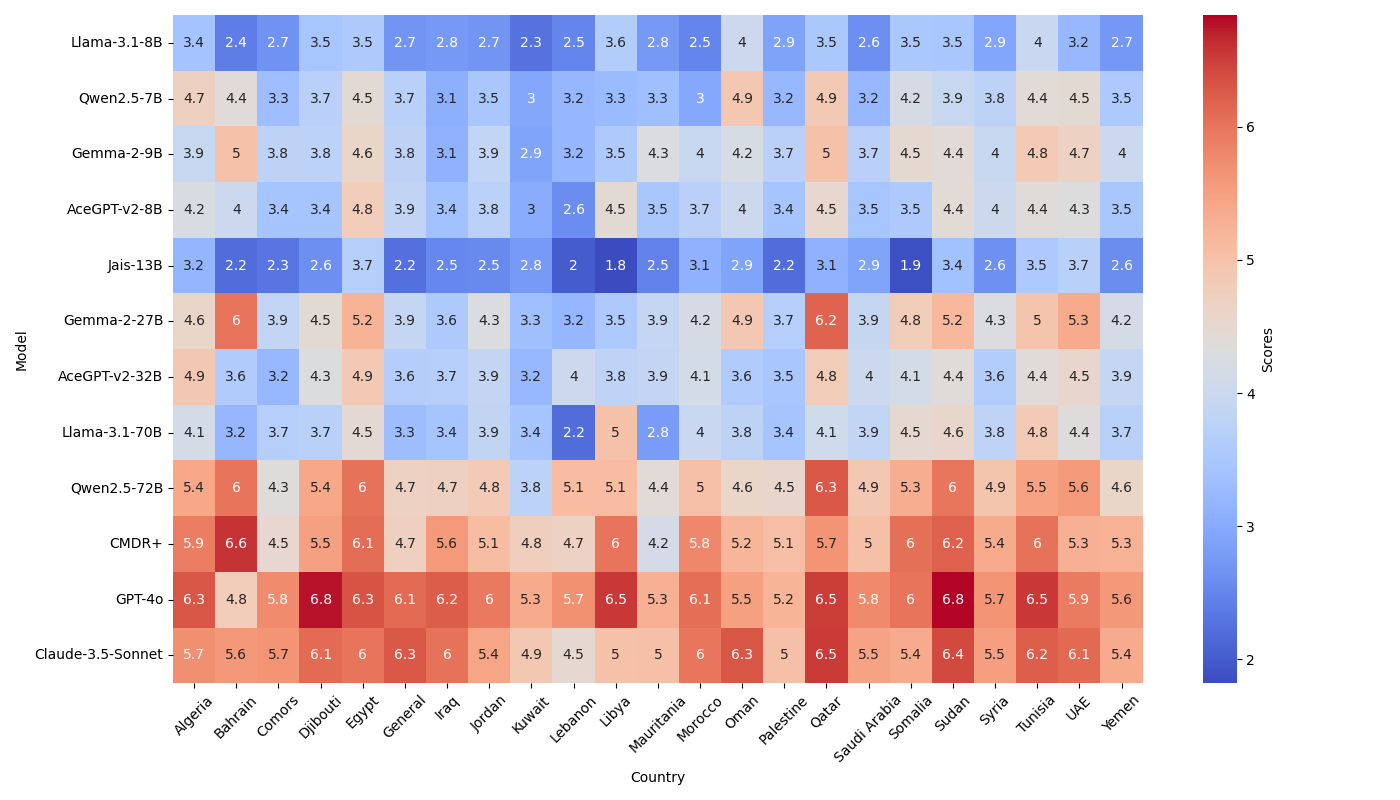}
        \caption{CMDR+ }
        \label{fig:cmdr_correctness_country_heatmap}
    \end{subfigure}
    \hfill
    \begin{subfigure}[b]{\columnwidth}
        \centering
        \includegraphics[width=\columnwidth]{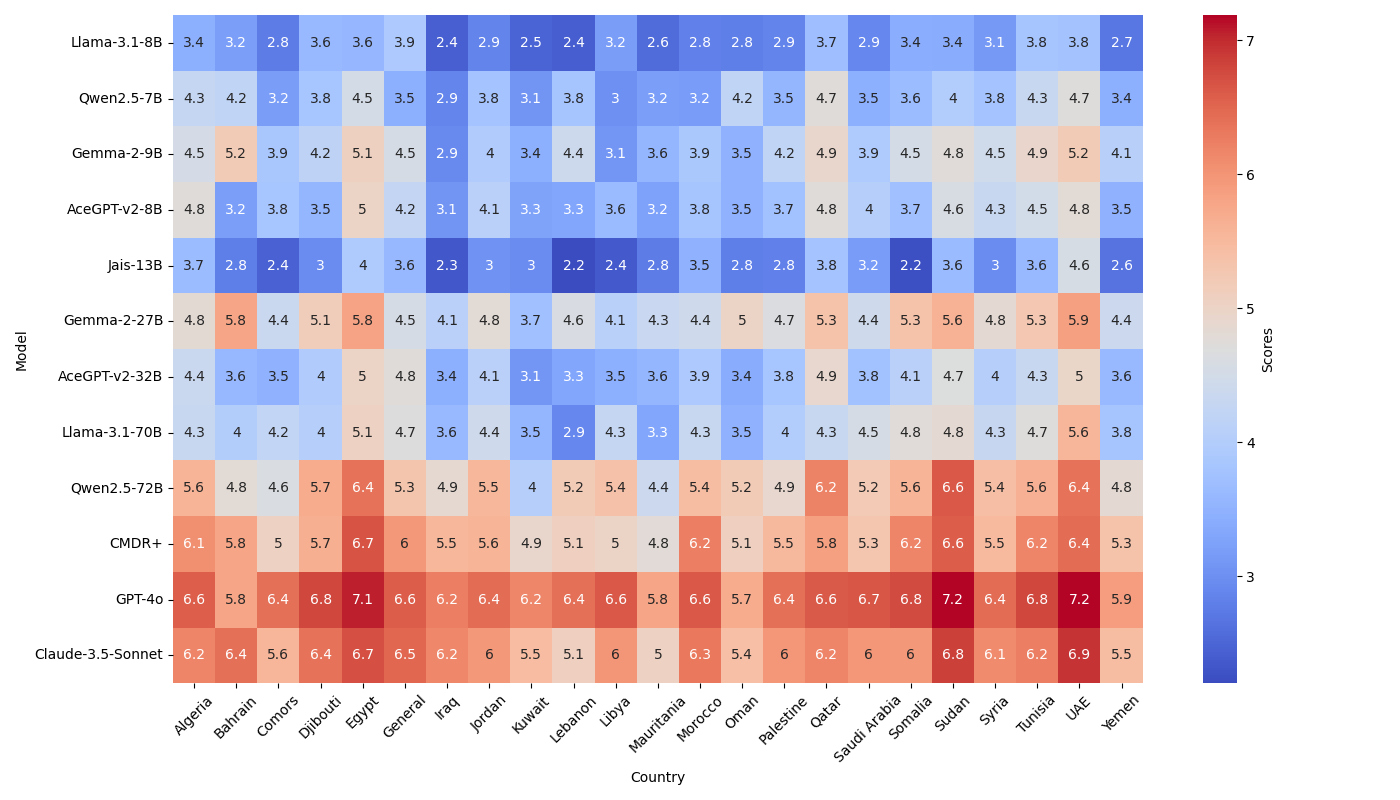}
        \caption{GPT-4o}
        \label{fig:gpt4_correctness_country_heatmap}
    \end{subfigure}
    \hfill
    \begin{subfigure}[b]{\columnwidth}
        \centering
        \includegraphics[width=\columnwidth]{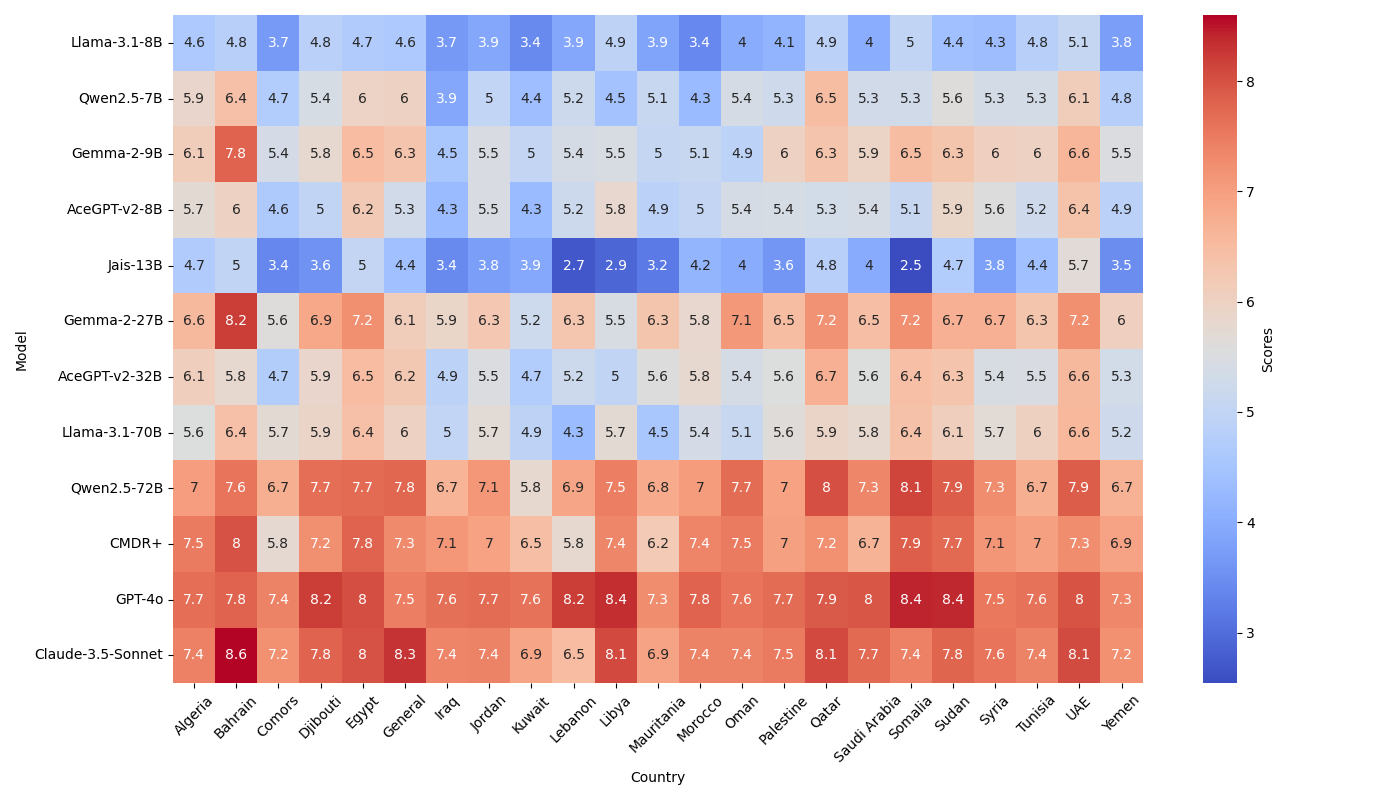}
        \caption{Qwen2.5-72B}
        \label{fig:qwen_correctness_country_heatmap}
    \end{subfigure}
    \caption{LLM-as-judge correctness scores across Arabic countries.}
    \label{fig:llm-judge-correctness}
\end{figure}

\subsubsection{Ablation Study on Model Size and Performance}\label{append:small_model_ablation}
\begin{table*}[ht]
\centering
\resizebox{0.6\textwidth}{!}{%
\begin{tabular}{lcccccc}
\toprule
\textbf{Country} & \textbf{Phi-3.5-Mini} & \textbf{Gemma-2-2B} & \textbf{Qwen2.5-1.5B} & \textbf{Qwen2.5-3B} & \textbf{LLaMA-3.2-1B} & \textbf{LLaMA-3.2-3B} \\
\midrule
Algeria        & 3.66 & 2.92 & 2.31 & 3.42 & 1.67 & 2.60 \\
Bahrain        & 3.00 & 4.00 & 2.20 & 3.40 & 1.40 & 2.20 \\
Comoros        & 1.79 & 2.70 & 1.80 & 2.23 & 1.85 & 2.02 \\
Djibouti       & 3.00 & 2.89 & 2.55 & 2.66 & 1.56 & 2.18 \\
Egypt          & 3.23 & 3.04 & 2.72 & 3.15 & 1.72 & 2.28 \\
General        & 2.45 & 2.65 & 2.47 & 2.55 & 1.55 & 2.10 \\
Iraq           & 2.43 & 2.74 & 1.33 & 1.78 & 1.64 & 1.74 \\
Jordan         & 3.08 & 2.71 & 2.31 & 2.84 & 1.67 & 2.18 \\
Kuwait         & 2.37 & 2.21 & 1.54 & 2.17 & 1.63 & 1.73 \\
Lebanon        & 1.90 & 1.70 & 2.20 & 2.70 & 1.40 & 1.50 \\
Libya          & 3.09 & 4.18 & 2.18 & 2.00 & 1.55 & 1.82 \\
Mauritania     & 2.86 & 3.04 & 2.18 & 2.75 & 1.69 & 1.92 \\
Morocco        & 2.14 & 2.48 & 1.73 & 2.38 & 1.42 & 1.96 \\
Oman           & 3.10 & 3.30 & 2.40 & 3.30 & 1.60 & 2.30 \\
Palestine      & 2.66 & 2.61 & 2.18 & 2.37 & 1.50 & 1.97 \\
Qatar          & 2.49 & 3.28 & 1.72 & 2.60 & 1.70 & 2.67 \\
Saudi Arabia   & 2.41 & 2.97 & 2.39 & 2.63 & 1.69 & 1.96 \\
Somalia        & 2.89 & 4.51 & 2.08 & 2.59 & 1.92 & 2.54 \\
Sudan          & 3.06 & 3.32 & 2.24 & 2.64 & 1.66 & 2.30 \\
Syria          & 2.83 & 2.54 & 2.19 & 2.51 & 1.61 & 2.20 \\
Tunisia        & 3.82 & 3.24 & 2.67 & 3.51 & 1.84 & 2.60 \\
UAE            & 3.27 & 3.27 & 2.47 & 3.14 & 1.79 & 2.41 \\
Yemen          & 2.93 & 3.02 & 1.88 & 2.69 & 1.69 & 2.01 \\
\bottomrule
\end{tabular}
}
\caption{Average correctness scores of Small LLMs models across countries using CMDR+.}
\label{tab:small_llm_scores}
\end{table*}
In our ablation study, we observed a notable disparity in performance between smaller and larger LLMs. As indicated in Table~\ref{tab:small_llm_scores}, the smaller models—those with fewer than 4 billion parameters—achieved an average correctness score of 2.40, whereas the larger models attained a significantly higher average score of 4.01 on the correctness metric. This suggests that smaller LLMs may lack the capacity to effectively handle the complexity of the culturally rich and dialectal content in our dataset. Consequently, we did not include these smaller models in our primary comparisons with larger LLMs. The correctness scores were evaluated using the \texttt{Command R+} model as the evaluator, providing a consistent benchmark across all assessments.

\subsection{Human Evaluation Results}
Table \ref{tab:human_eval_msa} presents average human-evaluation correctness scores for MSA instructions per country.

\begin{table*}[b]
\centering
\resizebox{0.60\textwidth}{!}{
\begin{tabular}{lccccc}
\toprule
\textbf{Country} & \textbf{AceGPT-v2-32B} & \textbf{Llama-3.1-8B} & \textbf{Qwen2.5-72B} & \textbf{Claude-3-5-Sonnet} & \textbf{Jais-13b} \\ \midrule
Algeria     & 6.58 & 5.50 & 5.33 & 5.17 & 6.67 \\ 
Bahrain     & 3.25 & 3.00 & 3.33 & 5.00 & 2.58 \\ 
Comoros     & 5.67 & 5.00 & 2.17 & 4.17 & 4.08 \\ 
Djibouti    & 5.33 & 4.25 & 5.42 & 4.08 & 3.67 \\ 
Egypt       & 5.83 & 5.00 & 5.17 & 6.83 & 4.33 \\ 
General     & 5.75 & 1.92 & 3.92 & 4.58 & 4.45 \\ 
Iraq        & 3.83 & 3.42 & 4.08 & 5.75 & 3.67 \\ 
Jordan      & 2.92 & 2.33 & 4.25 & 5.00 & 4.67 \\ 
Kuwait      & 2.83 & 3.83 & 2.50 & 3.58 & 3.09 \\ 
Lebanon     & 4.58 & 2.42 & 4.67 & 5.75 & 2.92 \\ 
Libya       & 3.92 & 2.58 & 5.33 & 5.08 & 2.33 \\ 
Mauritania  & 1.67 & 2.00 & 2.75 & 3.27 & 1.67 \\ 
Morocco     & 3.50 & 2.33 & 3.17 & 4.92 & 4.92 \\ 
Oman        & 4.00 & 3.00 & 5.50 & 4.92 & 3.50 \\ 
Palestine   & 3.25 & 2.83 & 3.75 & 5.17 & 3.92 \\ 
Qatar       & 2.92 & 2.25 & 3.75 & 3.08 & 3.17 \\ 
Saudi Arabia & 1.09 & 3.42 & 5.00 & 4.25 & 1.75 \\ 
Somalia     & 3.83 & 4.75 & 5.17 & 5.50 & 3.50 \\ 
Sudan       & 4.33 & 4.17 & 5.92 & 4.42 & 3.58 \\ 
Syria       & 5.67 & 5.50 & 6.40 & 7.25 & 4.17 \\ 
Tunisia     & 4.92 & 6.58 & 4.08 & 4.36 & 3.18 \\ 
UAE         & 3.83 & 2.75 & 4.83 & 4.58 & 2.42 \\ 
Yemen       & 5.17 & 4.75 & 4.67 & 5.58 & 3.75 \\ \bottomrule
\end{tabular}}
\caption{Average human-evaluation correctness scores for MSA instructions per country.}
\label{tab:human_eval_msa}
\end{table*}

\end{document}